\documentclass{article}
\usepackage{frExamplee}
\usepackage{graphicx}
\usepackage{setspace}
\usepackage{booktabs}

\usepackage{siunitx}
\usepackage{mathtools}
\usepackage{gensymb}
\usepackage{float}
\usepackage[tight]{subfigure}
\usepackage{caption} %

\usepackage{tikz}
\usetikzlibrary{decorations.text,calc,shapes.geometric,shapes.callouts}

\usepackage{xspace}
\usepackage{color}
\usepackage[textsize=scriptsize]{todonotes} %
\newcommand{\revision}[1]{\textcolor{black}{#1}\xspace}

\usepackage{natbib}
\usepackage{arydshln}

\title{Towards Automated Sample Collection and Return\\ in Extreme Underwater Environments}

\author{Gideon Billings\\
Naval Architecture and Marine Engineering\\
University of Michigan\\
Ann Arbor, MI 48109, USA\\
\texttt{gidobot@umich.edu}
\And
Matthew R.\ Walter\\
Toyota Technological Institute at Chicago\\
Chicago, IL, 60637, USA
\And
Oscar Pizarro\\
Australian Centre for Field Robotics\\
University of Sydney\\
NSW 2006, Australia
\And
Matthew Johnson-Roberson\\
Naval Architecture and Marine Engineering\\
University of Michigan\\
Ann Arbor, MI 48109, USA
\And
Richard Camilli\\
Applied Ocean Physics and Engineering\\
Woods Hole Oceanographic Institution\\
Deep Submergence Laboratory\\
Woods Hole, MA, 02543, USA
}

\usepackage{hyperref}

\hypersetup{
    colorlinks=true,
    citecolor=black,
    linkcolor=blue,
    urlcolor=blue,
    pdfauthor={Gideon Billings, Richard Camilli, Matthew R. Walter, Matthew Johnson-Roberson},
    pdftitle={Towards Automated Sample Collection and Return in Extreme Underwater Environments}
}

\begin{document}

\newpage
\maketitle

\begin{abstract}

\revision{In this report, we present the system design, operational strategy, and results of coordinated multi-vehicle field demonstrations of autonomous marine robotic technologies in search-for-life missions within the Pacific shelf margin of Costa Rica and the Santorini-Kolumbo caldera complex, which serve as analogs to environments that may exist in oceans beyond Earth. This report focuses on the automation of ROV manipulator operations for targeted biological sample-collection-and-return from the seafloor. In the context of future extraterrestrial exploration missions to ocean worlds, an ROV is an analog to a planetary lander, which must be capable of high-level autonomy. Our field trials involve two underwater vehicles, the SuBastian ROV and the Nereid Under Ice (NUI) hybrid ROV for mixed initiative (i.e., teleoperated or autonomous) missions, both equipped 7-DoF hydraulic manipulators. We describe an adaptable, hardware-independent computer vision architecture that enables high-level automated manipulation. The vision system provides a 3D understanding of the workspace to inform manipulator motion planning in complex unstructured environments. We demonstrate the effectiveness of the vision system and control framework through field trials in increasingly challenging environments, including the automated collection and return of biological samples from within the active undersea volcano, Kolumbo. Based on our experiences in the field, we discuss the performance of our system and identify promising directions for future research.}

\end{abstract}

\section{Introduction}

A growing body of evidence suggests that the Earth is not unique in containing liquid water~\citep{lewis1971satellites, gaeman2012sustainability, malin2000evidence, khurana1998induced}, an essential ingredient for carbon-based life. Recent indications of water geysers emanating from moons of Saturn and Jupiter, including Enceladus~\citep{nimmo2007shear} and Europa~\citep{arnold2019magnetic}, suggest that they may contain subsurface oceans with active hydrothermal venting~\citep{lowell2005hydrothermal, hsu2015ongoing}. Here on Earth, ocean floor hydrothermal systems and cold seep sites have long been known to host diverse chemosynthetic ecosystems that rely on the redox potentials of deep Earth fluids emitted from these sites to derive biochemical energy~\citep{jannasch1979chemosynthetic, brooks1987deep}, and may serve as analogs for oases of life elsewhere in our solar system and beyond. However, exploration for life within the distant oceans of Europa and Enceladus remains a daunting technological challenge. Robotic submersible vehicles equipped with manipulators provide a practical means for sample analysis and collection, enabling flexibility and dexterity without requiring precise and energetically costly positioning of the vehicle. Planetary landers such as the Mars Rovers have historically relied on human teleoperated manipulation using manually generated scripts~\citep{8396726, fong2001collaborative, leger2005remote} to collect samples. However, teleoperation of robotic subsea vehicles within these putative ocean worlds is impractical because of high communication latencies (e.g., on the order of an hour for Europa). Thus, robotic missions must be capable of fully automated manipulation.

Marine robotic platforms such as remotely operated vehicles (ROVs) equipped with manipulators provide a useful testbed to develop automated manipulation and sampling technologies as analogs for space missions. Although Earth's gravitational constant is higher than Europa and Enceladus, these moons' estimated ice thicknesses of up to 30\,km~\citep{Iess78, billings2005great} are expected to present operational challenges, such as extreme pressure, near-freezing temperatures, and corrosion that are similar to Earth's deep ocean environments. While autonomous underwater vehicles (AUVs) have been used for under-ice surveys for nearly 50 years~\citep{francois1972unmanned}, deep ocean missions that require sample collection and return using manipulator arms are generally conducted using ROVs under direct human piloted control with cable-tethered communication. Only limited attempts at autonomous manipulation have been made in natural ocean environments~\citep{sauvim2009, SHIM2010, rauvi2010, sanz2013trident, triton2015, SIVCEV2018153}. \revision{The comparative lag in subsea manipulator autonomy behind terrestrial systems can be at least partly attributed to commercial systems being historically designed for direct teleoperation, with limited command modes and feedback, low control loop frequency, and poor repeatability~\citet{SIVCEV2018431}. Despite these challenges, we demonstrate an automation framework that is compatible with existing commercial manipulator systems and that automates many high level tasks, while reducing risk through visual based scene understanding and pilot supervision.}

\begin{figure}[!t]
    \centering
    \includegraphics[width=\linewidth]{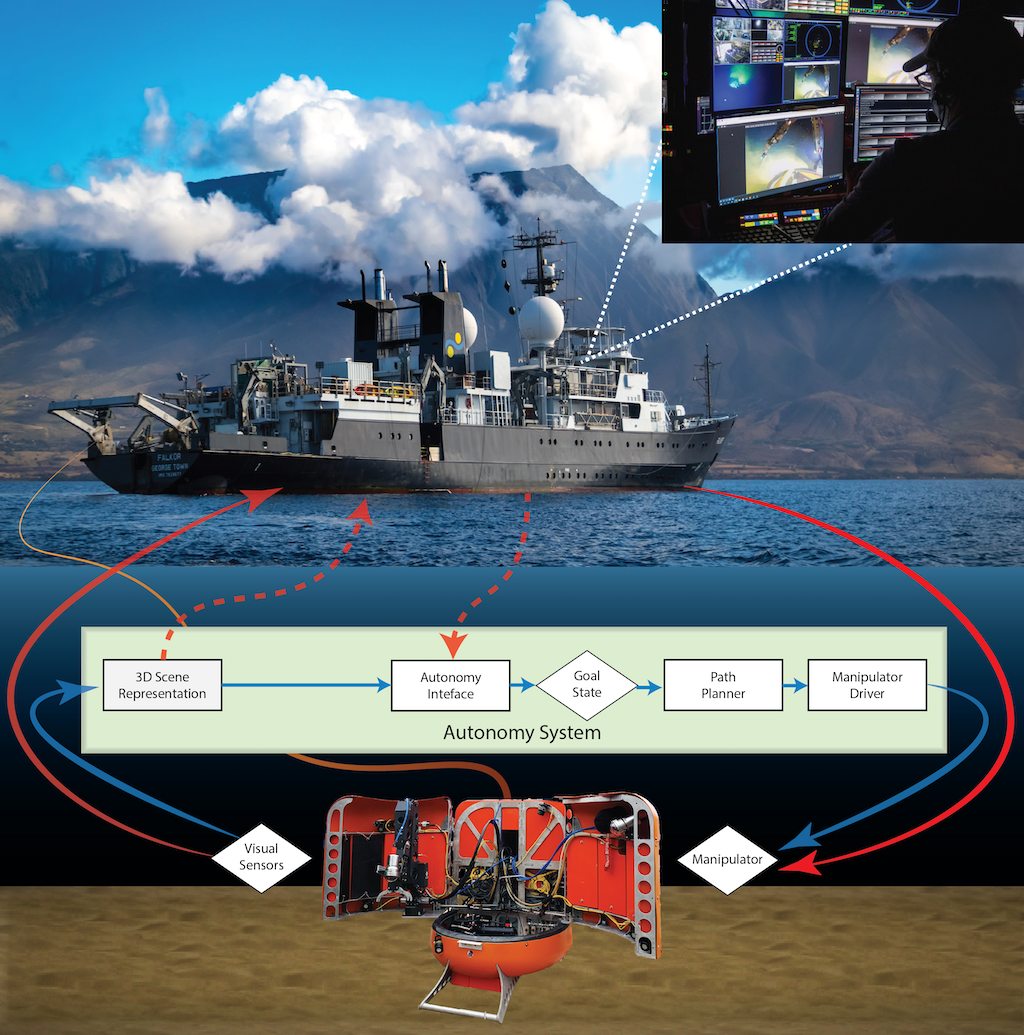}
\caption{Conceptual graphic of the our control system for an underwater intervention vehicle. The autonomy system runs on a topside desktop computer with visual sensor data and manipulator coms streamed over a high bandwidth tether from the vehicle. Solid \textcolor{red}{red} flow lines represent standard teleoperated control from a surface ship. \textcolor{blue}{Blue} flow lines represent our automated system. \textcolor{red}{Red} dashed lines represent interfacing between the pilot and the autonomous system, where, in this work, the pilot acts as the high level task planner and interfaces with the automated system through a graphical scene representation and task level controller. Eventually, the pilot would be replaced with an automated mission planner that could issue high level tasks.}
    \label{fig:Concept}
\end{figure}

In this paper, we consider the challenge of automated subsea manipulation and sample collection using existing ROV platforms as a technology analog for an under-ice exploration missions to Europa or Enceladus. We discuss the challenges that deep seafloor environments pose to automated robotic intervention and propose an architecture that overcomes many of these challenges. The system that we describe can be integrated on existing ROVs with minimal hardware requirements, namely, a vehicle-mounted stereo camera and a manipulator-mounted fisheye camera. We investigate the practical use of our perception methods to estimate the vehicle configuration, dynamically localize tools, and ground the transform between the natural scene reconstruction and the structured vehicle workspace. The manipulator control and vision processes that serve as the basis of this automation framework can be readily adapted to a variety of hardware configurations, making them suitable for a wide range of robotic platforms, including space flight systems. We demonstrate the flexibility of this framework through separate field trials performed with two different classes of ROVs equipped with substantively different manipulators. \revision{Figure~\ref{fig:Concept} shows a conceptual diagram of how our system integrates with an ROV. In the current system implementation, a topside machine performs all processing using camera and manipulator data streamed from the vehicle over a high-bandwidth tether.}

We conducted testing and field trials in progressively challenging environments, initially in laboratory settings and tank testing, followed by 11 dive missions at the Central American Pacific shelf margin of Costa Rica to operational depths of approximately 1800\,m. This area of the Costa Rican accretionary prism is a well studied region with localized ocean floor fluid expulsion sites that host diverse assemblages of extremophile organisms~\citep{hensen2004fluid, krause2014microbial, levin2015biodiversity, sahling2008fluid, silver2000fluid}. Following the completion of the Costa Rica expedition, our team conducted a series of five dive missions to depths of 500\,m within the potentially hazardous craters of the Kolumbo and Santorini calderas. These sites of active volcanism contain localized areas of high-temperature hydrothermal venting causing environmental hypercapnia~\citep{camilli2015kallisti, carey2013co2}, which host non-calcifying chemosynthetic organisms that may resemble those that arose early on Earth, prior to the advent of its oxidizing atmosphere.

The paper is organized as follows. We begin in Section~\ref{section:background} with an overview of previous work on automated underwater manipulation.  Section~\ref{section:system} briefly outlines our strategy for mission planning and operations, and describes in detail the architecture of our automated manipulation system. Section~\ref{section:results} examines the results of experimental missions during field demonstrations using the \textit{SuBastian} ROV~\citep{campbell_bingham_williams_2019} and \textit{Nereid Under Ice (NUI)} Hybrid Remotely Operated Vehicle (HROV)~\citep{bowen2014design}. Section~\ref{section:discussion} draws on these field results and experiences to discuss advances as well as the limitations and potential failure modes of our perception and control methods and examines how this research may help to advance both automated ROV operations here on Earth and future space flight missions to explore for life within ocean worlds elsewhere. Section~\ref{section:conclusion} identifies promising directions for future research.

\section{Background}
\label{section:background}

There is a rich body of literature on underwater vehicle manipulator system (UVMS) control. This section provides a brief review of the work most related to our approach which have demonstrated their methods in experimental trials. For a thorough discussion of the prior work on UVMS systems, we refer the reader to~\citet{SIVCEV2018431}.

 \citet{ishimi1991manipulation} and \citet{broome1995subsea} describe some of the pioneering work on automating UVMSs, where demonstrations included 3D graphical renderings of an ROV's configuration and workspace, real-time visualization of manipulator motion plans, and Cartesian space end-effector control. \revision{\citet{yann2015} demonstrated free-floating stereovision based servoing of the \textit{Twin-Burger 2} AUV outfitted with an unconventional 2-DoF manipulator to collect vacuum samples of free-floating targets in a pool.} More recent works under the large-scale research projects RAUVI~\citep{rauvi2010}, TRIDENT~\citep{sanz2013trident}, TRITON~\citep{triton2015}, PANDORA~\citep{pandora2015}, and MARIS~\citep{simetti2017autonomous} focus on tightly coupled control of the 140\,kg displacement \textit{Girona 500} AUV outfitted with a customized electric manipulator to perform free-floating intervention tasks. The PANDORA project explores the ability to learn the vehicle and manipulator trajectories by demonstration. The other projects combine vehicle and manipulator motion generation under a task priority framework, where the manipulator control law is a function of the vehicle velocity. Building on these works, the MERBOTS project~\citep{merbots2017} offers a significant advancement towards automated UVMS control by integrating the ROS-based MoveIt! motion planning framework with the intervention AUV to generate combined vehicle and manipulator motion trajectories in Cartesian space for free-floating intervention tasks. \revision{In very recent work, \citet{Nishida2019} demonstrated free-floating autonomous sample collection of shells from the seafloor using the 400\,kg displacement \textit{Tuna-Sand2} AUV outfitted with a custom extendable suction device.} While this body of work provides key advancements towards automated free-floating intervention, limitations make it difficult for many actively operated UVMSs to adopt these methods. Among them, integrating such a tightly coupled control system with existing UVMS platforms would require significant modification to the software architecture, which is particularly problematic for commercial systems. Additionally, the dynamic coupling effect between the vehicle and manipulator during free-floating intervention can strongly affect the trajectory tracking performance, necessitating very slow actuation of the vehicle and manipulator. Lastly, this control approach is designed for high-precision electric manipulators that support velocity-based control, whereas most manipulators on operational UVMS platforms are hydraulic and support only position set point commands with limited precision and repeatability.

\revision{Hydraulic manipulators have orders of magnitude higher power-to-weight ratios compared to their electric counterparts and are generally more reliable, making them the manipulator of choice for commercial ROV systems. Though recent commercial electric manipulators have entered the market, their significantly higher power requirements make them practical only for ROVs that have power supplied over a tether. For vehicles like the \textit{NUI} HROV, which carries all power onboard, low-power hydraulic manipulators remain the most practical choice. However, the limited precision and feedback of hydraulic manipulators present challenges for automation, and little work exists that addresses these challenges.} \citet{hildebrandt2009} demonstrated precision control of a hydraulic manipulator to plug a deep-sea connector. \citet{SHIM2010} %
perform pre-programmed motion following and operator control of a hydraulic work class manipulator. Using Cartesian space end-effector control, they demonstrate operator-guided push-core sampling in the deep ocean. \citet{zhang2019} perform visual servoing and target grasping with a custom 7-DoF hydraulic manipulator. \citet{SIVCEV2018153} demonstrate impressive visual servoing of a working class hydraulic manipulator using position-based control, with feedback provided by fiducials detected from a wrist-mounted camera. Their results include grasping and turning T-bar valves and tracking targets in motion with the end-effector.

Our system builds on these prior approaches to UVMS control, where we demonstrate the effective integration of the MoveIt!~motion planning framework~\citep{moveit2014} with a work class ROV manipulator system for automated planning and control in obstructed scenes. We take a decoupled approach to manipulator control that assumes the vehicle holds station (i.e., rests on the bottom) during the manipulation task. This assumption is motivated by the goal of having the system widely transferable among existing ROV systems. This decoupled approach enables our manipulator control system to be integrated externally from the existing UVMS control systems, providing high-level autonomy with flexibility to be integrated onto a wide array of ROV classes and manipulator arms, including both electric and hydraulic systems.

Important to automating UVMSs are the problems of visual scene understanding and target localization, whether the target be a tool to grasp, a valve to turn, or a sample location in an unstructured environment. \revision{Subsea perception is a particularly challenging problem for a number of reasons: turbidity degrades image quality; evenly lighting the scene is very difficult; variable wavelength-dependent absorption and scattering properties of the water column attenuate light and reduce color and photometric contrast; and gathering underwater datasets for developing computer vision methods is expensive. Despite these challenges, computer vision remains the primary means of performing target localization for automated UVMS platforms. Most prior works on UVMS automation rely on fiducials or known geometric shapes that retain high contrast underwater.} \citet{sauvim2009} use large spherical markers attached to a target and a circle shape edge detector algorithm to localize the marker from a video feed. \citet{sanz2013trident} localize a black box object on a harbor seabed by first constructing a visual mosaic from a pre-intervention survey dive with a downward-facing stereo camera, and then matching an image template of the black box to the mosaic. \citet{pandora2015} localize a known panel during intervention operations by registering interest points against a template image. They then estimate the orientation of valves on the panel based on edge detection. \citet{triton2015} and \citet{merbots2017} use fiducial markers to localize a panel with \textit{a priori} known relative positions of the turn valves and connector plugs. \citet{merbots2017} also use fiducials on the end-effector of the manipulator to update the manipulator calibration in real-time. \citet{simetti2017autonomous} use color and geometric shape segmentation of RGB images to detect the pose of a cylindrical pipe of known size. Under the DexROV project, \citet{Birk2018DexterousUM} process stereo point clouds into a 3D occupancy map, while also using fiducial markers to detect and localize a panel with known structure that was projected into the planning scene.

Building on the long history of fiducials as a robust visual cue for underwater computer vision methods, our work extends the use of fiducials to detect the pose of graspable tools carried on-board the ROV, estimate dynamic vehicle configurations in real-time, and ground the relative reference frames in the planning environment. We demonstrate the use of fiducials in a way that is practical for field deployments with an underlying vision system that can effectively localize tools and target objects within the workspace, as well as reconstruct the workspace for obstructed motion planning. Fiducials also enable the collection of annotated image datasets in natural deep seabed environments that support the development of advanced perception methods for scene reconstruction, and target detection and localization.

\section{System Overview}
\label{section:system}

The following sections provide an overview of the different components of our system, including the mission architecture for field operations and the methods for perception and control.

\subsection{Mission and Vehicle Platform Architecture}

\begin{figure}[!t]
    \centering
    \includegraphics[height=3in]{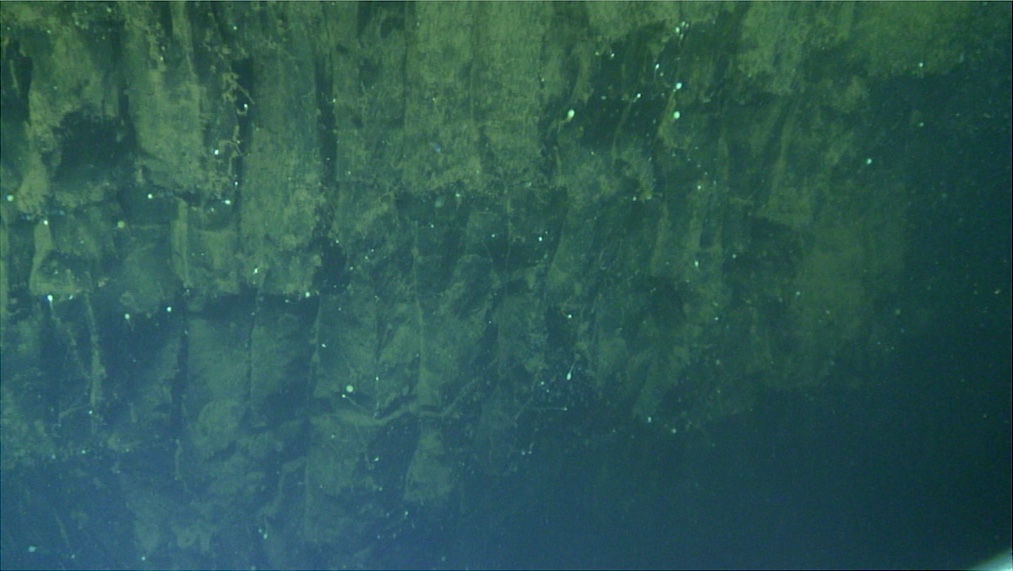}
    \caption{Photograph taken by the \textit{NUI} vehicle within the Kolumbo volcano crater that shows an overhanging vertical wall of columnar lava. Colonization of the lava surfaces by relatively uncommon lollipop sponges (\textit{Stylocordyla pellita}) are visible as white dots within the image.}
    \label{fig:Columnar_lava_w_overhang}
\end{figure}

Field demonstration and validation include two research cruises, conducted east of the Cocos and Caribbean tectonic subduction zone along Central America's Pacific continental margin (9.0\,N 84.5\,W), and within the Kolumbo and Santorini Calderas of the Hellenic volcanic arc in the southern Aegean Sea (36.52\,N 25.48\,E and 36.45\,N 25.39\,E, respectively). The sites, which are known to host oases of chemosynthetic communities associated with hydrothermal and seafloor hydrocarbon seeps, were chosen as NASA TRL-6 demonstration locations for analog astrobiology exploration missions. These campaigns utilized a sequentially nested survey method with a coordinated team of heterogeneous robotic platforms that relied on automated planing tools to rapidly synthesize vehicle missions in response to newly acquired information~\citep{vrolijk2021using, ayton2019measurement}. To better approximate an analog space flight mission scenario, surface ships operated as orbiters, conducting multibeam sonar bathymetric mapping of the Pacific~\citep{CostRicaData} and Aegean~\citep{nomikou2019advanced} campaign sites, with coverage areas of \SI{2,000}{\,km\squared} at 30\,m resolution and  \SI{48}{\,km\squared} at 10\,m resolution, respectively. These maps informed the mission planning for autonomous underwater gliders (AUG), which acted as long-range in-situ reconnaissance drones, conceptually similar to NASA's \textit{Ingenuity} and \textit{Dragonfly} vehicles, conducting reconnaissance missions of between 1\,km and 500\,km in length at standoff distances to within 15\,m of ocean floor obstacles in order to identify potential areas of scientific interest~\citep{vrolijk2021using, camilli2020improving}. Automated AUG mission planning considered resource (e.g., time and power) and risk constraints~\citep{timmons2016preliminary, timmons2019risk}, and adaptively replanned missions based on inferred sites of scientific interest that correlated with the presence of active ocean floor hydrocarbon cold seeps and hydrothermal vents. Using information gained by the surface ship sonar and AUG missions, the automated planning process then generated viable mission sequences that the ROV used to investigate areas of highest estimated information gain~\citep{vrolijk2021using}. During these missions, the ROV acted as a lander, outfitted with a manipulator for automated sample collection and return. The hazardous deep ocean environments explored as part of these ROV missions are considered probable analogs for environments (Fig.~\ref{fig:Columnar_lava_w_overhang}) that may exist on other ocean worlds.

\begin{table}[h!]
\caption{Comparison of \textit{SuBastian} and \textit{NUI} configurations.}
\begin{center}
\resizebox{16cm}{!}{
  \begin{tabular}{cccccccccc}
     \toprule
      \multicolumn{1}{c}{} &
        \shortstack{Depth rating \\ meters} &
        \shortstack{Displacement \\ kilograms} &
        \shortstack{Lateral excursion \\ (tethered) meters} &
        \shortstack{Power draw \\ (typical) watts} &
        \shortstack{Endurance \\ hours} &
        \shortstack{Manipulator \\ type} &
        \shortstack{Manipulator reach \\ meters} &
        \shortstack{Payload capacity \\ kilograms}\\ %
      \midrule
     \textit{SuBastian} & $4500$ & $3200$ & $<500$ & $~40000$ & unlimited & $2$ x  $7$-DoF & $1.9$ & $200$ \\
      \textit{NUI} & $2000$ & $2000$ & $20000$ & $\;\hphantom{0}2500$ & $6$ to $8$ & $ 7$-DoF & $1.3$ & $100$\\
      \bottomrule
  \end{tabular}}
\end{center}
\label{tab:ROV comparison}
\end{table}
The two ROVs used for these demonstration campaigns, \textit{SuBastian} and \textit{NUI} are substantially different in design and purpose (Table~\ref{tab:ROV comparison}). \textit{SuBastian} is an exemplar of modern deep ocean work class ROVs, with its power, communications, and navigation net provided via an armored cable by its attendant surface ship, the \textit{R/V Falkor}. \textit{SuBastian} is equipped with twin 7-DoF Schilling Titan-4 hydraulic manipulator arms (Schilling Robotics, Davis, California) and is a fully teleoperated vehicle that can operate at horizontal excursions of up to 500\,m laterally from the \textit{R/V Falkor}. In contrast, \textit{NUI} is a HROV that relies on its own battery power and uses an un-armored fiber optic link (roughly the diameter of a human hair) for optional communication with an attendant surface ship, and can operate as both an ROV and an AUV. When in tethered ROV mode, \textit{NUI}'s power and telemetry architecture enables lateral excursions of up to 20\,km from the attendant surface ship. To aid hydrodynamic efficiency, \textit{NUI} has articulating bow doors that can be closed and act as a fairing during transits and AUV missions. In contrast to \textit{SuBastian}'s twin Titan-4 architecture, which is configured to maximize ROV work area and dexterity, NUI's starboard door is equipped with a single, custom 7-DoF hydraulic manipulator (Kraft Telerobotics, Overland Park, Kansas) that is optimized for energy efficiency. This emphasis on efficiency comes at the expense of reductions in available payload and work space, the usable range of motion, control precision, lighting field, and available viewing perspectives.

\subsection{Perception}

\begin{figure}[!t]
    \centering
    \subfigure[Wrist-mounted fisheye camera]{\includegraphics[height=3in]{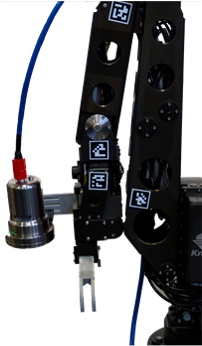}\label{fig:fisheye-nht-rov}}\hfil
    \subfigure[Vehicle-mounted stereo camera]{\includegraphics[height=3in]{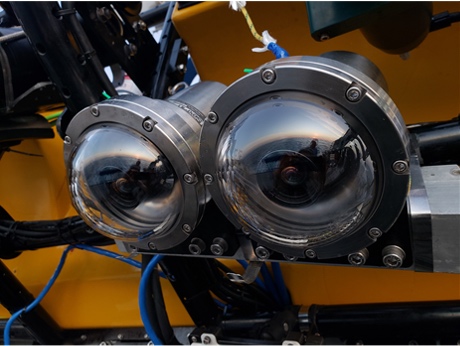}\label{fig:stereo-nht-rov}}
    \caption{The vision system for autonomy is composed of \subref{fig:fisheye-nht-rov} a wrist-mounted fisheye camera and \subref{fig:stereo-nht-rov} a vehicle-mounted stereo pair (shown here mounted on the \textit{SuBastian} ROV). The vision system can be easily integrated onto existing vehicles.}
    \label{fig:nht_smirc}
\end{figure}

\begin{figure}[!t]
    \centering
    \subfigure[Far fisheye view]{\includegraphics[width=0.45\linewidth]{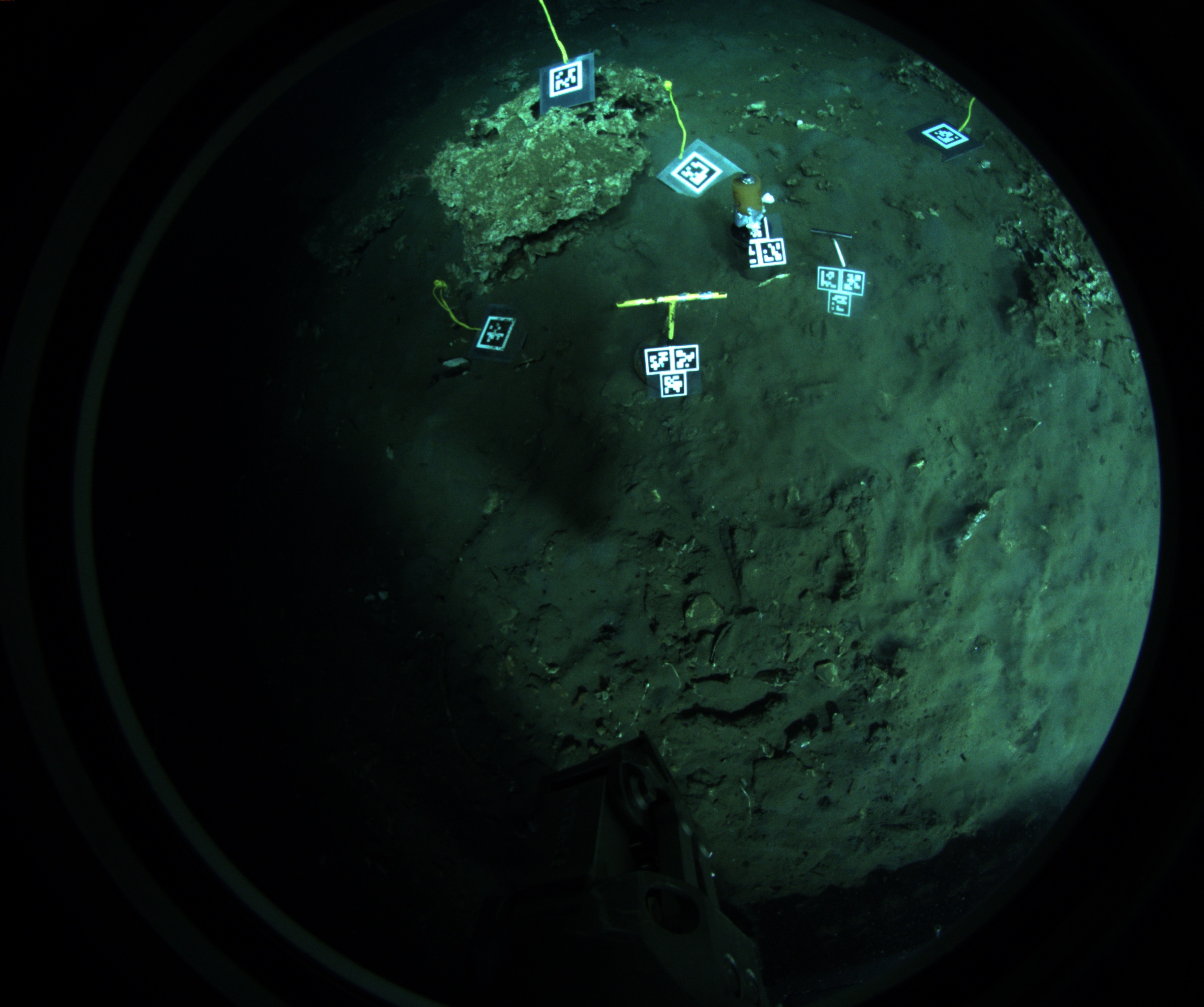}}\hfil
    \subfigure[Close fishye view]{\includegraphics[width=0.45\linewidth]{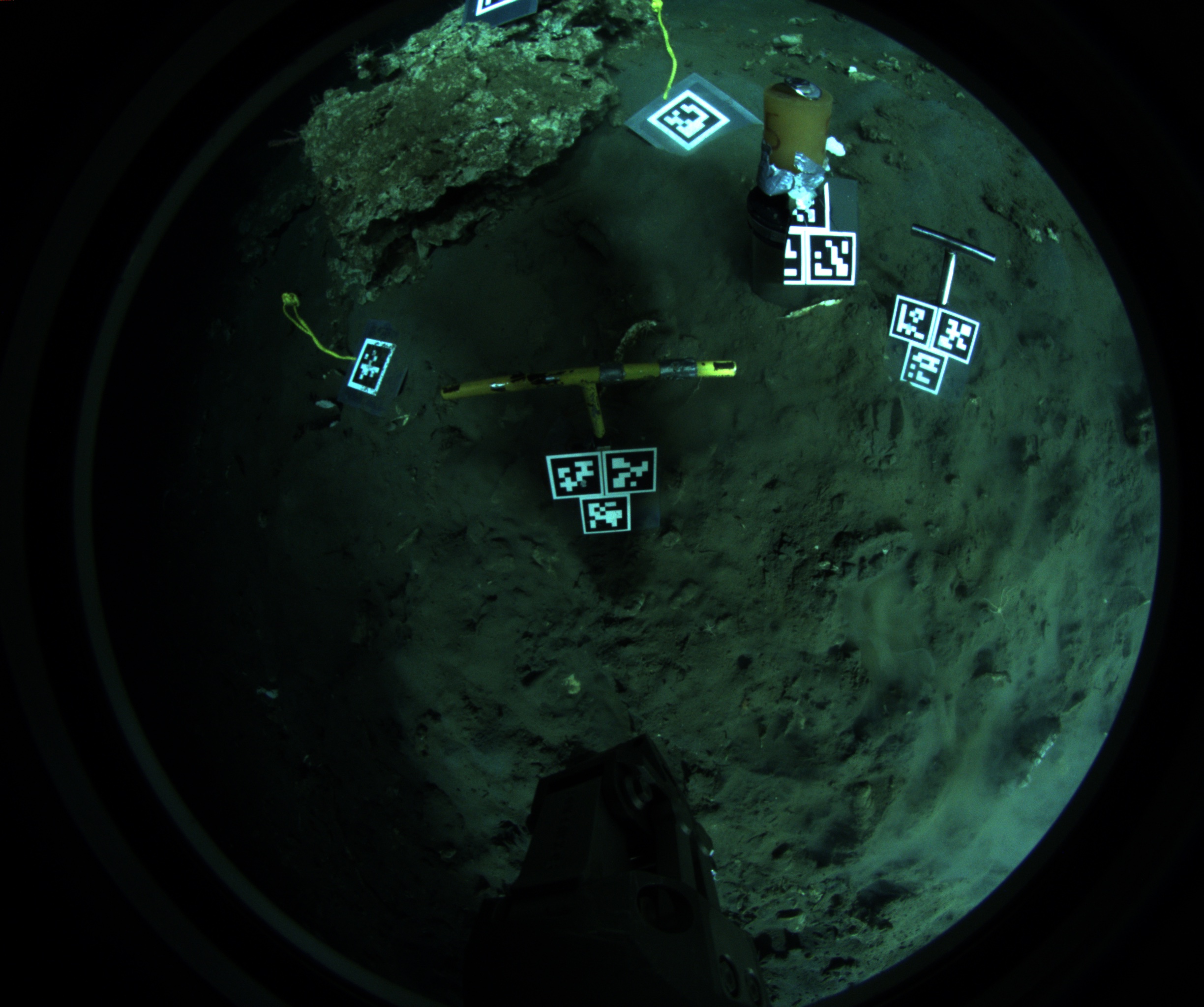}}\\
    \subfigure[Far perspective view]{\includegraphics[width=0.45\linewidth]{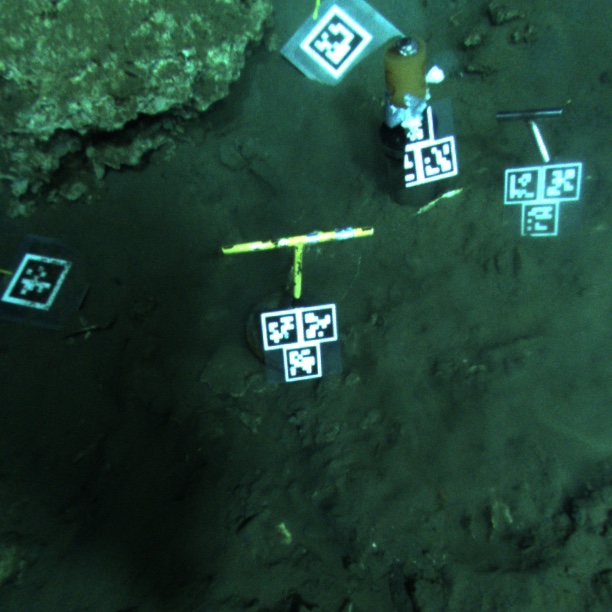}}\hfil
    \subfigure[Close perspective view]{\includegraphics[width=0.45\linewidth]{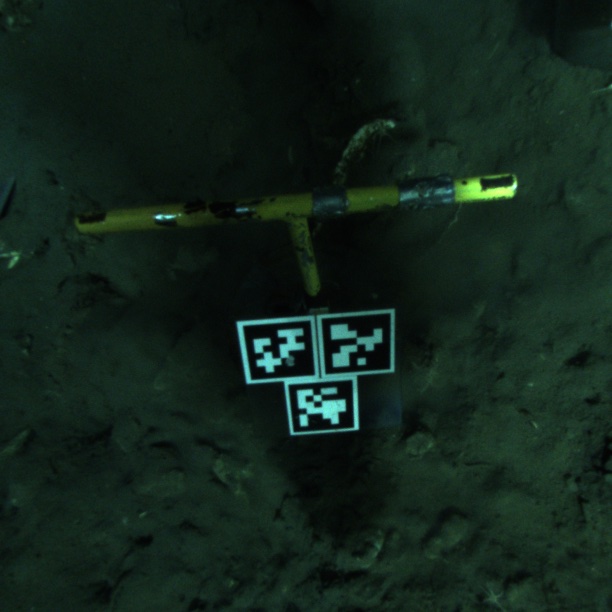}}\\
    \caption{A comparison of (top) the full view of the wrist-mounted fisheye camera in an underwater scene at close and far range compared to (bottom) a 60\degree perspective rectification, which illustrates the significant increase in the field-of-view provided by a fisheye lens compared to a conventional perspective lens. This increased field-of-view provides significantly better contextual awareness to the vision and manipulation systems, especially when working at close range to the target, which is typical for manipulation tasks.}
    \label{fig:fishvspersp}
\end{figure}

A vehicle capable of automated intervention must have an effective means to self-localize within the environment and visually reconstruct the workspace to complete the mission tasks. We adopt a vision system consisting of computer vision cameras, that takes into consideration three primary criteria. First, the system must be capable of generating a 3D reconstruction of the manipulator workspace, enabling the motion planner to avoid obstacles and generate safe, collision-free paths. Second, the system must be able to localize a set of known objects, such as tools, and guide the manipulator to grasp them. Third, the system must easily integrate with existing robotic platforms. Our vision system is composed of a vehicle chassis-mounted stereo camera pair with a fixed-baseline and a manipulator wrist-mounted fisheye camera (Fig.~\ref{fig:nht_smirc}). The stereo pair observes the manipulator workspace, including part of the tool tray and the scene working area. The system uses the stereo to generate 3D point clouds of the workspace for scene reconstruction, assist with localizing tools in the tool tray, and visually track dynamic vehicle reference frames that are otherwise not observable (e.g., the position of the \textit{NUI} HROV doors). The wrist-mounted fisheye camera provides a wide-angle view of the scene, and is used to detect objects and acquire dynamic viewpoints of the scene, which may be occluded or outside the field-of-view of the stereo pair. The wide field-of-view of the fisheye compared to a perspective camera enables clear views of objects and scene context at both close and far range (Fig.~\ref{fig:fishvspersp}), which is advantageous for manipulation.

All three cameras are Blackfly model BFLY-PGE-50S5C-C (FLIR, Wilsonville, OR). The stereo cameras use the VS Technology SV-0614H 6\,mm f/1.4 lens (VS Technology Corporation, Tokyo, JP), and the fisheye lens is the Fujinon FE185C086HA-1 2.7\,mm f/1.8 (Fujinon, Tokyo, JP). The camera housings are custom fabricated with titanium shells and dome viewports (Sexton Corporation, Salem, OR), with a depth rating of 6000\,m. A hardware trigger synchronizes the cameras. We calibrate the cameras using images of a checkerboard that the ROV manipulator moves throughout each camera's field-of-view while the vehicle is submerged. We calibrate the stereo cameras using the ROS stereo camera calibration package. We calibrate the fisheye camera using the Kalibr toolbox~\citep{kalibr2006}. Because the usable field-of-view for the fisheye camera is less than 180\degree ~due to occlusions from the housing, we use the pinhole projection model with equidistant distortion. We verify that both the stereo and fisheye calibrations achieve sub-pixel reprojection errors for the checkerboard corners.

\subsubsection{Tool Handle Pose Estimation}

\begin{figure}[!t]
    \centering
    \subfigure[AprilTag mount for tools]{\includegraphics[height=3in]{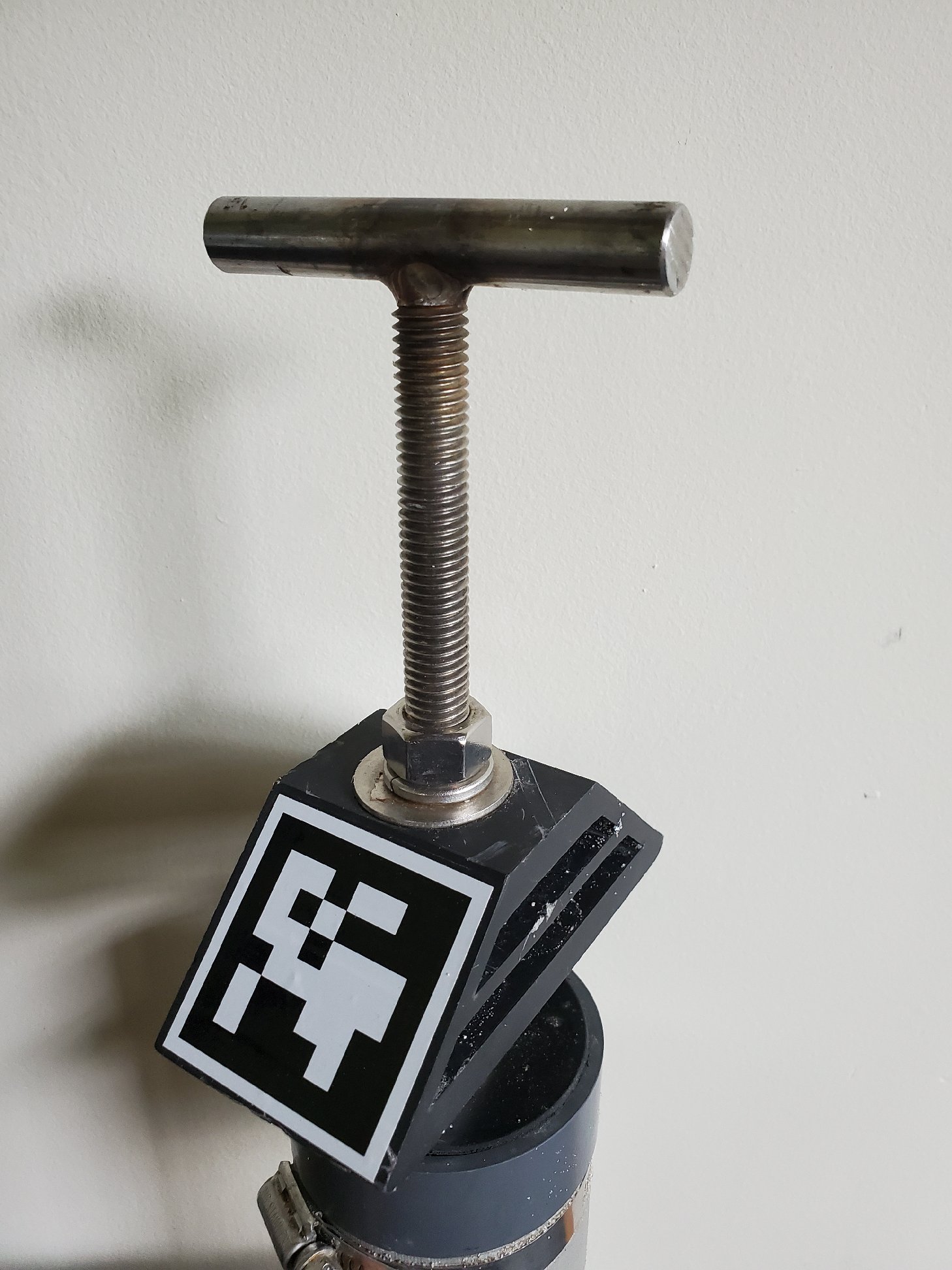}\label{fig:thandle-a}}\hfil
    \subfigure[Fisheye view of tools in tool tray]{\includegraphics[height=3in]{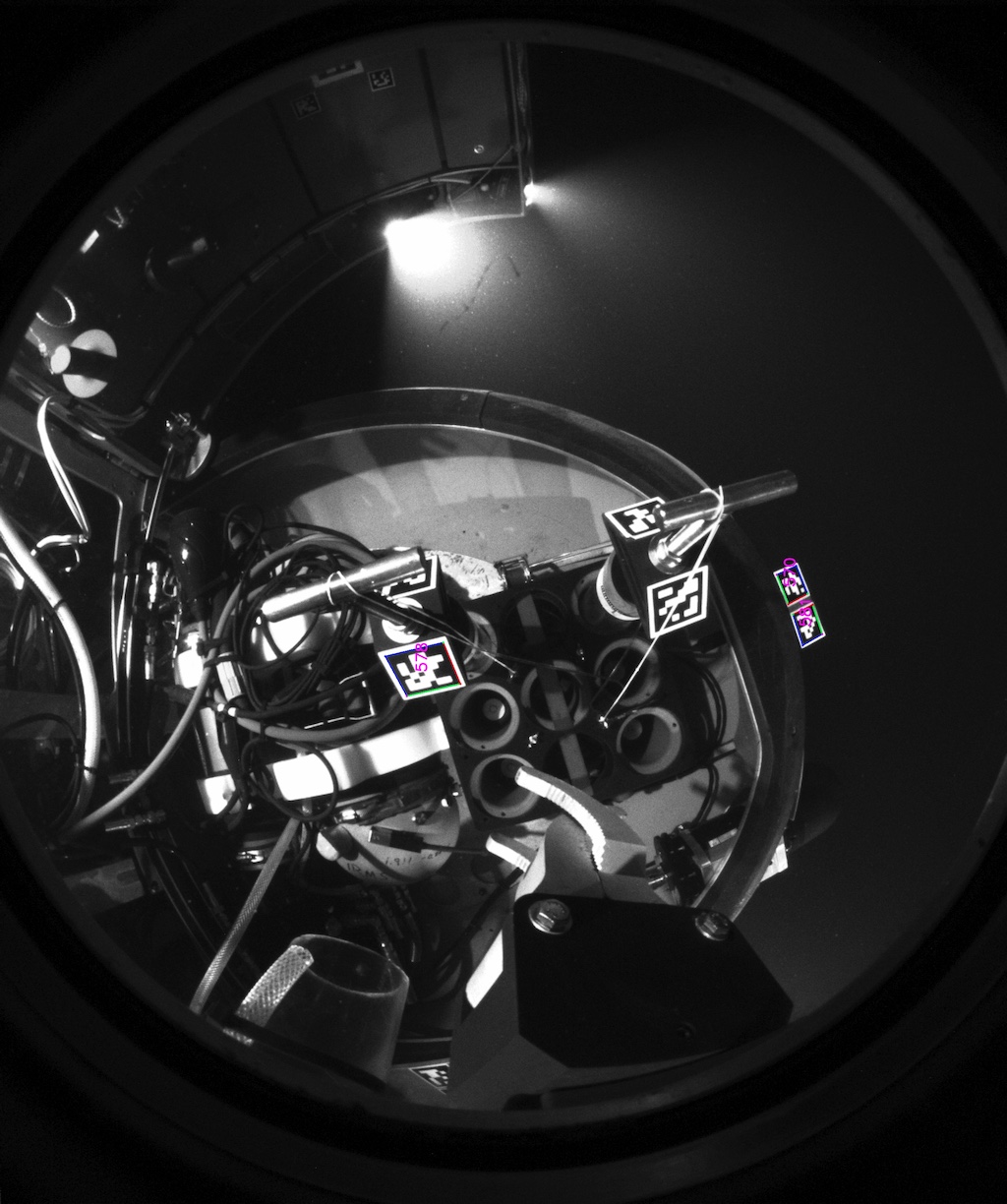}\label{fig:thandle-b}}
    \caption{A single type of t-handle was used to manipulate the different tools. The vision system localizes the t-handles using  \subref{fig:thandle-a} AprilTags affixed to 3D-printed mounts located beneath the t-handle. These tags are detected in \subref{fig:thandle-b} images of the ROV tool tray from the wrist-mounted fisheye camera.}
    \label{fig:thandle}
\end{figure}
Tools carried by the ROV must be localized by the vision system before they can be grasped. It is general practice in ROV operations to use a single type of handle on every tool to provide consistency for ROV pilots. Given a known type of tool and its model, the vision system need only localize the handle for a tool to be grasped and manipulated.

\revision{Using data collected with our vision system during the field trials, we developed a novel deep learning-based method, SilhoNet~\citep{billings2019silhonet} and SilhoNet-Fisheye~\citep{billings2020silhonet}, that estimates the pose of tool handles detected from the wrist-mounted fisheye camera, without the need for fiducials. SilhoNet uses an intermediate silhouette representation to regress the detected object poses. This silhouette representation improves pose regression performance and facilitates training the network on synthetic data, which is especially beneficial when real training data is limited, as is the case for underwater environments. This method achieves promising results on the recorded datasets, but was not ready for integration with the system during the field trials.}

\revision{During our field demonstrations, we relied on AprilTag markers~\citep{olson11} to localize the tool handles. Our choice of the AprilTag marker was motivated by the results of~\citet{apriltagunderwater}, which show that AprilTags yield the best performance in underwater environments, with the lowest sensitivity to turbidity and variable lighting conditions in comparison to other popular fiducial markers. In this study, the minimum marker size detectable in an image was approximately 20 pixels, which, for the $50 \times 50$\,mm markers used in our system, equates to an expected maximum detection range of approximately 1.0\,m for the fisheye camera and 2.4\,m for the stereo cameras. These distances are within the typical working ranges of the manipulators used in our demonstrations.} We designed 3D printed mounts that screw onto the t-handle bases and hold AprilTag vinyl stickers (Fig.~\ref{fig:thandle}, right).

\revision{We use the ROS TagSLAM package~\citep{pfrommer2019tagslam} to detect the fiducials from the wrist-mounted fisheye camera. TagSLAM is built on the GTSAM~\citep{dellaert2012factor} factor graph library and uses the ISAM2~\citep{kaess2012isam2} incremental optimizer for efficient run-time performance. TagSLAM operates in a transform tree completely separate from the world planning environment. Within the TagSLAM environment, the fisheye camera is set as the origin, while the tools with the tag mounts are set as dynamic objects. We optimize the pose of each detected tool with respect to the fisheye camera frame using TagSLAM. The optimized tool pose with respect to the fisheye frame is projected into the world frame through the manipulator kinematics. If the fisheye camera loses sight of a tool, the tool pose within the world scene remains static until the tool is tracked again with TagSLAM.}

\subsection{Control}

While many existing methods tightly couple vehicle and manipulator motion planning and control, our approach decouples the manipulator and imaging system from other systems on the ROV. This makes it easier to integrate the system with different ROVs and also minimizes risk to the vehicle, as the automation system runs independently of the vehicle's software stack. This approach also mimics standard ROV operation procedures, in which one pilot controls the vehicle while another pilot controls the manipulator. Our system seeks to replace the direct pilot control of the manipulator with a high-level automation interface that naturally integrates with standard ROV operational procedures. \revision{A current limitation of this control approach is a fixed-base assumption while the manipulator is activated. During manipulator operations, the ROV is assumed to be set down on the seabed and essentially acts as a fixed-base manipulator platform during a sampling tasks. When a manipulator command is executed, our system assumes that the scene state remains static until the activation is completed. This assumption of fixing the vehicle position before activating the manipulator follows the standard practice for operating work-class ROVs.}

\begin{figure}[!t]
    \centering
    \includegraphics[width=1.0\linewidth]{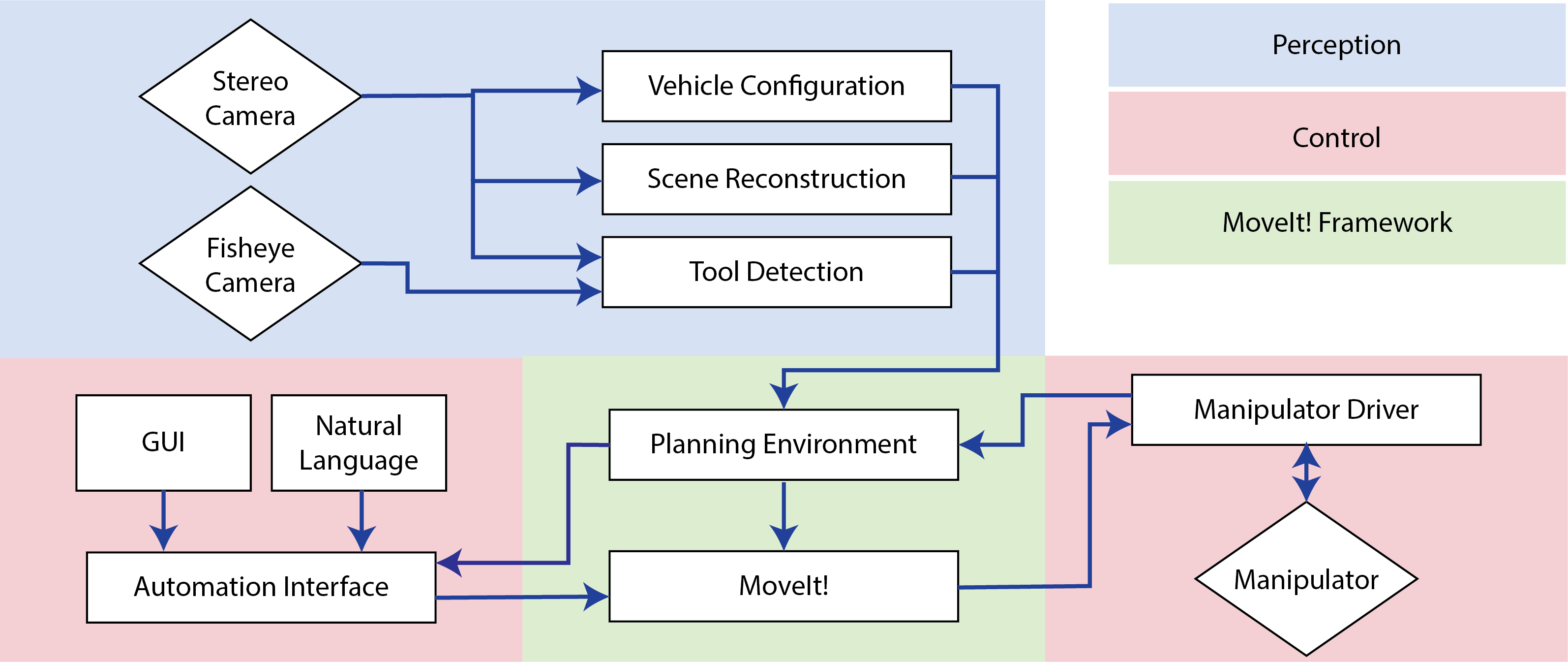}
    \caption{A diagram of the overall system, where rectangular blocks represent processes and diamond-shaped blocks represent hardware. Blocks in blue relate to perception. Blocks in red relate to (left) high- and (right) low-level control. Blocks in green are part of the MoveIt! framework around which our system is built. Our system uses the stereo camera to estimate the vehicle configuration (e.g., the pose of the doors on the \textit{NUI} HROV), generate point clouds of the scene that can be fused to produce a 3D reconstruction of the scene, and assist with tool localization. The fisheye camera is used to localize tools and obtain dynamic viewpoints of the workspace. For low-level control, a driver implements a position-based trajectory controller, which integrates between MoveIt! and the manipulator valve controller. For high-level control, we implemented an automation interface to MoveIt! that supports high-level commands. In this work, we implement this interface using a graphical front-end as well as a preliminary demonstration using natural language.}\label{fig:system}
\end{figure}

Figure~\ref{fig:system} shows a diagram of our system architecture. We use the MoveIt!~Motion Planning Framework~\citep{moveit2014} to integrate the outputs of the perception system into the planning scene and to generate collision-free motion plans. MoveIt!\ directly supports a diverse set of state-of-the-art motion planners and inverse kinematic (IK) solvers. For this work, we used the RRT$^*$ planner~\citep{karaman11a} with the KDL IK solver. We visualize the planning scene using RVIZ, with out-of-the-box integration with MoveIt!. We generated a kinematic description of the manipulator and vehicle from CAD models, and configured a motion planning environment with MoveIt!. A low-level driver for the manipulator exposes a position trajectory control interface to MoveIt!\ and interprets motion plans as command packets that it sends to the manipulator. Most work class hydraulic manipulators support only position setpoint commands. For this work, the system encodes the target joint positions and sends them directly to the manipulator valve controller.

\subsubsection{Calibration Procedure}

\begin{figure}[!t]
    \centering
    \includegraphics[width=0.8\linewidth]{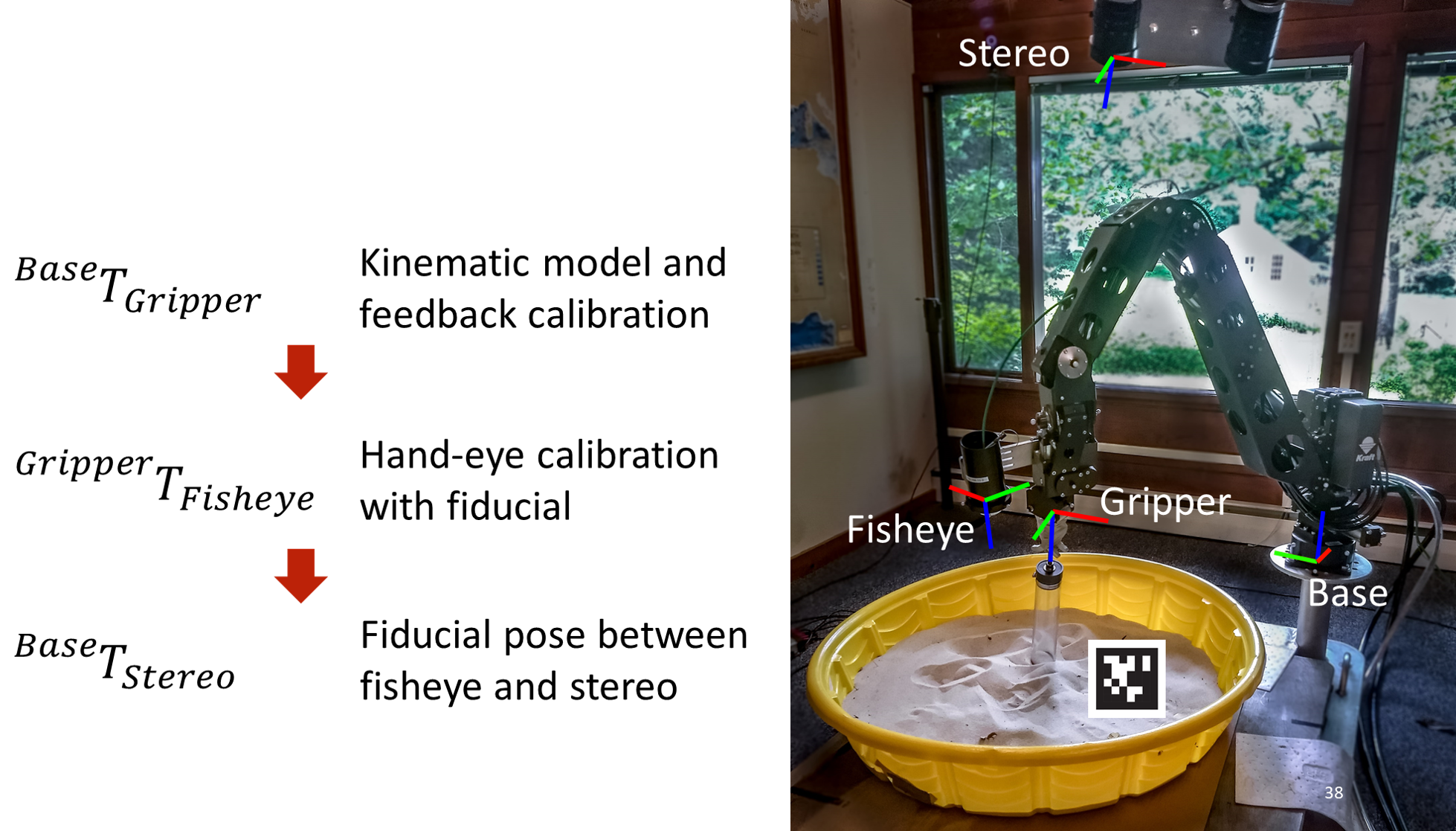}
    \caption{(right) An image of our testbed consisting of a Kraft TeleRobotics manipulator, a fisheye camera mounted to the end-effector, and a overhead stereo camera. Together with the manipulator base frame, there are four references frames (left) which must be calibrated in order to fuse sensor data into a common reference frame and to plan the motion of the arm. Calibration is performed in the order shown on the left, where each transform enables calibrating the next in a bootstrapping manner. The fiducial in the image is included to indicate that AprilTags were placed statically in the workspace to obtain the Gripper-to-Fisheye and Base-to-Stereo calibrations.}
    \label{fig:testbed_frames}
\end{figure}

Figure~\ref{fig:testbed_frames} illustrates the coordinate frame transforms that must be calibrated for motion planning and kinematic-based control of the manipulator. The end-effector pose follows from the kinematic chain of transforms from the manipulator base frame through each consecutive link, where each transform is parameterized by the joint angle. Hydraulic manipulators generally provide limited joint feedback from position sensors like potentiometers or resolvers, which must be calibrated to the kinematic model. For this work, we assume a linear interpolation between the feedback values at the joint limits.  We calibrate the hand-eye transform between the fisheye camera and the wrist link by detecting the fisheye-relative pose of an AprilTag positioned at a fixed location relative to the manipulator base. We perform these detections for a set of different kinematic configurations and then optimize the hand-eye transform using the ROS \verb+easy_handeye+ package~\citep{tsai1989new}. When the manipulator base is rigidly fixed relative to the stereo pair, as would be the case for most ROV configurations, the transformation between the stereo pair and the manipulator base frame is calibrated by detecting the pose of a vehicle-affixed AprilTag in the scene in both the left stereo and the wrist-mounted fisheye cameras. We transform the pose of the fiducial from the fisheye camera frame to the manipulator base frame through the kinematic chain, giving the stereo-to-base frame transform as the difference in the tag pose between the two frames. For the \textit{NUI} vehicle, where the stereo is not fixed relative to the manipulator, we used a different approach to estimate the stereo-to-base transform in real-time (see Section~\ref{sec:NUIdoor}).

\subsubsection{Pick-and-Place Interface}

\begin{figure}[!h]
    \centering
    \includegraphics[width=0.6\linewidth]{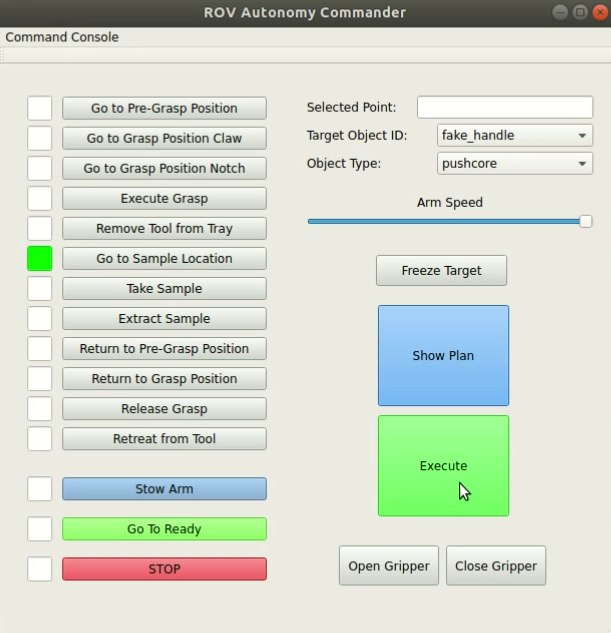}
    \caption{A simple interface to the automated system allows the user to configure and step through the automated pick-and-place pipeline. The motion plan for each step is visualized in the planning scene and is only executed upon confirmation by the user, which provides a high-level of safety for the system to be deployed on ocean-going systems.}
    \label{fig:interface}
\end{figure}
Our autonomy framework targets manipulation operations that involve pick-and-place tasks, such as taking a push-core sample. We implement a simple front-end interface (Fig.~\ref{fig:interface}) that allows a user to step through a pick-and-place state machine that automates each step of the process, while maintaining a high level of safety through human oversight. The interface visualizes the manipulator motion plan at each step and only proceeds to execute the plan after the operator provides confirmation. The interface allows the user to select a target among a set of tools detected in the scene and then activate a sequence of automated steps to grasp and manipulate the target using pre-defined grasp points. An interactive marker enables the user to indicate the desired sample location in the 3D planning scene. Besides the pick-and-place state machine controller, the interface enables one-click planning of the manipulator to a set of pre-defined poses, immediate stopping of any manipulator motion, and opening and closing of the gripper. The MoveIt!\ planning environment also allows the operator to command the manipulator to an arbitrary configuration within the workspace through an interactive 3D visualization.

\subsection{\revision{System Precision}}

\revision{The maximum precision of our system is limited by both kinematic and visual factors. The KRAFT manipulator uses 11 bit encoders, for an approximate per-joint angular resolution of 0.176\degree. When the arm is fully extended to 1.3\,m, the angular resolution for the shoulder joints equates to a metric arc length resolution of 4\,mm at the end-effector. However, this estimate does not account for non-linear effects in the hydraulic actuators, bias in the joint actuation, inaccurate feedback from the joint sensors, or flexing of the arm's mounting base/vehicle door, any of which may significantly reduce the kinematic accuracy of the system. The visual factors that limit precision include the accuracy of localizing the AprilTags from the fisheye camera and the resolution of the stereo reconstruction. Visual precision is dependent on the metric resolution of a pixel projected into the world. For a tag that is 1\,m from the fisheye camera, the pixel metric resolution is 1.3\,mm, which is the expected best precision for localizing the tags. High distortion of tags near the edges of the fisheye image is expected and has been observed to reduce the localization accuracy. When processing the stereo images to produce depth maps, the maximum working distance can be tuned based on the maximum disparity over which a feature match is searched across a rectified image pair. In our system, the maximum practical distance we target for stereo reconstruction is 3\,m, which is well beyond the manipulator reach and beyond which lighting and haze effects severely degraded the image quality. For a viewing range of 3\,m, the metric pixel resolution in the stereo view is 1.7\,mm. Due to feature smoothing by the SGM correlation window, the actual reconstructed spatial resolution is coarser. In practice, we have observed that the kinematic accuracy is the limiting factor on the precision of our system, due to the many sources of kinematic error in hydraulic manipulator systems.}

\section{Experiments and Field Results}
\label{section:results}

\subsection{Automated Pick-and-Place Demonstration on Testbed}

\begin{figure}[!t]
    \centering
    \subfigure[Detecting the t-handle]{\includegraphics[width=0.485\linewidth]{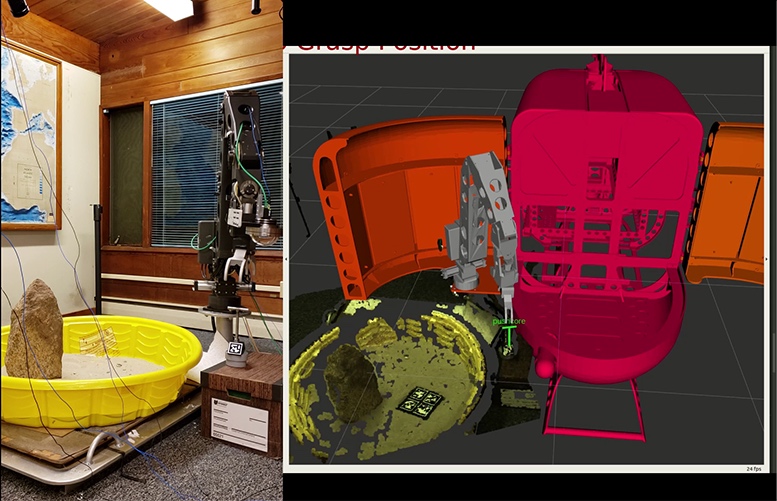}\label{fig:testbed_demo_a}} \hfill
    \subfigure[Grasping the t-handle]{\includegraphics[width=0.485\linewidth]{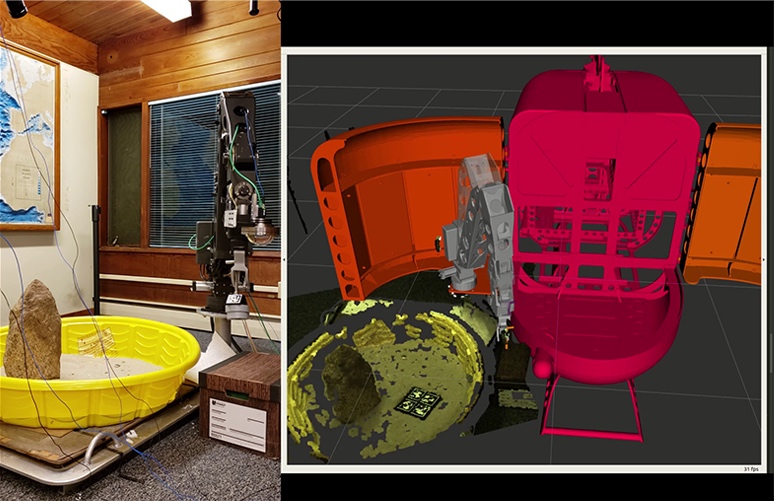}\label{fig:testbed_demo_b}}\\
    \subfigure[Moving to the sample location]{\includegraphics[width=0.485\linewidth]{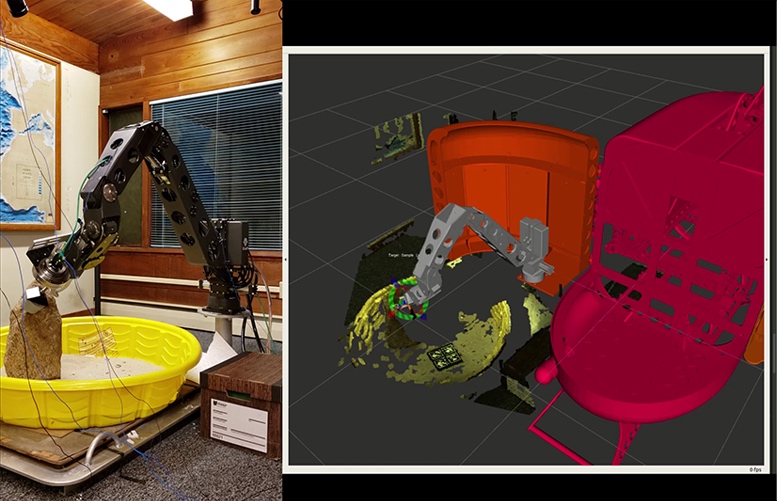}\label{fig:testbed_demo_c}} \hfill
    \subfigure[Replacing the t-handle]{\includegraphics[width=0.485\linewidth]{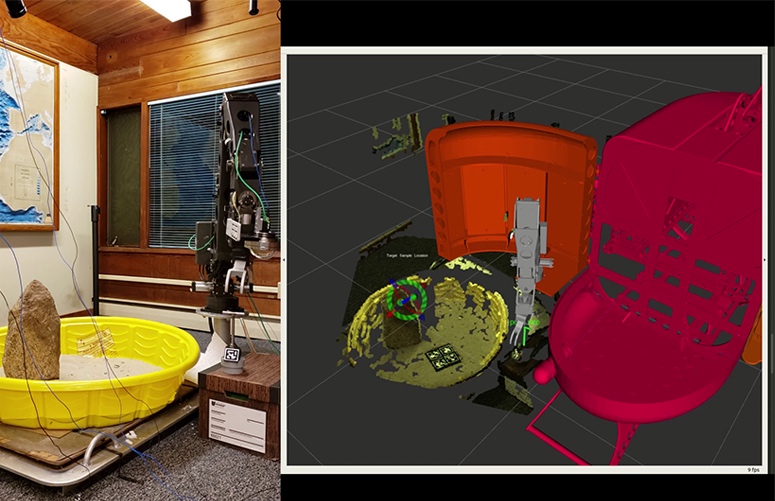}\label{fig:testbed_demo_d}}
    \caption{We demonstrated fully autonomous pick-and-place with a t-handle on a testbed with the same camera and manipulator hardware used on the \textit{NUI} HROV. First, \subref{fig:testbed_demo_a} the t-handle was detected from the fisheye camera using the AprilTags, and the handle pose was projected into the planning scene. Next, \subref{fig:testbed_demo_b} the manipulator was commanded to grasp the t-handle via the autonomy interface. Subsequently, \subref{fig:testbed_demo_c} a sample location was set in the planning scene with an interactive marker based on the projected stereo point cloud, the manipulator planned a motion to reach the sample location, and executed the plan after the user verified it. The manipulator was then \subref{fig:testbed_demo_d} commanded to return the t-handle to the location where it was first grasped. The rock in the environment was placed in a delicate balance on its end, yet the manipulator was controlled with enough precision to bring the tool into direct contact without knocking it over.}
    \label{fig:testbed_demo}
\end{figure}

To prove the viability of our system before deploying it in the field, we demonstrated the full pick-and-place pipeline on a hardware testbed (Fig.~\ref{fig:testbed_demo}) that mimics the configuration of the vision system and manipulator as they would be mounted on an ROV. The testbed includes a Kraft TeleRobotics manipulator identical to the one that we use for the field deployments with the \textit{NUI} vehicle. The planning environment simulates the manipulator being mounted on the \textit{NUI} HROV. The stereo point cloud is projected into the planning scene to inform placement of the sample marker. As described previously, we estimate the t-handle pose from the wrist-mounted fisheye camera by detecting the AprilTags mounted below the handle. We executed each step of the automated pick-and-place interface successfully, with no manual control input. The system grasped the t-handle  based on the detected pose, and the planner found and executed a manipulation path to the sample location marker, which was placed at a non-trivial angle, touching a rock in the scene. The rock was placed on its end in a delicately balanced position, and the manipulator was able to bring the tool into contact with the rock with enough precision that the rock remained standing. The tool was then returned to the position from where it was grasped. This full experiment was repeated multiple times, though not without some grasp failures, due to noise in the visually estimated pose of the t-handle.  However, the interface made it easy to recover from any failed step of the state-machine without ever requiring the operator take manual control of the manipulator.

\subsection{Real-Time Scene Reconstruction and Data Collection at the Costa Rican Pacific Shelf Margin}

\begin{figure}[!t]
    \centering
    \includegraphics[width=0.8\linewidth]{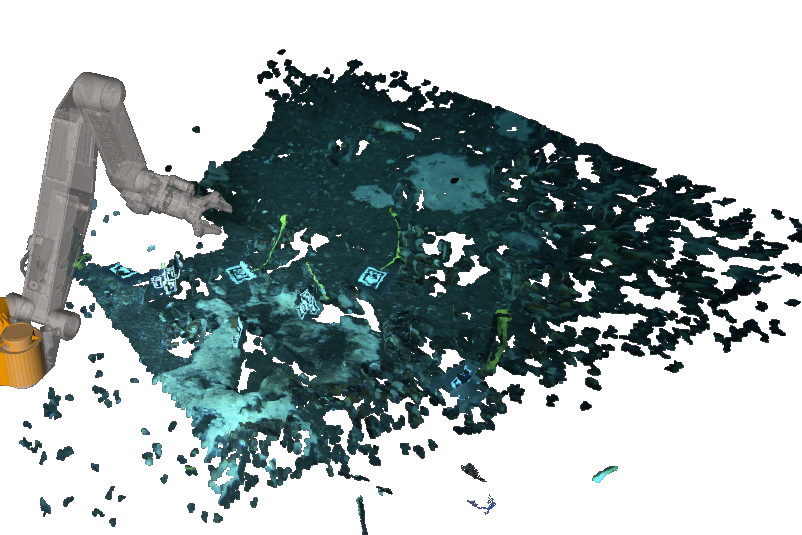}
    \caption{The vision system was integrated on the \textit{SuBastian} ROV operated by Schmidt Ocean Institute, where we demonstrated real-time visualization of the planning scene with a Schilling Titan-4 manipulator and the projected stereo point clouds. This also demonstrates the flexibility of the system to be integrated with different vehicles and manipulators.}
    \label{fig:titan_reconstruction}
\end{figure}

Demonstrations at the Pacific continental margin were conducted during a two-week research cruise aboard the R/V \textit{Falkor} using the \textit{SuBastian} ROV. The automated manipulation component of this expedition focused on a demonstration of the vision system and data collection to aid the development of visual methods. The integration time of our system took approximately two days during cruise mobilization, highlighting the relative ease and flexibility with which the system can be implemented on a variety vehicles and manipulators. Camera image data was streamed over a GigE interface at 3\,Hz to a topside workstation, which handled all processing and visualization. Joint encoder feedback from the manipulator was obtained by passively monitoring the serial communication between the manipulator and the ship's control computer. We visualized the real-time configuration of the manipulator in the 3D planning environment with the stereo point clouds projected into the scene. The point clouds were generated from the stereo imagery using the standard semi-global matching (SGM) method built into the ROS image processing pipeline, and the parameters were hand-tuned to achieve the best results. Figure~\ref{fig:titan_reconstruction} shows a frame from the real-time visualization captured on the seafloor during one of the dives. A good camera calibration combined with high water clarity, rich seafloor texture, and evenly distributed scene lighting resulted in high quality point clouds. These early results demonstrated the effectiveness of the vision system to capture the 3D structure of the workspace and the ability to fuse the information into a real-time scene representation that is useful for both manipulation planning and 3D visualization of the ROV configuration and planning environment.

\begin{figure}[!t]
    \centering
    \includegraphics[width=1.0\linewidth]{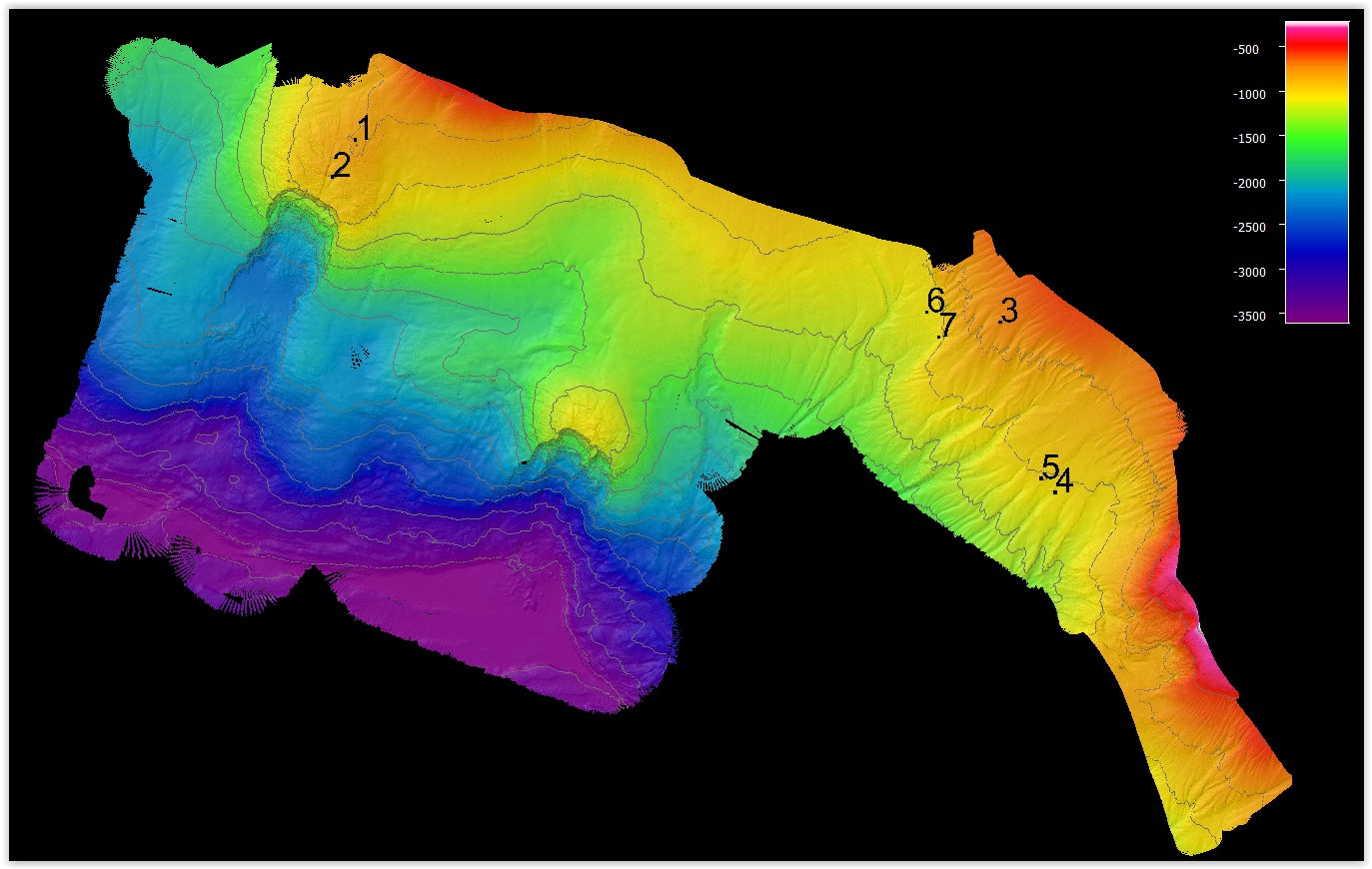}
    \caption{Bathymetric map of the survey area from the 2018 cruise on the Pacific continental margin showing data collection locations at seven different science goal sites, spanning over 62\,km (linear distance between Locations 1 and 4) and ranging in depth from 600\,m to 1100\,m. Depth contours are spaced at 250\,m intervals and the map is oriented with North up.}
    \label{fig:CostaRicaMap}
\end{figure}

During this expedition we collected an extensive dataset~\citep{billings2020silhonet} of synchronized stereo and wrist mounted fisheye images along with the manipulator joint feedback from a diverse set of seafloor environments (Fig.~\ref{fig:CostaRicaMap}). AprilTags mounted on plates were dispersed into the scenes to provide ground truth for the camera poses, and three different types of graspable handle objects were also randomly placed into the scenes. This dataset supported the development of our visual methods and is also intended to serve the underwater research community for the development of scene reconstruction, object detection, and pose estimation methods that work robustly in real seafloor environments.

\begin{figure}[!h]
    \centering
    \subfigure[Sample raw fisheye images from different sequences of the dataset]{\includegraphics[width=1.0\linewidth]{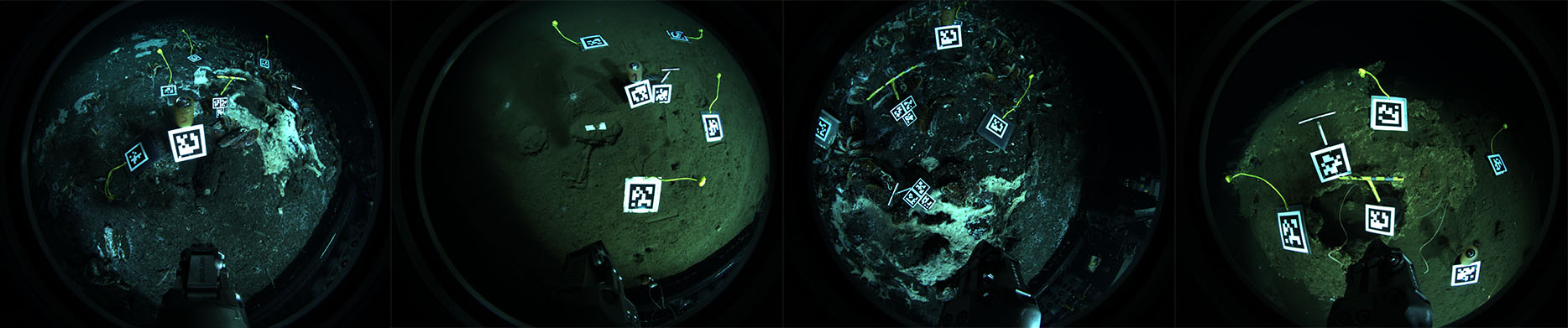}\label{fig:uwhandles_a}}\\
    \subfigure[Sample annotations from a single sequence of the dataset]{\includegraphics[width=1.0\linewidth]{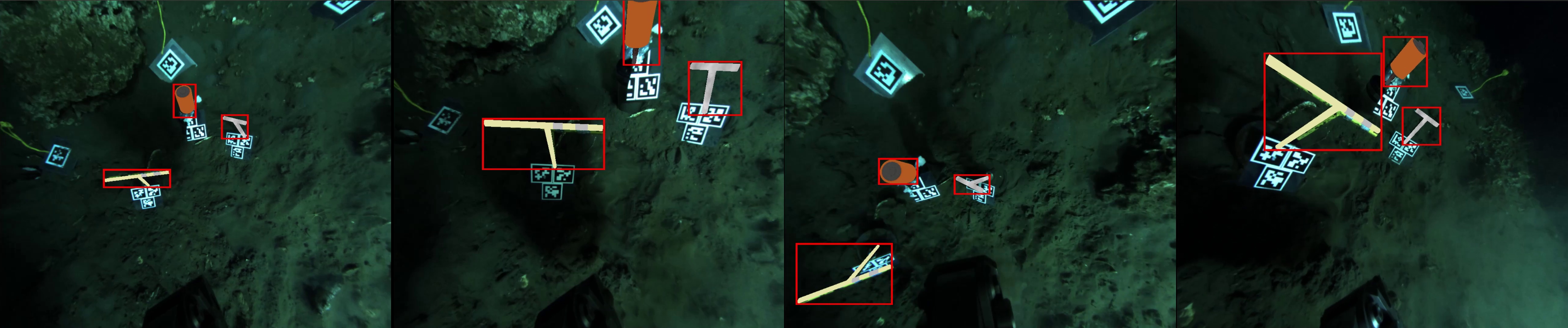}\label{fig:uwhandles_b}}
\caption{The fisheye imagery collected during the Costa Rica cruise was processed into a stand-alone dataset~\citep{billings2020silhonet}. The images are annotated with the bounding box and six-DoF pose of the tool handles placed in the workspace. The top row~\subref{fig:uwhandles_a} shows sample raw fisheye images from different sequences of the dataset, and the bottom row~\subref{fig:uwhandles_b} shows sample annotations from a single sequence in the dataset. The images are center rectified here only for purposes of visualization.}
    \label{fig:uwhandles}
\end{figure}

The fisheye images were processed into a standalone dataset with annotated 2D bounding boxes and 6D poses for the handle objects visible in each frame. This dataset was released with the SilhoNet-Fisheye publication~\citep{billings2020silhonet}. Figure~\ref{fig:uwhandles} shows sample images from this dataset. The combined dataset of stereo and fisheye imagery with synchronized manipulator joint feedback supported our development of visual methods for scene reconstruction.

\subsection{Automated Sample Collection and Return within Active Submarine Volcanoes}
\label{sec:NUIdoor}

\begin{figure}[!t]
    \centering
    \includegraphics[width=\linewidth]{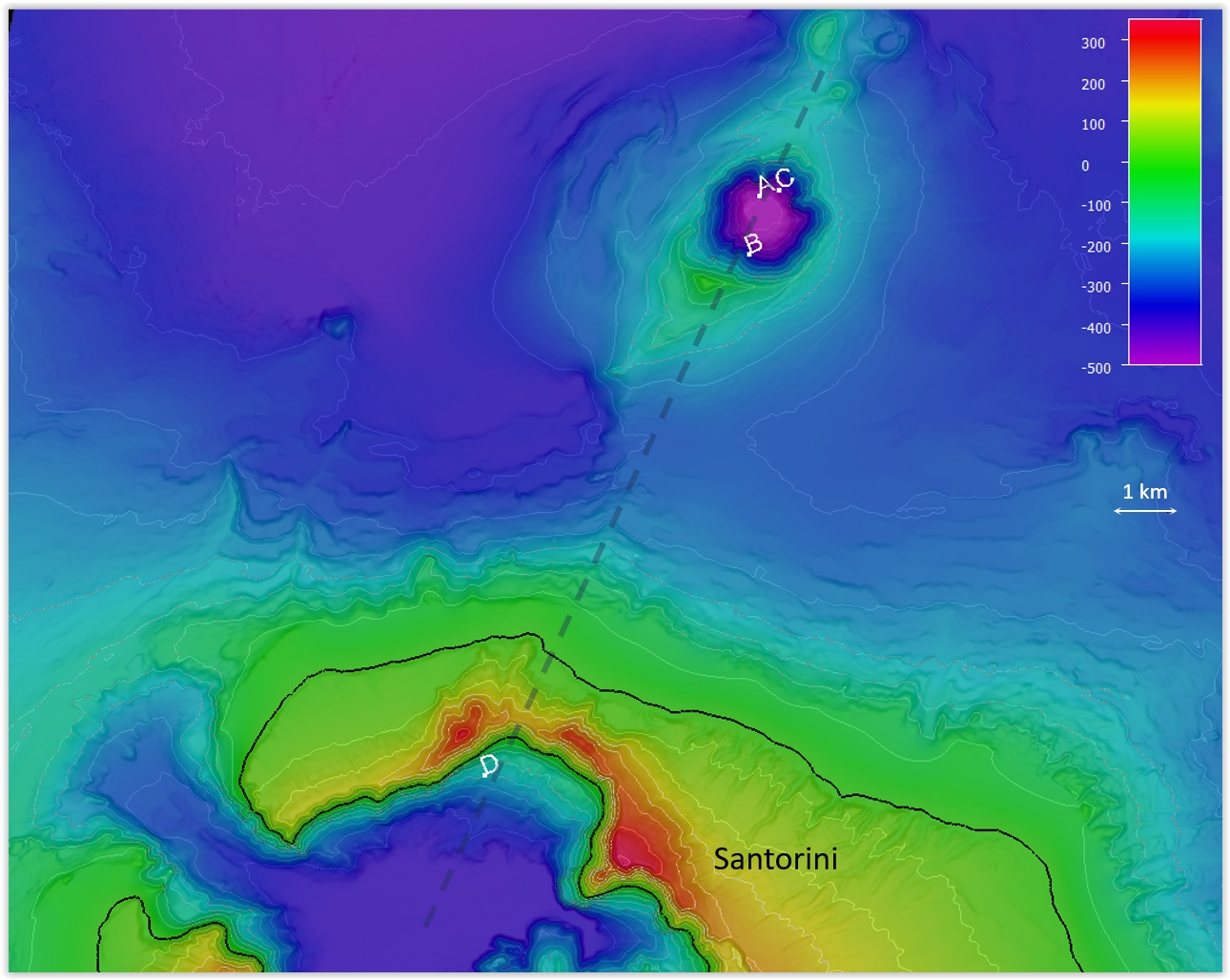}
    \caption{Map of automated sample collection locations, with regional bathymetry adapted from \citet{nomikou2012submarine,  nomikou2013morphological}. The sea level contour is indicated in black. The dashed line indicates the Christiana-Santorini-Kolumbo tectonic line~\citep{nomikou2012submarine}.  Locations marked A, B, and D indicate automated sample collection and return sites, and location C indicates the site where a natural language proof-of-concept demonstration was conducted. Sampling depths ranged from 240\,m to 501\,m }
    \label{fig:kolumbo}
\end{figure}
\begin{figure}[!h]
    \centering
    \includegraphics[width=1.0\linewidth]{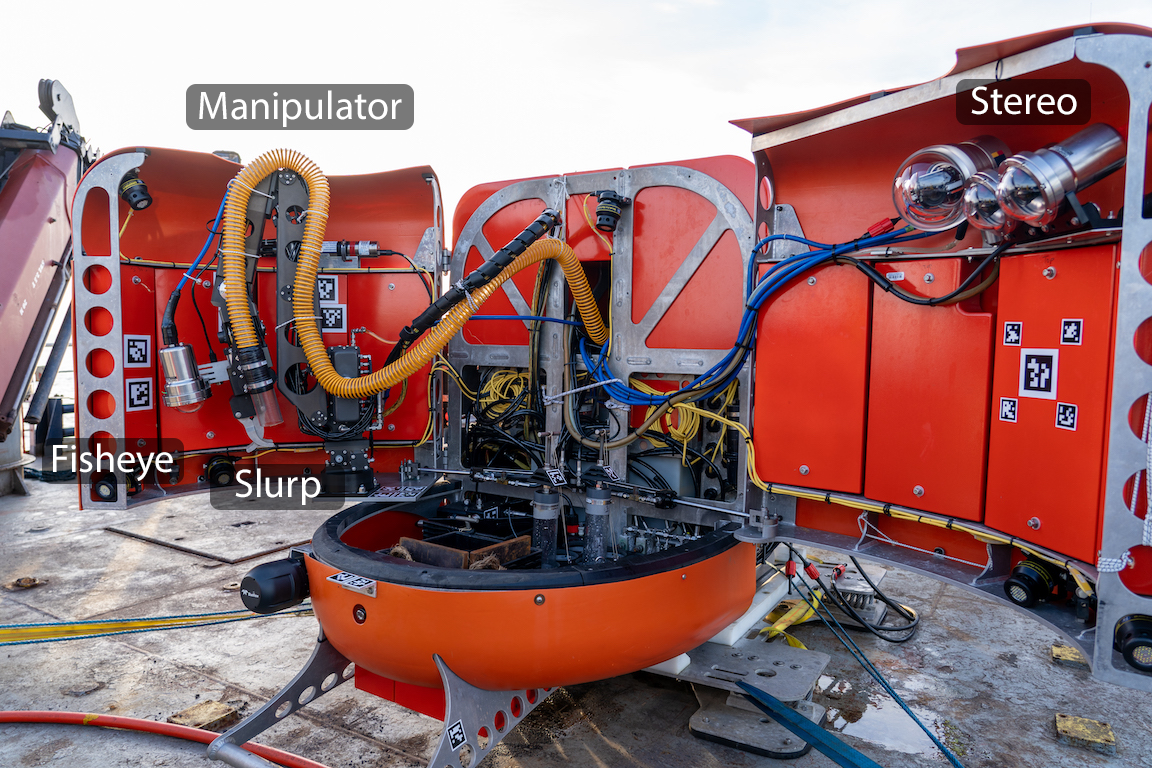}
    \caption{The \textit{NUI} vehicle is outfitted with clam shell doors that can be closed to reduce drag when cruising and opened to perform manipulation tasks. The manipulator is mounted to the starboard door and the stereo cameras are mounted to the port door.}
    \label{fig:nui}
\end{figure}

For exploration of the Kolumbo and Santorini calderas, \textit{NUI}'s manipulator was mounted to the starboard door and the stereo cameras were mounted to the port door (Fig.~\ref{fig:nui}). Having the manipulator and stereo camera on opposite articulating doors allowed for flexibility in configuring the position of the arm according to the specific manipulation task and enabled on-the-fly adjustment of the manipulator and stereo positions separately. Unfortunately, the doors are actuated using hydraulic rams that lack position feedback. For safe motion planning, it was necessary to estimate the door positions in real-time. The estimated door positions were used to update the kinematic configuration of the vehicle in the planning scene. However, we observed that the doors could flex, introducing some error in the kinematic estimates that negatively impacted the accuracy of the stereo point cloud projection into the planning scene. To minimize accumulated error in the transform between the stereo camera frame and the manipulator base frame, the stereo frame was referenced directly to the base frame in the ROS transform tree. The base frame was accurately localised directly from the stereo camera through detection of tags fixed to the manipulator base.

\begin{figure}[!h]
    \centering
    \includegraphics[height=1.8in]{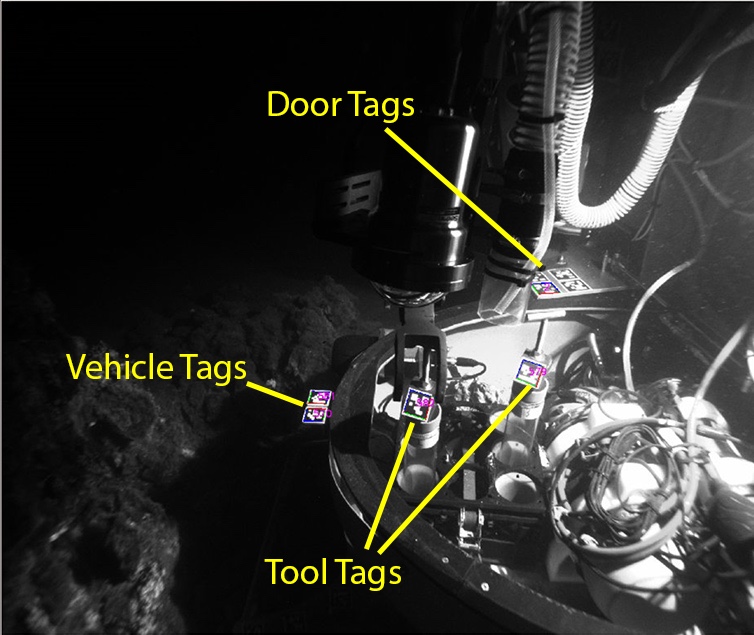}\hfill
    \includegraphics[height=1.8in]{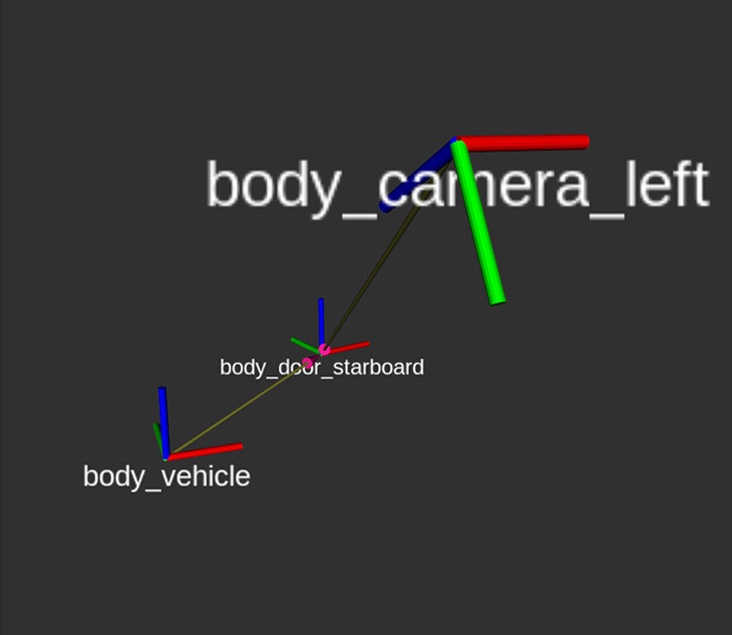}\hfill
    \includegraphics[height=1.8in]{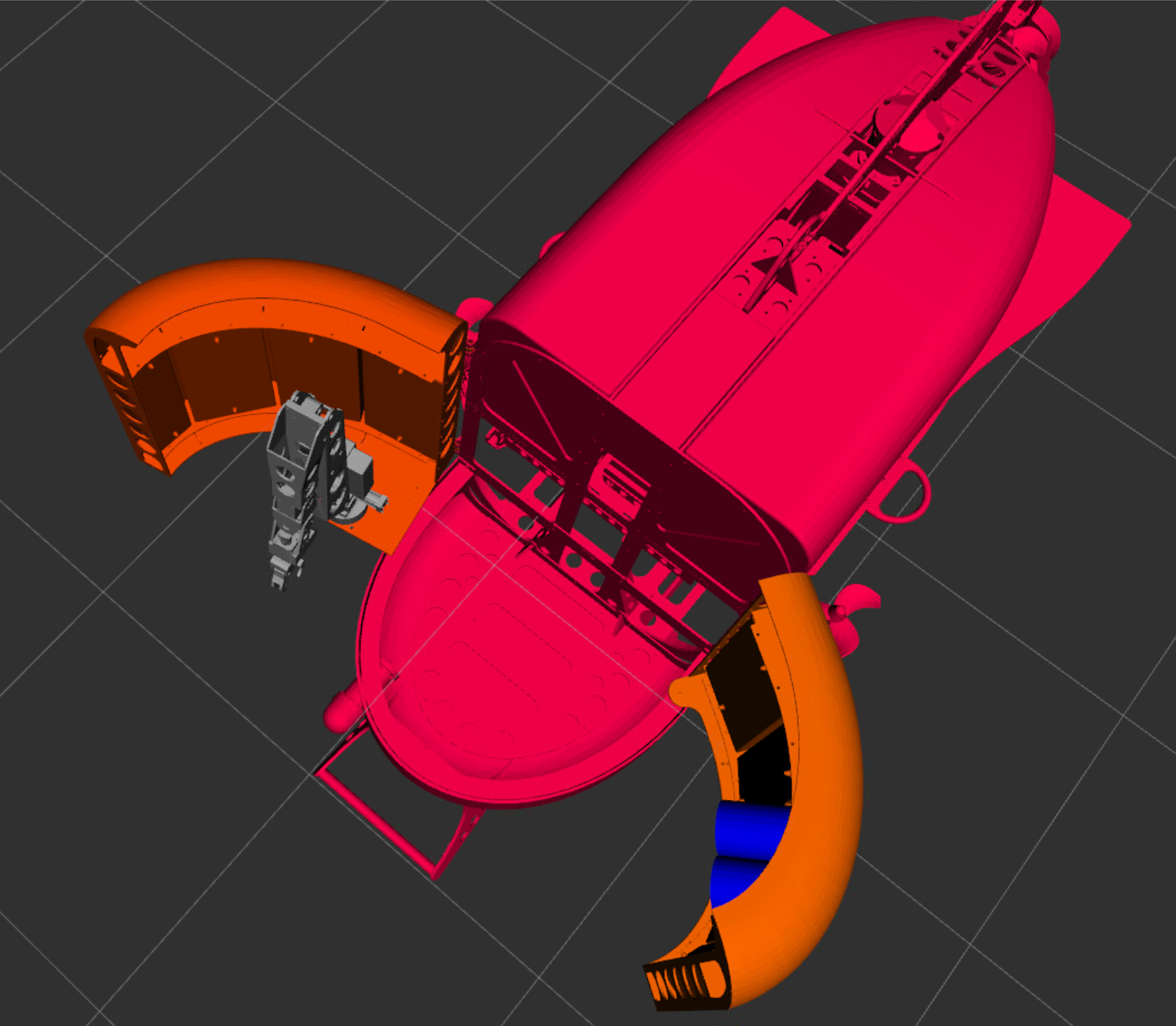}
    \caption{Fiducial-based visual SLAM from the left stereo camera was used to estimate the door angles in real-time using (left) tags mounted to the front of the vehicle frame and at the base of the manipulator on the starboard door. SLAM provided estimates of (middle) the relative transformations between the camera and the tag frames that were used (right) to estimate the door angles and update the vehicle model in the planning scene. The left stereo camera was also used in conjunction with the wrist mounted fisheye for (left) fiducial-based localization of tools.}
    \label{fig:doorslam}
\end{figure}

\begin{figure}[!h]
    \centering
    \includegraphics[width=0.6\linewidth]{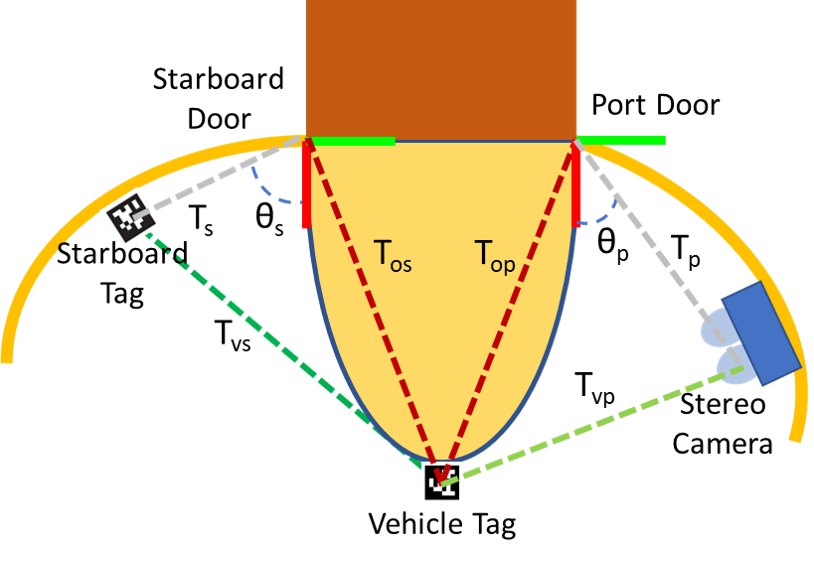}
    \caption{A 2D schematic of the \textit{NUI} HROV that relates the visual SLAM from the left stereo camera to the door positions. The green dashed lines represent transformations estimated from SLAM. The red dashed lines denote known transformations computed from the vehicle kinematic model and the measured position of the tags. The grey dashed lines represent the calculated transforms with respect to each door reference frame, which have a trigonometric relation to the door angles, $\theta_s$ and $\theta_p$.}
    \label{fig:doorjoints}
\end{figure}

To estimate the door positions in real-time, we affixed AprilTags to the starboard door and to the bow of the vehicle's payload bay (Fig.~\ref{fig:doorslam} (left)). The tags on the vehicle bow were mounted at a measured location relative to the vehicle reference frame, with the reference tag's $Z$-axis aligned with the $Z$-axis of the vehicle reference frame. The door joint axes of rotation were also aligned to the $Z$-axis of the vehicle frame, enabling  a simple trigonometric calculation of the door angles based on the relative tag locations in the $X$-$Y$ plane. We used the left stereo camera to track the relative pose of the AprilTags and used these estimates as observations in AprilTag-based visual SLAM~\citep{pfrommer2019tagslam} (Fig.~\ref{fig:doorslam}). Figure~\ref{fig:doorjoints} shows a schematic of the vehicle and visual SLAM system with the relevant transforms in the $X$-$Y$ plane used to calculate the door angles. The visual SLAM provided the relative translations between the vehicle tag frame and the starboard tag frame, $T_{vs}$, and between the vehicle tag frame and the stereo camera frame, $T_{vp}$. Given that the translations between the vehicle tag and the door joint frames, $T_{os}$ and $T_{op}$, were measured and known, the angle of the starboard and port doors, $\theta_s$ and $\theta_p$ respectively, were recovered as
\begin{subequations}
    \begin{align}
     \theta_s &= \arctan{\frac{T_{s,y}}{T_{s,x}}} - \theta_{s_0}\\
     \theta_p &= \arctan{\frac{T_{p,y}}{T_{p,x}}} - \theta_{p_0},
 \end{align}
\end{subequations}
where
\begin{subequations}
    \begin{align}
     T_s &= T_{vs}-T_{os}\\
     T_p &= T_{vp}-T_{op},
 \end{align}
\end{subequations}
where the $x$ and $y$ subscripts indicate the corresponding component of the translation vector, and $\theta_{s_0}$ and $\theta_{p_0}$ are the measured angle offsets.

\subsubsection{Planner Controlled Biological Sample Collection}
\begin{figure}[!h]
    \centering
    \subfigure[Taking slurp sample]{\includegraphics[width=0.8\linewidth]{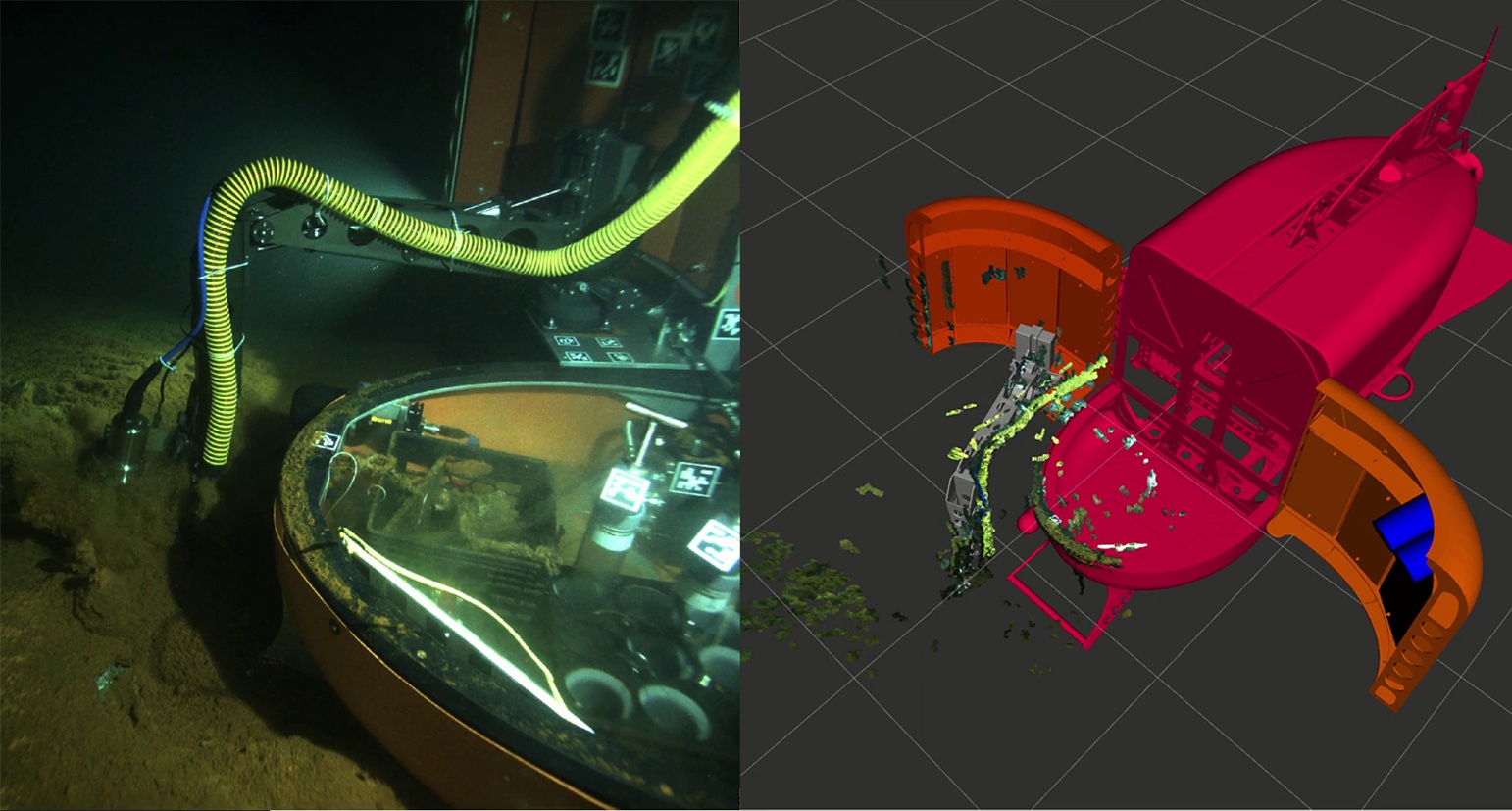}\label{fig:slurp_a}}\\
    \subfigure[Returning to home position]{\includegraphics[width=0.8\linewidth]{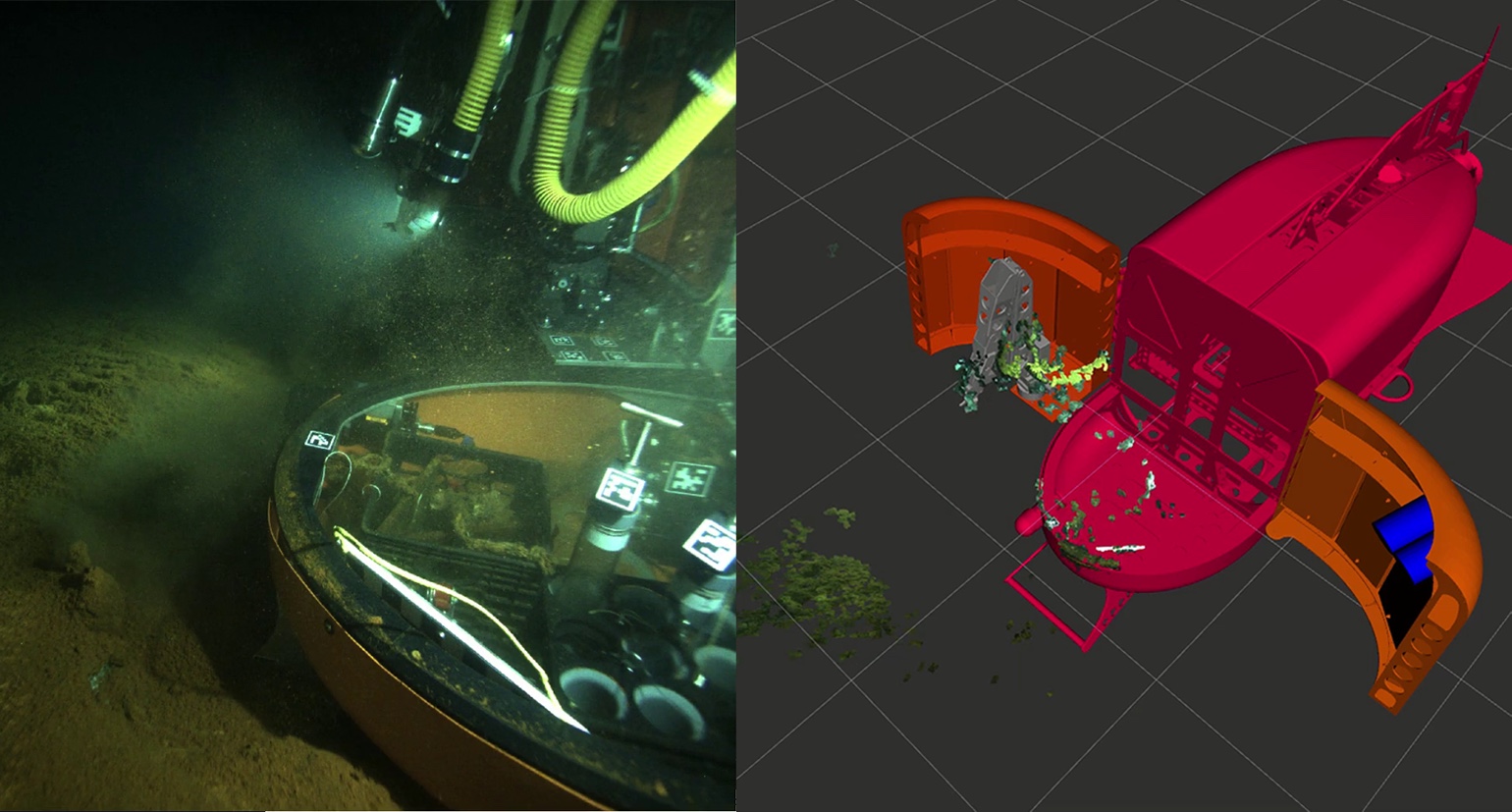}\label{fig:slurp_b}}
    \caption{An example of a successful planner controlled slurp collection of a bacterial mat, with the yellow slurp hose attached to the manipulator. The manipulator was \subref{fig:slurp_a} commanded to the desired slurp location through the automated planning interface and then \subref{fig:slurp_b} directed to return to its home position following the slurp collection.}
    \label{fig:slurp}
\end{figure}

The Kolumbo-Santorini expedition resulted in several scientific achievements, including verification of the persistence of Kalliste Limnes~\citep{camilli2015kallisti}, 3D reconstruction of extremophile habitats within the calderas' craters, and sampling of benthic fluids, seafloor sediments, and biological materials.  One of the most useful subsea tools for sample collection and return is a "slurp gun" vacuum sampler. For these operations, the slurp nozzle is in close proximity to the sample of interest and a vacuum pump sucks the sample through the hose into a collection chamber. To demonstrate automated slurp collection, we attached the slurp hose to the side of the manipulator, so that the end-effector could be commanded to the desired location to collect the slurp sample. We completed multiple successful sample collections, including that shown in Figure~\ref{fig:slurp}, where a slurp sample of a sediment microbial mat was collected using the planner interface to command the manipulator to the desired sample location, after which the manipulator was returned to the home position.

\subsubsection{Natural Language Control}

Subsea ROV missions require close collaboration between the ROV pilots and scientists. The primary means by which pilots and scientists communicate is through spoken language---scientists use natural language to convey specific mission objectives to ROV pilots (e.g., requesting that a sample be taken from a particular location), while the pilots engage in dialogue to coordinate their efforts. Natural language provides a flexible, efficient, and intuitive means for people to interact with our automated manipulation framework. The inclusion of a natural language interface would support our goal to realize a framework that can be integrated seamlessly with standard ROV operating practices and may eventually mitigate the need for a second pilot.

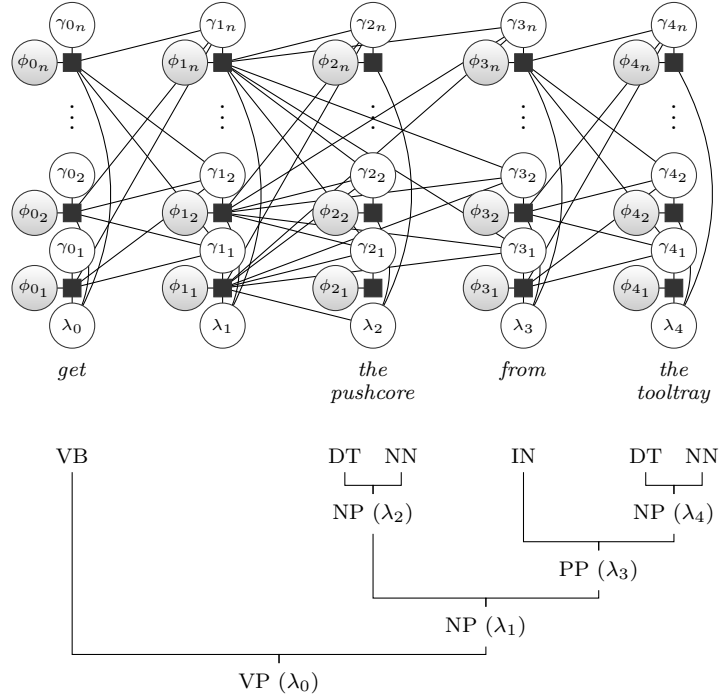
\begin{figure}[!h]
\centering
\begin{tikzpicture}[textnode/.style={anchor=mid,font=\tiny},nodeknown/.style={circle,draw=black!80,fill=white,minimum size=6mm,font=\tiny},nodeunknown/.style={circle,draw=black!80,fill=black!10,minimum size=6mm,font=\tiny,top color=white,bottom color=black!20},factor/.style={rectangle,draw=black!80,fill=black!80,minimum size=2mm,font=\tiny,text=white}]
\draw[-] (0,0.5) to (0,1);
\draw[-] (0,0.5) to [bend right=30] (0,2);
\draw[-] (0,0.5) to [bend right=30] (0,4);
\draw[-] (-0.5,1) to (0,1);
\draw[-] (-0.5,2) to (0,2);
\draw[-] (-0.5,4) to (0,4);
\draw[-] (0,1.5) to (0,1);
\draw[-] (0,2.5) to (0,2);
\draw[-] (0,4.5) to (0,4);
\draw[-] (2.0,1.5) to (0,1);
\draw[-] (2.0,2.5) to (0,1);
\draw[-] (2.0,4.5) to (0,1);
\draw[-] (2.0,1.5) to (0,2);
\draw[-] (2.0,2.5) to (0,2);
\draw[-] (2.0,4.5) to (0,2);
\draw[-] (2.0,1.5) to (0,4);
\draw[-] (2.0,2.5) to (0,4);
\draw[-] (2.0,4.5) to (0,4);
\draw[-] (4.0,1.5) to (2.0,1);
\draw[-] (4.0,2.5) to (2.0,1);
\draw[-] (4.0,4.5) to (2.0,1);
\draw[-] (4.0,1.5) to (2.0,2);
\draw[-] (4.0,2.5) to (2.0,2);
\draw[-] (4.0,4.5) to (2.0,2);
\draw[-] (4.0,1.5) to (2.0,4);
\draw[-] (4.0,2.5) to (2.0,4);
\draw[-] (4.0,4.5) to (2.0,4);
\draw[-] (2.0,0.5) to (2.0,1);
\draw[-] (2.0,0.5) to [bend right=30] (2.0,2);
\draw[-] (2.0,0.5) to [bend right=30] (2.0,4);
\draw[-] (1.5,1) to (2.0,1);
\draw[-] (1.5,2) to (2.0,2);
\draw[-] (1.5,4) to (2.0,4);
\draw[-] (2.0,1.5) to (2.0,1);
\draw[-] (2.0,2.5) to (2.0,2);
\draw[-] (2.0,4.5) to (2.0,4);
\draw[-] (6.0,1.5) to (2.0,1);
\draw[-] (6.0,2.5) to (2.0,1);
\draw[-] (6.0,4.5) to (2.0,1);
\draw[-] (6.0,1.5) to (2.0,2);
\draw[-] (6.0,2.5) to (2.0,2);
\draw[-] (6.0,4.5) to (2.0,2);
\draw[-] (6.0,1.5) to (2.0,4);
\draw[-] (6.0,2.5) to (2.0,4);
\draw[-] (6.0,4.5) to (2.0,4);
\draw[-] (4.0,0.5) to (2.0,1);
\draw[-] (4.0,0.5) to [bend right=30] (4.0,2);
\draw[-] (4.0,0.5) to [bend right=30] (4.0,4);
\draw[-] (3.5,1) to (4.0,1);
\draw[-] (3.5,2) to (4.0,2);
\draw[-] (3.5,4) to (4.0,4);
\draw[-] (4.0,1.5) to (4.0,1);
\draw[-] (4.0,2.5) to (4.0,2);
\draw[-] (4.0,4.5) to (4.0,4);
\draw[-] (6.0,0.5) to (6.0,1);
\draw[-] (6.0,0.5) to [bend right=30] (6.0,2);
\draw[-] (6.0,0.5) to [bend right=30] (6.0,4);
\draw[-] (5.5,1) to (6.0,1);
\draw[-] (5.5,2) to (6.0,2);
\draw[-] (5.5,4) to (6.0,4);
\draw[-] (6.0,1.5) to (6.0,1);
\draw[-] (6.0,2.5) to (6.0,2);
\draw[-] (6.0,4.5) to (6.0,4);
\draw[-] (8.0,1.5) to (6.0,1);
\draw[-] (8.0,2.5) to (6.0,1);
\draw[-] (8.0,4.5) to (6.0,1);
\draw[-] (8.0,1.5) to (6.0,2);
\draw[-] (8.0,2.5) to (6.0,2);
\draw[-] (8.0,4.5) to (6.0,2);
\draw[-] (8.0,1.5) to (6.0,4);
\draw[-] (8.0,2.5) to (6.0,4);
\draw[-] (8.0,4.5) to (6.0,4);
\draw[-] (8.0,0.5) to (8.0,1);
\draw[-] (8.0,0.5) to [bend right=30] (8.0,2);
\draw[-] (8.0,0.5) to [bend right=30] (8.0,4);
\draw[-] (7.5,1) to (8.0,1);
\draw[-] (7.5,2) to (8.0,2);
\draw[-] (7.5,4) to (8.0,4);
\draw[-] (8.0,1.5) to (8.0,1);
\draw[-] (8.0,2.5) to (8.0,2);
\draw[-] (8.0,4.5) to (8.0,4);
\node[nodeknown] (p0) at (0,0.5) {};
\node[textnode] (l0) at (0,-0.1) {\footnotesize{\textit{get}}};
\node[font=\tiny] (p0label) at (0,0.5) {$\lambda_{0}$};
\node[nodeunknown] (c01) at (-0.5,1) {};
\node[font=\tiny] (c01label) at (-0.5,1) {$\phi_{0_{1}}$};
\node[nodeunknown] (c02) at (-0.5,2) {};
\node[font=\tiny] (c02label) at (-0.5,2) {$\phi_{0_{2}}$};
\node[nodeunknown] (c0n) at (-0.5,4) {};
\node[font=\tiny] (c0nlabel) at (-0.5,4) {$\phi_{0_{n}}$};
\node[nodeknown] (g01) at (0,1.5) {};
\node[font=\tiny] (g01label) at (0,1.5) {$\gamma_{0_{1}}$};
\node[nodeknown] (g02) at (0,2.5) {};
\node[font=\tiny] (g02label) at (0,2.5) {$\gamma_{0_{2}}$};
\node[] (g0dots) at (0,3.375) {$\vdots$};
\node[nodeknown] (g0n) at (0,4.5) {};
\node[font=\tiny] (g0nlabel) at (0,4.5) {$\gamma_{0_{n}}$};
\node[factor] (f01) at (0,1) {};
\node[factor] (f02) at (0,2) {};
\node[factor] (f0n) at (0,4) {};
\node[nodeknown] (p1) at (2.0,0.5) {};
\node[font=\tiny] (p1label) at (2.0,0.5) {$\lambda_{1}$};
\node[nodeunknown] (c11) at (1.5,1) {};
\node[font=\tiny] (c11label) at (1.5,1) {$\phi_{1_{1}}$};
\node[nodeunknown] (c12) at (1.5,2) {};
\node[font=\tiny] (c12label) at (1.5,2) {$\phi_{1_{2}}$};
\node[nodeunknown] (c1n) at (1.5,4) {};
\node[font=\tiny] (c1nlabel) at (1.5,4) {$\phi_{1_{n}}$};
\node[nodeknown] (g11) at (2.0,1.5) {};
\node[font=\tiny] (g11label) at (2.0,1.5) {$\gamma_{1_{1}}$};
\node[nodeknown] (g12) at (2.0,2.5) {};
\node[font=\tiny] (g12label) at (2.0,2.5) {$\gamma_{1_{2}}$};
\node[] (g1dots) at (2.0,3.375) {$\vdots$};
\node[nodeknown] (g1n) at (2.0,4.5) {};
\node[font=\tiny] (g1nlabel) at (2.0,4.5) {$\gamma_{1_{n}}$};
\node[factor] (f11) at (2.0,1) {};
\node[factor] (f12) at (2.0,2) {};
\node[factor] (f1n) at (2.0,4) {};
\node[textnode] (l2) at (4.0,-0.4) {\footnotesize{\textit{pushcore}}};
\node[textnode] (l2) at (4.0,-0.1) {\footnotesize{\textit{the}}};
\node[nodeknown] (p2) at (4.0,0.5) {};
\node[font=\tiny] (p2label) at (4.0,0.5) {$\lambda_{2}$};
\node[nodeunknown] (c21) at (3.5,1) {};
\node[font=\tiny] (c21label) at (3.5,1) {$\phi_{2_{1}}$};
\node[nodeunknown] (c22) at (3.5,2) {};
\node[font=\tiny] (c22label) at (3.5,2) {$\phi_{2_{2}}$};
\node[nodeunknown] (c2n) at (3.5,4) {};
\node[font=\tiny] (c2nlabel) at (3.5,4) {$\phi_{2_{n}}$};
\node[nodeknown] (g21) at (4.0,1.5) {};
\node[font=\tiny] (g21label) at (4.0,1.5) {$\gamma_{2_{1}}$};
\node[nodeknown] (g22) at (4.0,2.5) {};
\node[font=\tiny] (g22label) at (4.0,2.5) {$\gamma_{2_{2}}$};
\node[] (g2dots) at (4.0,3.375) {$\vdots$};
\node[nodeknown] (g2n) at (4.0,4.5) {};
\node[font=\tiny] (g2nlabel) at (4.0,4.5) {$\gamma_{2_{n}}$};
\node[factor] (f21) at (4.0,1) {};
\node[factor] (f22) at (4.0,2) {};
\node[factor] (f2n) at (4.0,4) {};
\node[textnode] (l3) at (6.0,-0.1) {\footnotesize{\textit{from}}};
\node[nodeknown] (p3) at (6.0,0.5) {};
\node[font=\tiny] (p3label) at (6.0,0.5) {$\lambda_{3}$};
\node[nodeunknown] (c31) at (5.5,1) {};
\node[font=\tiny] (c31label) at (5.5,1) {$\phi_{3_{1}}$};
\node[nodeunknown] (c32) at (5.5,2) {};
\node[font=\tiny] (c32label) at (5.5,2) {$\phi_{3_{2}}$};
\node[nodeunknown] (c3n) at (5.5,4) {};
\node[font=\tiny] (c3nlabel) at (5.5,4) {$\phi_{3_{n}}$};
\node[nodeknown] (g31) at (6.0,1.5) {};
\node[font=\tiny] (g31label) at (6.0,1.5) {$\gamma_{3_{1}}$};
\node[nodeknown] (g32) at (6.0,2.5) {};
\node[font=\tiny] (g32label) at (6.0,2.5) {$\gamma_{3_{2}}$};
\node[] (g3dots) at (6.0,3.375) {$\vdots$};
\node[nodeknown] (g3n) at (6.0,4.5) {};
\node[font=\tiny] (g3nlabel) at (6.0,4.5) {$\gamma_{3_{n}}$};
\node[factor] (f31) at (6.0,1) {};
\node[factor] (f32) at (6.0,2) {};
\node[factor] (f3n) at (6.0,4) {};
\node[textnode] (l4) at (8.0,-0.4) {\footnotesize{\textit{tooltray}}};
\node[textnode] (l4) at (8.0,-0.1) {\footnotesize{\textit{the}}};
\node[nodeknown] (p4) at (8.0,0.5) {};
\node[font=\tiny] (p4label) at (8.0,0.5) {$\lambda_{4}$};
\node[nodeunknown] (c41) at (7.5,1) {};
\node[font=\tiny] (c41label) at (7.5,1) {$\phi_{4_{1}}$};
\node[nodeunknown] (c42) at (7.5,2) {};
\node[font=\tiny] (c42label) at (7.5,2) {$\phi_{4_{2}}$};
\node[nodeunknown] (c4n) at (7.5,4) {};
\node[font=\tiny] (c4nlabel) at (7.5,4) {$\phi_{4_{n}}$};
\node[nodeknown] (g41) at (8.0,1.5) {};
\node[font=\tiny] (g41label) at (8.0,1.5) {$\gamma_{4_{1}}$};
\node[nodeknown] (g42) at (8.0,2.5) {};
\node[font=\tiny] (g42label) at (8.0,2.5) {$\gamma_{4_{2}}$};
\node[] (g4dots) at (8.0,3.375) {$\vdots$};
\node[nodeknown] (g4n) at (8.0,4.5) {};
\node[font=\tiny] (g4nlabel) at (8.0,4.5) {$\gamma_{4_{n}}$};
\node[factor] (f41) at (8.0,1) {};
\node[factor] (f42) at (8.0,2) {};
\node[factor] (f4n) at (8.0,4) {};
\node[textnode] (pt1) at (7.625,-1.25) {\footnotesize{DT}};
\node[textnode] (pt2) at (8.375,-1.25) {\footnotesize{NN}};
\node[textnode] (pt3) at (6.0,-1.25) {\footnotesize{IN}};
\node[textnode] (pt4) at (3.625,-1.25) {\footnotesize{DT}};
\node[textnode] (pt5) at (4.375,-1.25) {\footnotesize{NN}};
\node[textnode] (pt6) at (0,-1.25) {\footnotesize{VB}};
\node[textnode] (pt7) at (8.0,-2.0) {\footnotesize{NP $\left(\lambda_{4}\right)$}};
\draw[] (pt1) to (7.625,-1.625) to (8.0,-1.625) to (pt7);
\draw[] (pt2) to (8.375,-1.625) to (8.0,-1.625) to (pt7);
\node[textnode] (pt8) at (4.0,-2.0) {\footnotesize{NP $\left(\lambda_{2}\right)$}};
\draw[] (pt4) to (3.625,-1.625) to (4.0,-1.625) to (pt8);
\draw[] (pt5) to (4.375,-1.625) to (4.0,-1.625) to (pt8);
\node[textnode] (pt9) at (7.0,-2.75) {\footnotesize{PP $\left(\lambda_{3}\right)$}};
\draw[] (pt3) to (6.0,-2.375) to (7.0,-2.375) to (pt9);
\draw[] (pt7) to (8.0,-2.375) to (7.0,-2.375) to (pt9);
\node[textnode] (pt10) at (5.5,-3.5) {\footnotesize{NP $\left(\lambda_{1}\right)$}};
\draw[] (pt8) to (4.0,-3.125) to (5.5,-3.125) to (pt10);
\draw[] (pt9) to (7.0,-3.125) to (5.5,-3.125) to (pt10);
\node[textnode] (pt11) at (2.75,-4.25) {\footnotesize{VP $\left(\lambda_{0}\right)$}};
\draw[] (pt6) to (0,-3.875) to (2.75,-3.875) to (pt11);
\draw[] (pt10) to (5.5,-3.875) to (2.75,-3.875) to (pt11);
\end{tikzpicture}
\caption{A visualization of (top) the DCG factor graph for the expression ``get the pushcore from the tooltray" aligned with (bottom) the associated parse tree. Shaded nodes denote observed random variables, while those rendered in white are latent.} \label{fig:dcg-example}
\end{figure}
Using the \textit{NUI} HROV, we performed a proof-of-concept demonstration of an architecture that allows user control of an ROV manipulator using natural language provided as text or speech using a cloud-based speech recognizer. We frame natural language understanding as a symbol grounding problem~\citep{harnad90}, whereby the objective is to map words in the utterance to their corresponding referents in a symbolic representation of the robot's state and action spaces. Consistent with contemporary approaches to language understanding, we formulate grounding as probabilistic inference over a learned distribution that models this mapping. In particular, given the syntactic parse of a natural language command $\Lambda$, we employ maximum a posteriori inference over the power set of referent symbols $\mathcal{P}({\Gamma})$
\begin{equation}
  \Gamma^* = \underset{\mathcal{P}}{\text{arg max}} \; p( {\Gamma}  \vert {\Lambda}, S).
  \label{eqn:nlu}
\end{equation}
where $S$ is a variable that denotes the robot's model of the environment (e.g., the type and location of different tools). We model this distribution using the Distributed Correspondence Graph (DCG)~\cite{howard14}, a factor graph (Fig.~\ref{fig:dcg-example}) that approximates the conditional probabilities of a Boolean correspondence variable $\phi_{i_j}$ that indicates the association between a specific symbol $\gamma_{i_j} \in \Gamma$, which may correspond to an object, action, or location, and each word $\lambda_i \in \Lambda$. Critically, the composition of the DCG factor graph follows the hierarchical structure of language. The model is trained on corpora of annotated examples (i.e., words from natural language utterances paired with their corresponding groundings), whereby we independently learn the conditional probabilities for the different language elements, such as nouns (e.g., ``the pushcore'', ``tool'', and ``tool tray''), verbs (e.g., ``retrieve'', ``release'', and ``stow''), and prepositions (e.g., ``inside'' and ``towards''). Together with the fact that the factor graph exploits the compositional nature of language, the DCG model is able to generalize beyond the specific utterances present in the training data.

\begin{figure}[!h]
    \centering
    \subfigure[Language commands the system to plan a path to the sample location]{\includegraphics[width=1.0\linewidth]{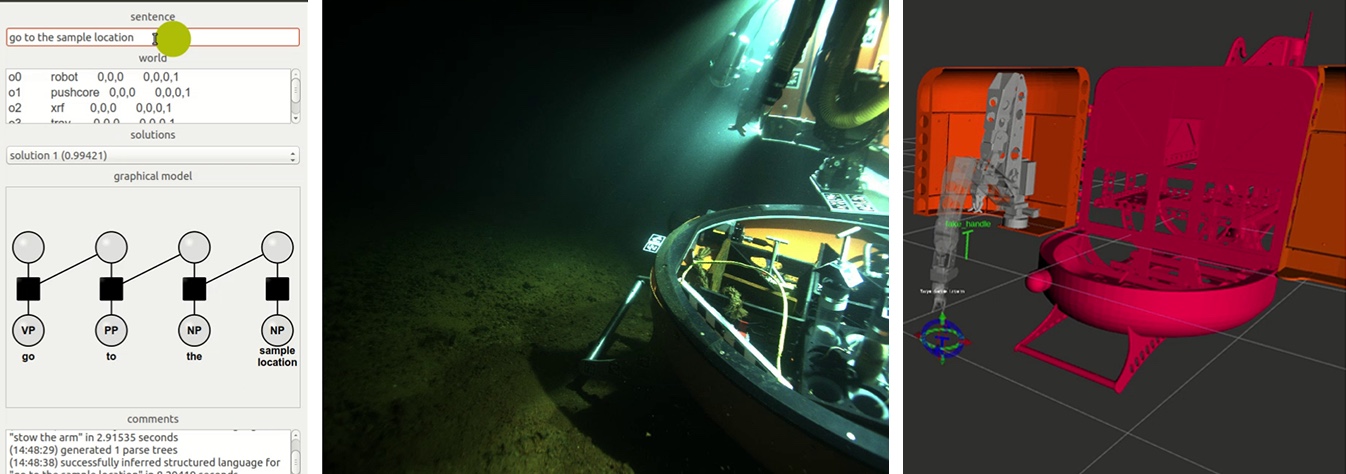}\label{fig:audio_a}}\\
    \subfigure[Language command to execute planned path]{\includegraphics[width=1.0\linewidth]{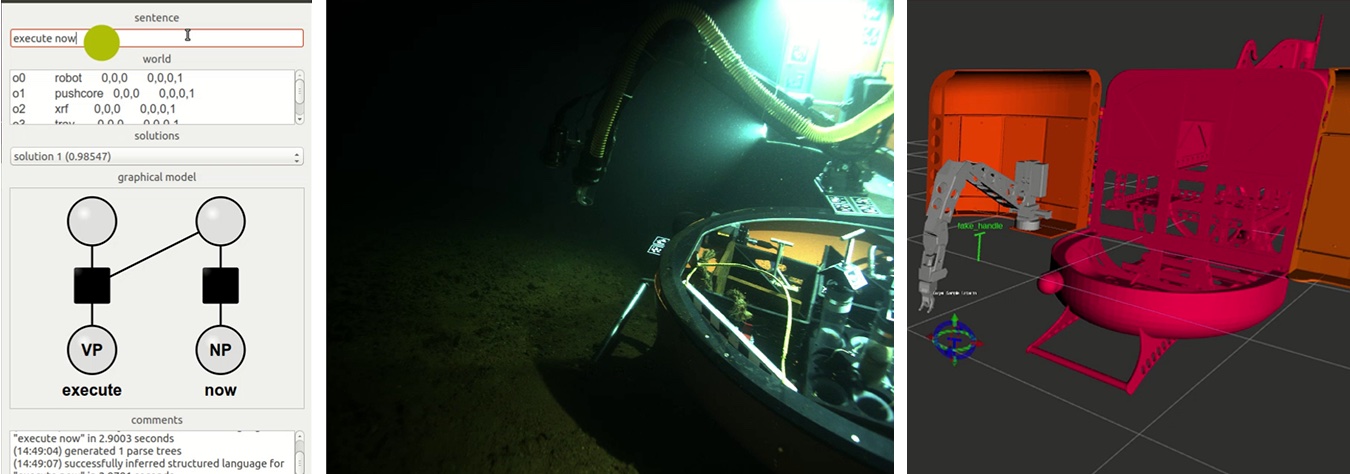}\label{fig:audio_b}}
    \caption{Demonstration of a proof-of-concept framework that enabled operators to interact with our autonomous manipulation architecture using natural language. Given input in the form of free-form text, either entered by the operator or output by a cloud-based speech recognizer, we (left) infer the meaning of the command using a probabilistic language model. \subref{fig:audio_a} In the case of the command to ``go to the sample location'', our system (top-right) determines the goal configuration and solves for a collision-free path in configuration space. \subref{fig:audio_b} Given the command to ``execute now'', the manipulator then (bottom-right) executes the planned path to the goal.}
    \label{fig:audio}
\end{figure}
For our initial implementation, the space of symbols $\Gamma$ included the tools that the arm was able to grasp and the different steps that comprised the state machine underlying the pick-and-place pipeline. Figure~\ref{fig:audio} presents an example from a deployment at the Kolumbo caldera in which natural language  was used to initiate path planning to the sample location and then to command the manipulator to execute the planned path. Several tests were conducted in which the manipulator was commanded through natural language input to move to a location specified by the sample marker in the planning interface and then return to the home position. These tests demonstrated the flexibility of our system to incorporate different operational modalities through high level abstraction.

\subsection{\revision{Performance Analysis}}

\begin{figure}[!h]
    \centering
    \includegraphics[width=1.0\linewidth]{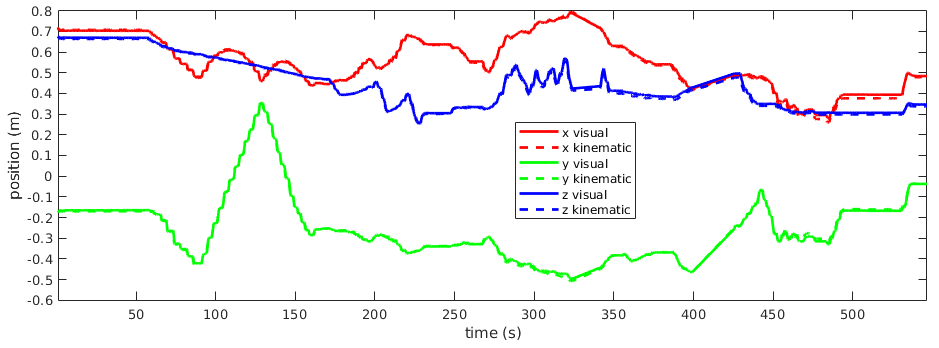}
    \caption{Plot of the TagSLAM estimated trajectory (visual) of the fisheye camera versus the trajectory estimated from the manipulator joint feedback (kinematic). The trajectory is plotted separately for each coordinate axis with respect to the manipulator base frame.}
    \label{fig:trajectories}
\end{figure}

\revision{We evaluated the overall accuracy of the calibrated kinematic and visual system on the testbed. For this evaluation, we placed an AprilTag grid in the scene and activated every joint of the manipulator while keeping the tag grid in view of the fisheye camera. We used TagSLAM~\citet{pfrommer2019tagslam} to generate a visual SLAM estimated trajectory of the fisheye camera, and we used the manipulator joint feedback to also generate a kinematic based trajectory. These trajectories are plotted against each other in figure~\ref{fig:trajectories}. The overall mean error between the kinematic and visual based trajectories is 1.16cm, the maximum trajectory error is 3.27cm, and the standard deviation is 0.65cm. These results are a conservative estimate of the system calibration accuracy as there are several sources of error: the visual SLAM accuracy degrades when the tags are near the edge of the fisheye image; the agreement between the kinematic and visual based pose of the fisheye camera depends on the accuracy of the hand-eye calibration; the joint feedback and fisheye images are not synchronized; and the SLAM and kinematic reference frames were mapped to each other through a single fisheye frame estimate of the tag grid pose, projected from the fisheye frame through the kinematic chain to the manipulator base frame. However, we have demonstrated in our experimental trials that the system accuracy is good enough to perform high level automation tasks.}

\revision{The first five joints of the KRAFT manipulator are controlled through joint position set-point commands. We analyzed the manipulator control response for bias or hysteresis, as these are known issues with hydraulic actuation. Figure~\ref{fig:joint_path_testbed} shows plots of the commanded versus feedback positions during actuation of each joint of the testbed manipulator. The figure also shows a histogram of errors, binned at 0.5\degree, between the commanded and followed joint trajectories. All of the joints except the wrist pitch exhibit small bias and no major hysteresis is evident. The wrist pitch exhibits a bias of approximately 1.5\degree, which is significant, but did not prevent completion of high level automation tasks. Figure~\ref{fig:joint_path_nui} shows the same plots for the \textit{NUI} HROV manipulator made from data recorded during the field trials in Greece. The elbow and wrist joints exhibit little bias or hysteresis. However, both of the shoulder joints exhibited high bias, particularly the shoulder yaw joint, which had a bias of approximately 8\degree. This high bias prevented the completion of a pick-and-place manipulation task during the field trial. Our current control system relies on the manipulator valve controller to move to the desired set-point and does not account for bias in the control response. It will be critical in the future to incorporate an adaptive controller into the system that can account for bias and hysteresis in the hydraulic actuators.}

\begin{figure}
    \centering
    \subfigure[Shoulder Yaw Trajectory]{\includegraphics[width=0.5\linewidth,trim={4cm 8.5cm 4cm 8.5cm},clip]{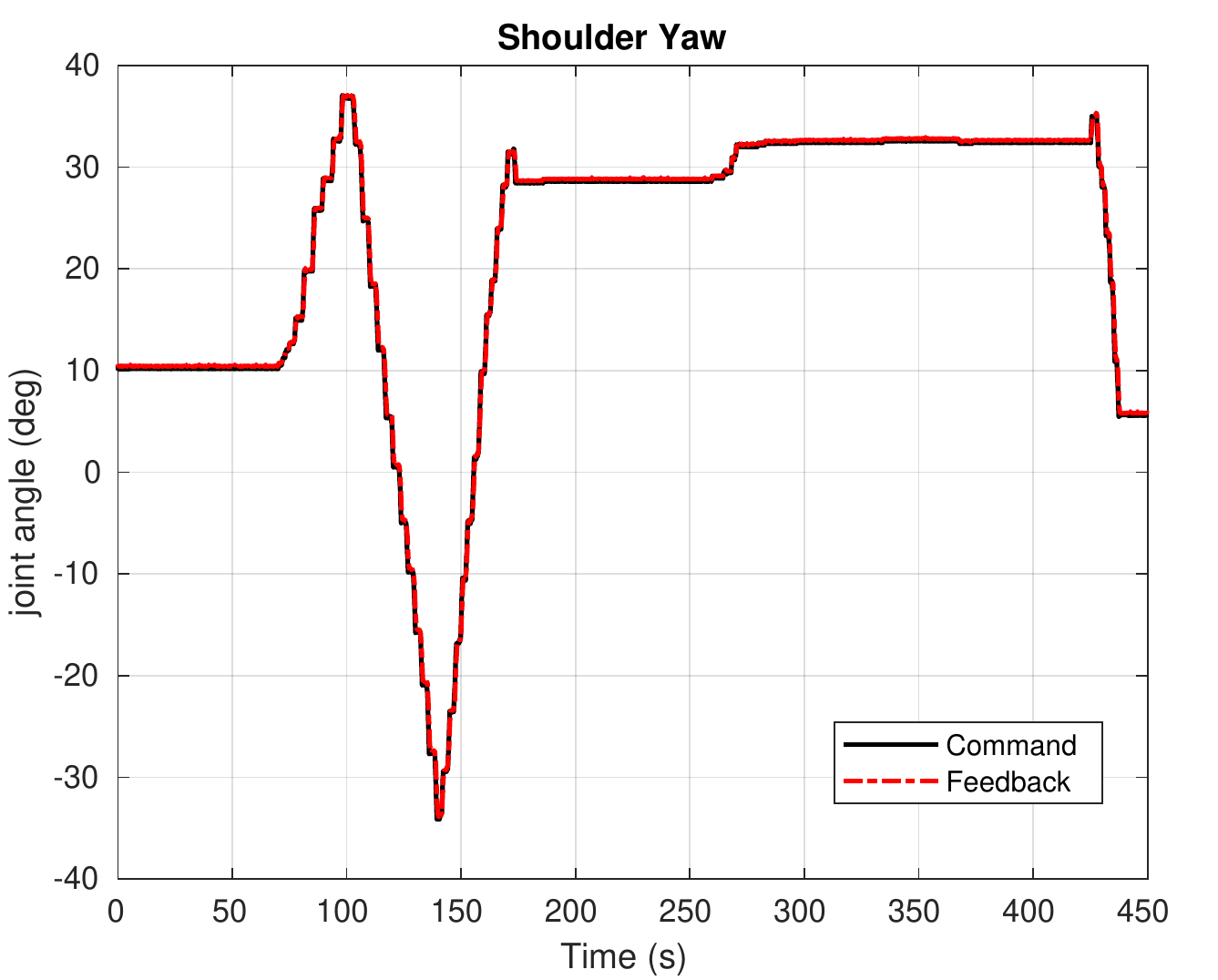}}\hfil
    \subfigure[Shoulder Yaw Error Histogram]{\includegraphics[width=0.5\linewidth,trim={4cm 8.5cm 4cm 8.5cm},clip]{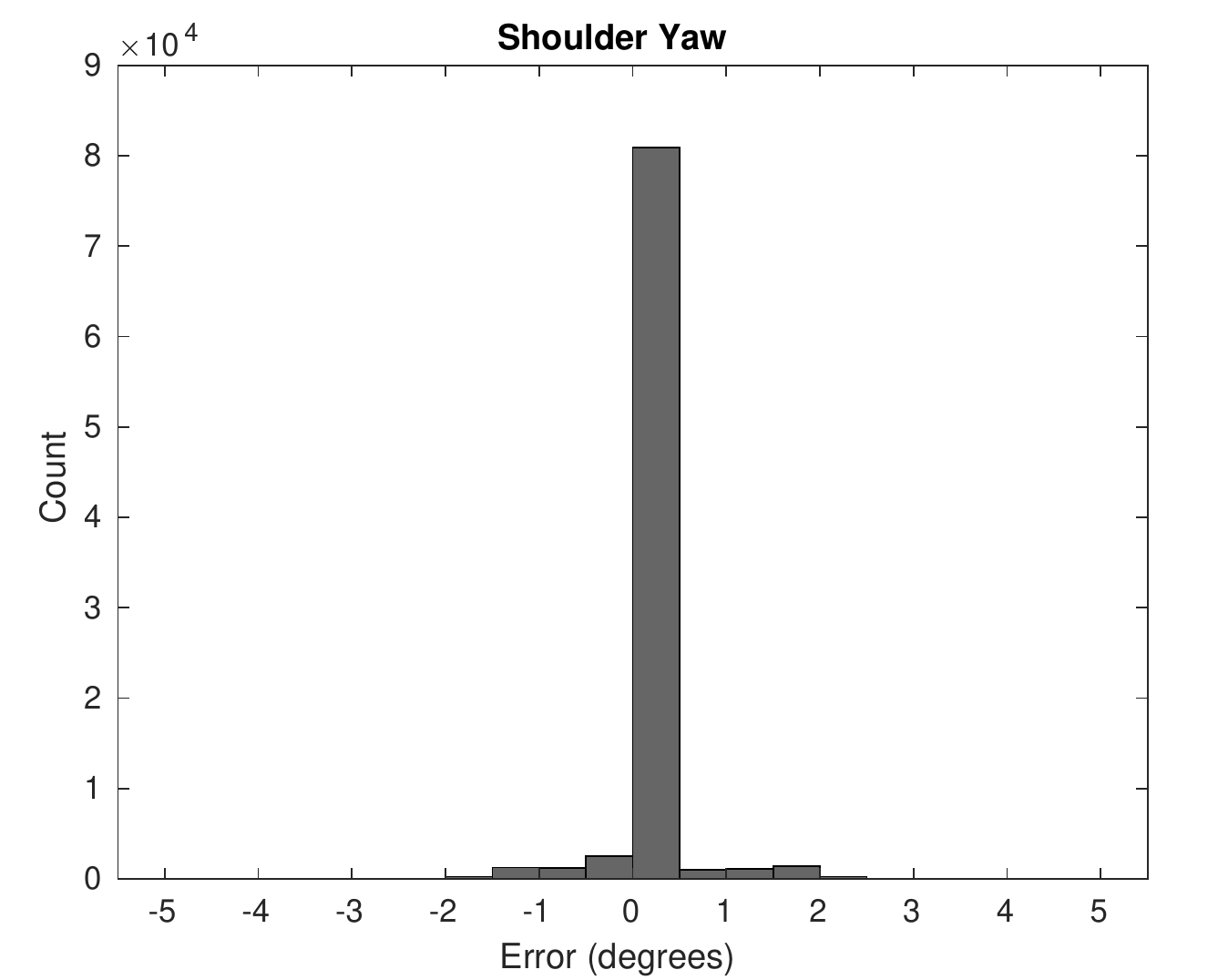}}\\
    \subfigure[Shoulder Pitch Trajectory]{\includegraphics[width=0.5\linewidth,trim={4cm 8.5cm 4cm 8.5cm},clip]{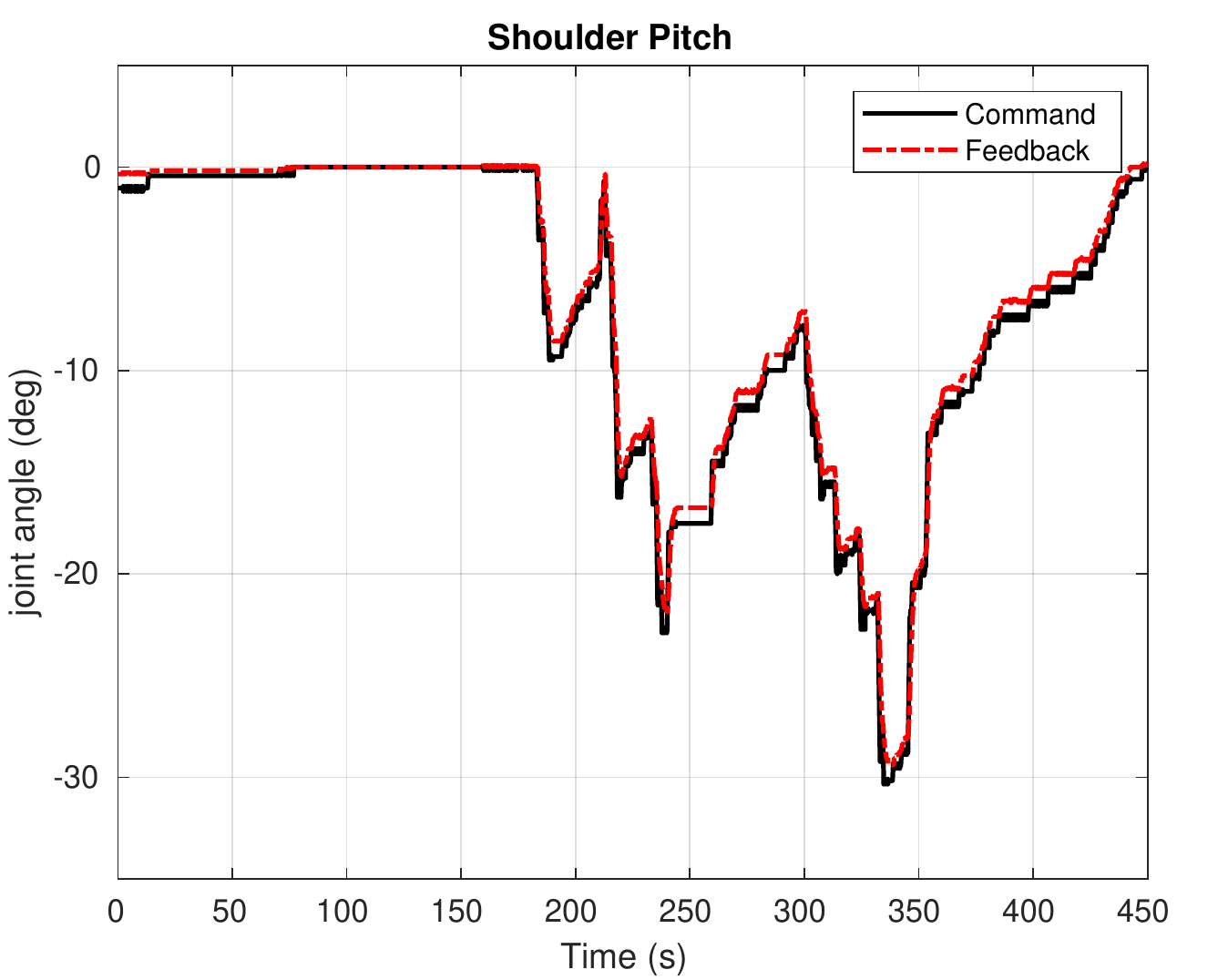}}\hfil
    \subfigure[Shoulder Pitch Error Histogram]{\includegraphics[width=0.5\linewidth,trim={4cm 8.5cm 4cm 8.5cm},clip]{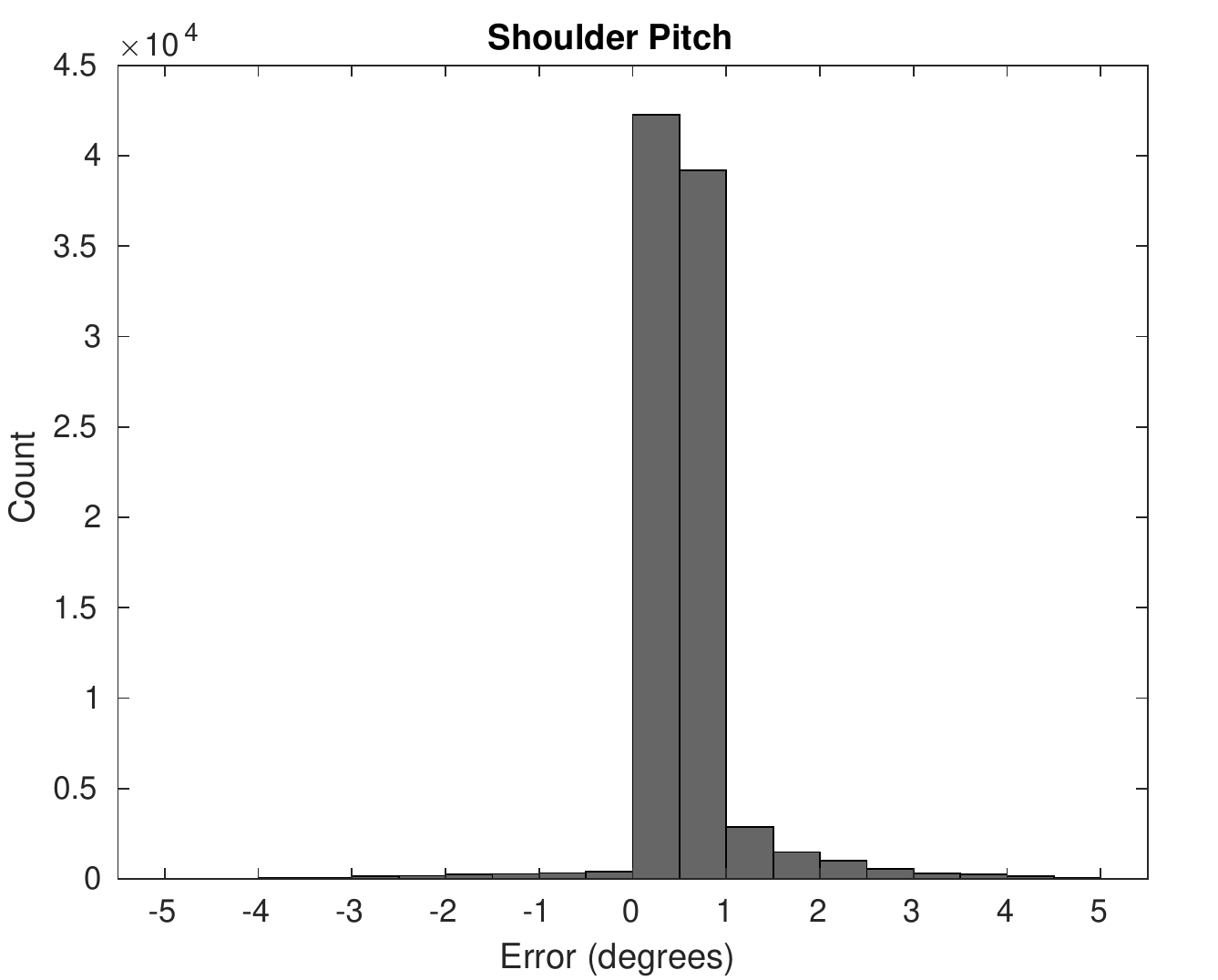}}\\
    \subfigure[Elbow Trajectory]{\includegraphics[width=0.5\linewidth,trim={4cm 8.5cm 4cm 8.5cm},clip]{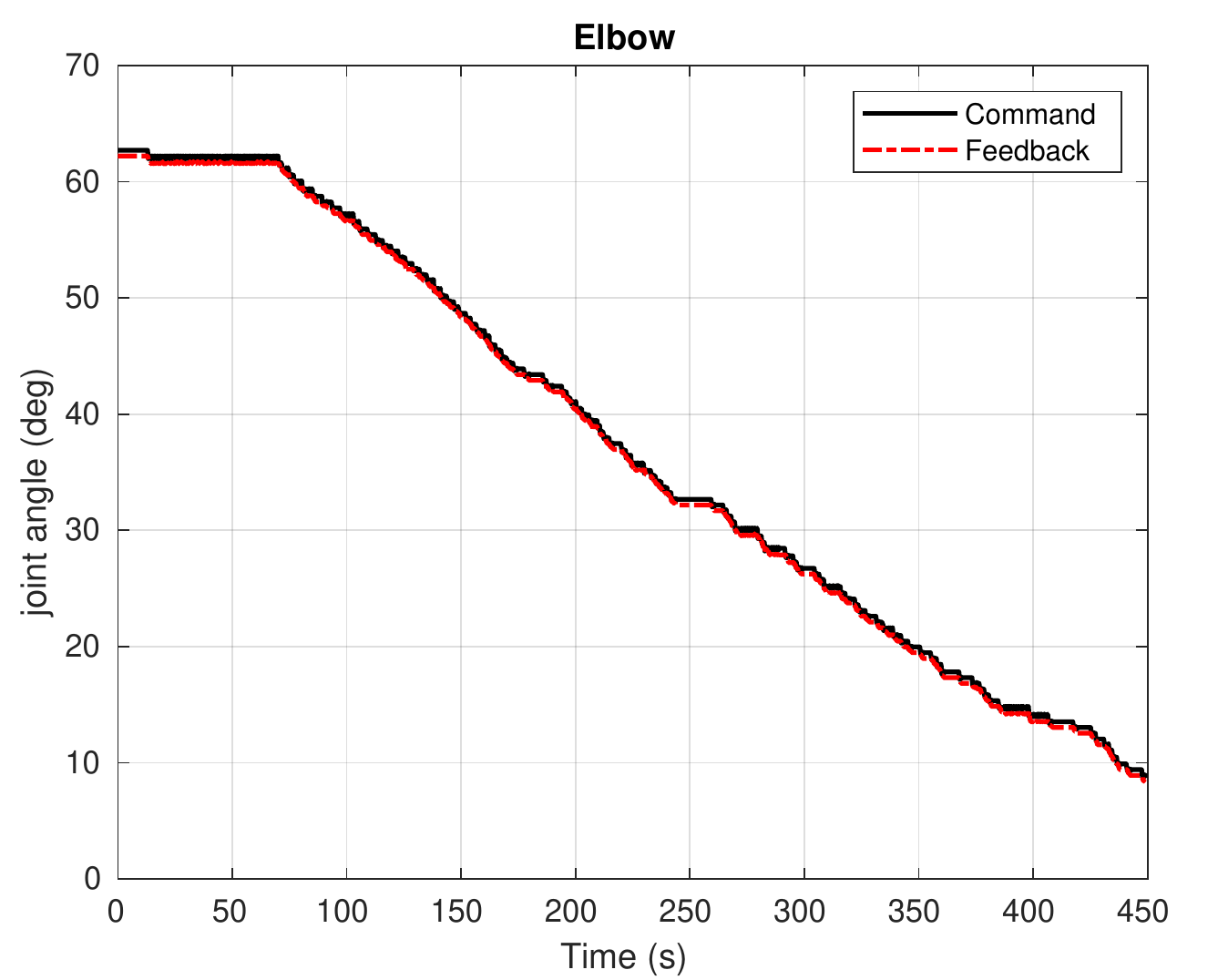}}\hfil
    \subfigure[Elbow Error Histogram]{\includegraphics[width=0.5\linewidth,trim={4cm 8.5cm 4cm 8.5cm},clip]{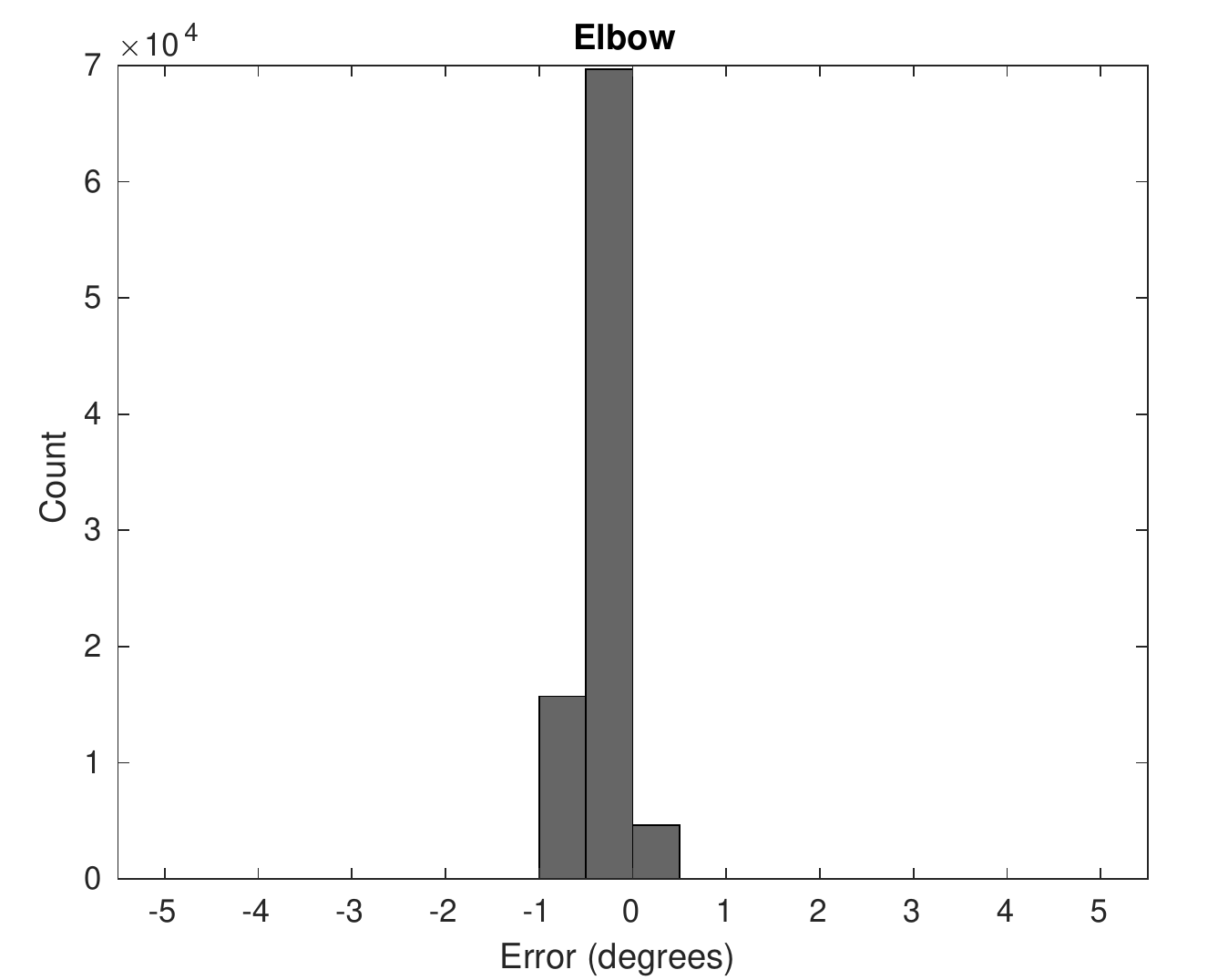}}
    \phantomcaption{} %
\end{figure}
\begin{figure}[!ht]\ContinuedFloat
    \centering
    \subfigure[Wrist Pitch Trajectory]{\includegraphics[width=0.5\linewidth,trim={4cm 8.5cm 4cm 8.5cm},clip]{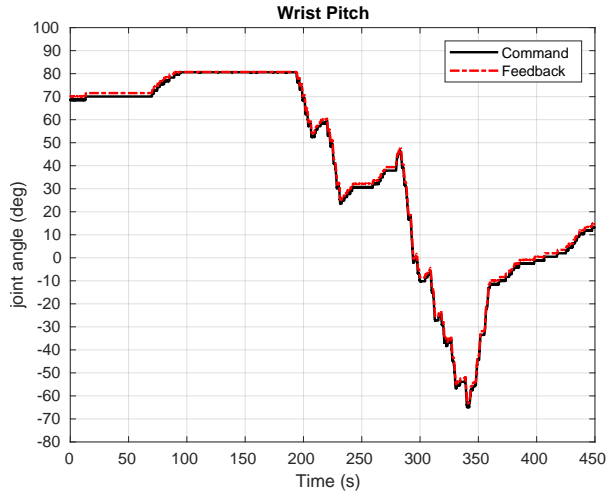}}\hfil
    \subfigure[Wrist Pitch Error Histogram]{\includegraphics[width=0.5\linewidth,trim={4cm 8.5cm 4cm 8.5cm},clip]{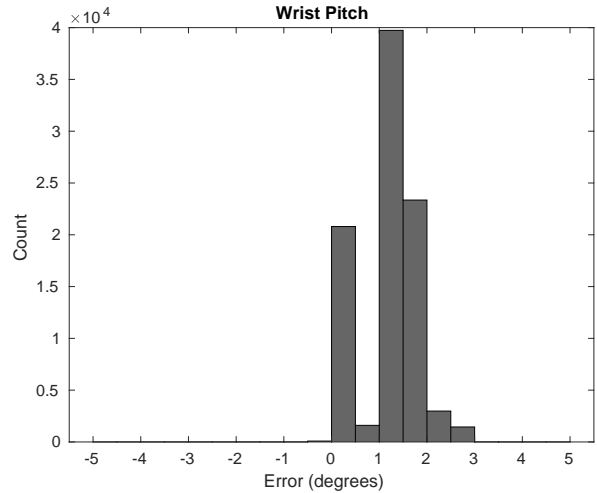}}\\
    \subfigure[Wrist Yaw Trajectory]{\includegraphics[width=0.5\linewidth,trim={4cm 8.5cm 4cm 8.5cm},clip]{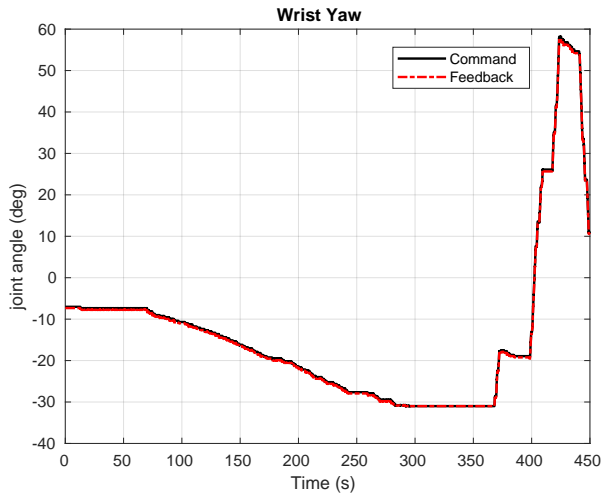}}\hfil
    \subfigure[Wrist Yaw Error Histogram]{\includegraphics[width=0.5\linewidth,trim={4cm 8.5cm 4cm 8.5cm},clip]{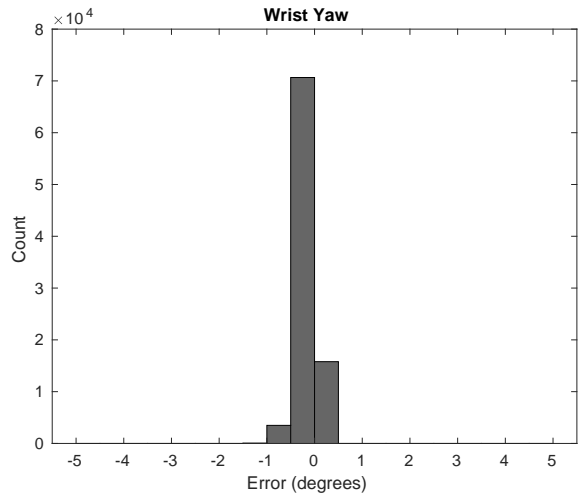}}\\
    \caption{Plot of commanded versus followed joint trajectories for the \textit{testbed} manipulator.}
    \label{fig:joint_path_testbed}
\end{figure}

\begin{figure}
    \centering
    \subfigure[Shoulder Yaw Trajectory]{\includegraphics[width=0.5\linewidth,trim={4cm 8.5cm 4cm 8.5cm},clip]{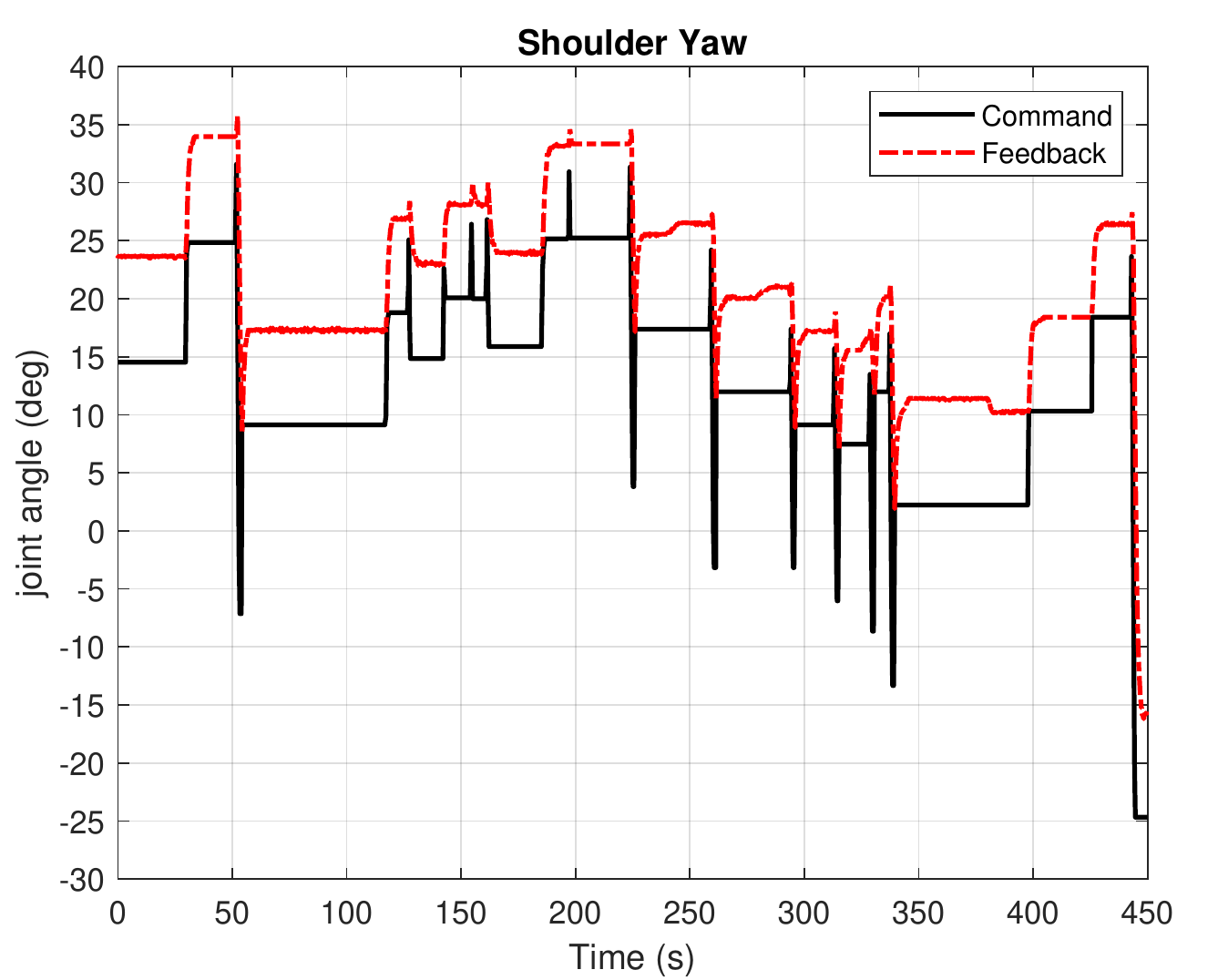}}\hfil
    \subfigure[Shoulder Yaw Error Histogram]{\includegraphics[width=0.5\linewidth,trim={4cm 8.5cm 4cm 8.5cm},clip]{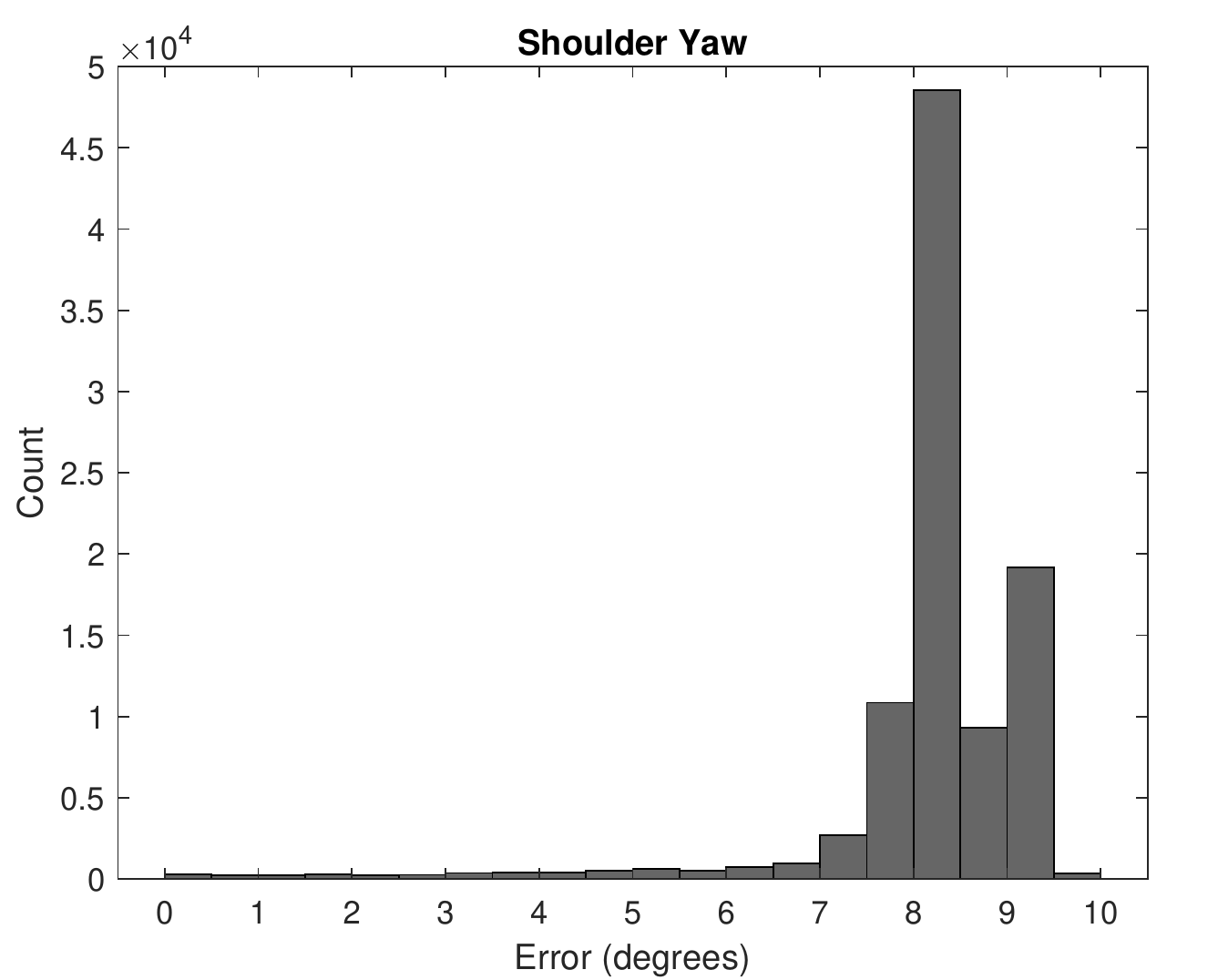}}\\
    \subfigure[Shoulder Pitch Trajectory]{\includegraphics[width=0.5\linewidth,trim={4cm 8.5cm 4cm 8.5cm},clip]{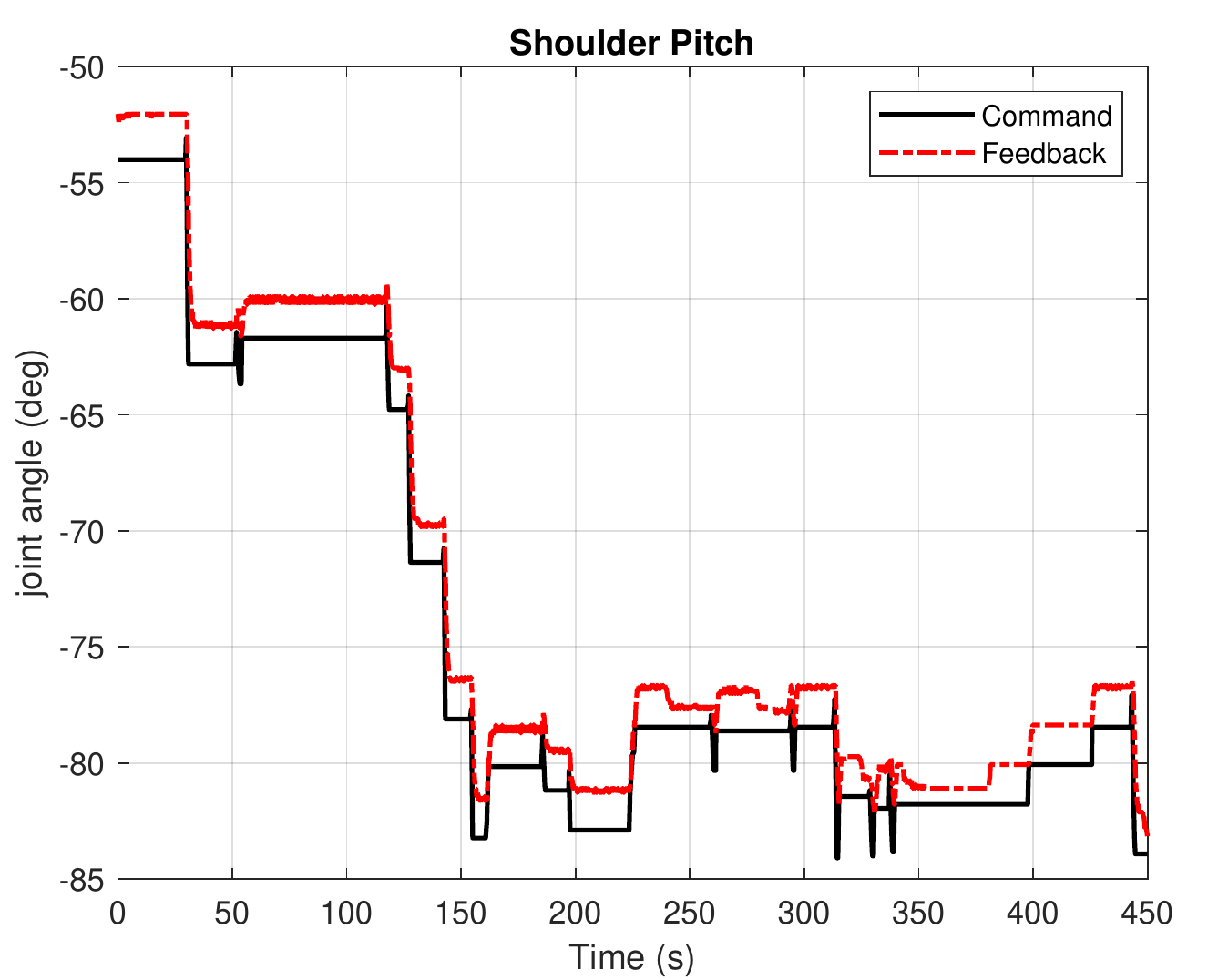}}\hfil
    \subfigure[Shoulder Pitch Error Histogram]{\includegraphics[width=0.5\linewidth,trim={4cm 8.5cm 4cm 8.5cm},clip]{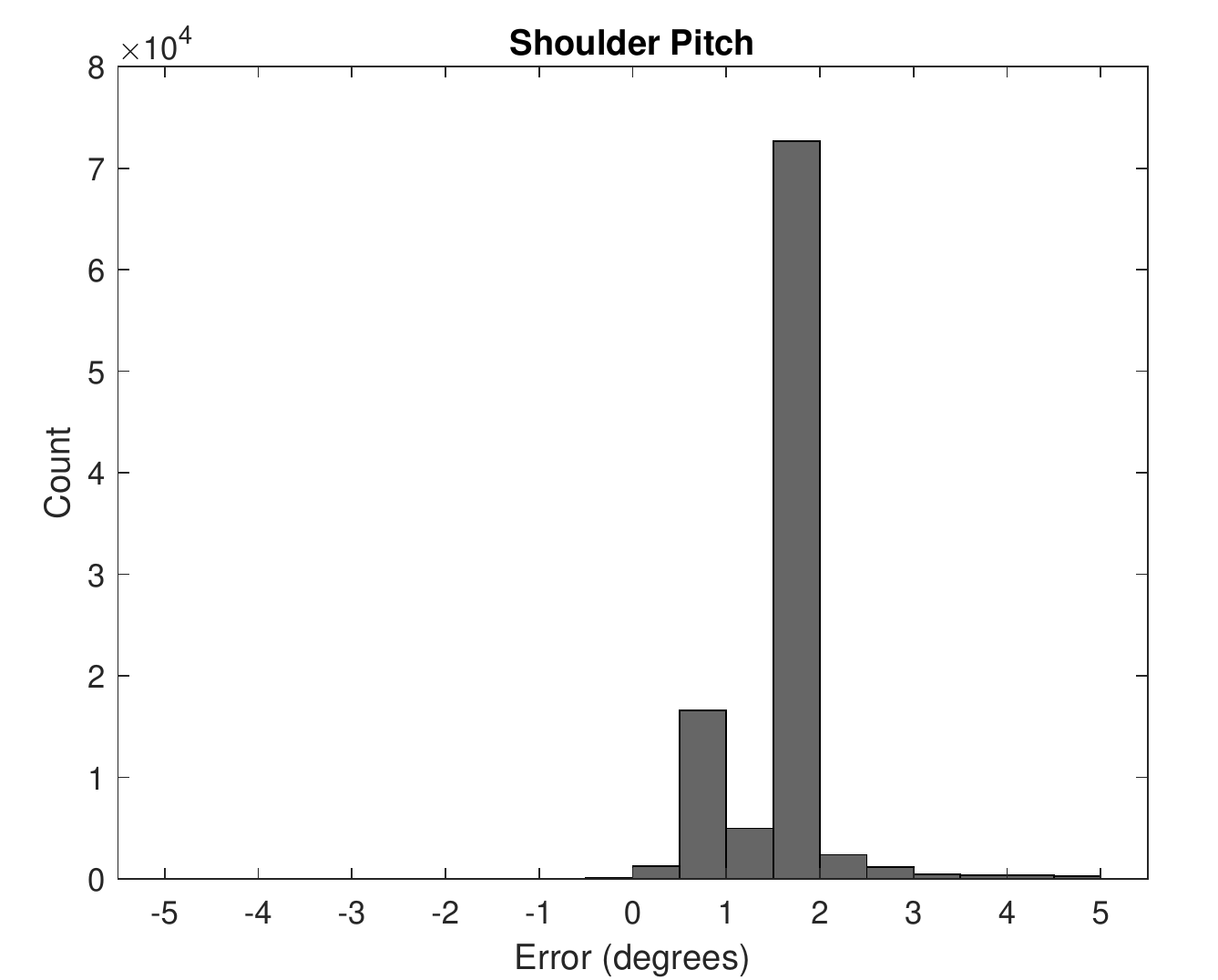}}\\
    \subfigure[Elbow Trajectory]{\includegraphics[width=0.5\linewidth,trim={4cm 8.5cm 4cm 8.5cm},clip]{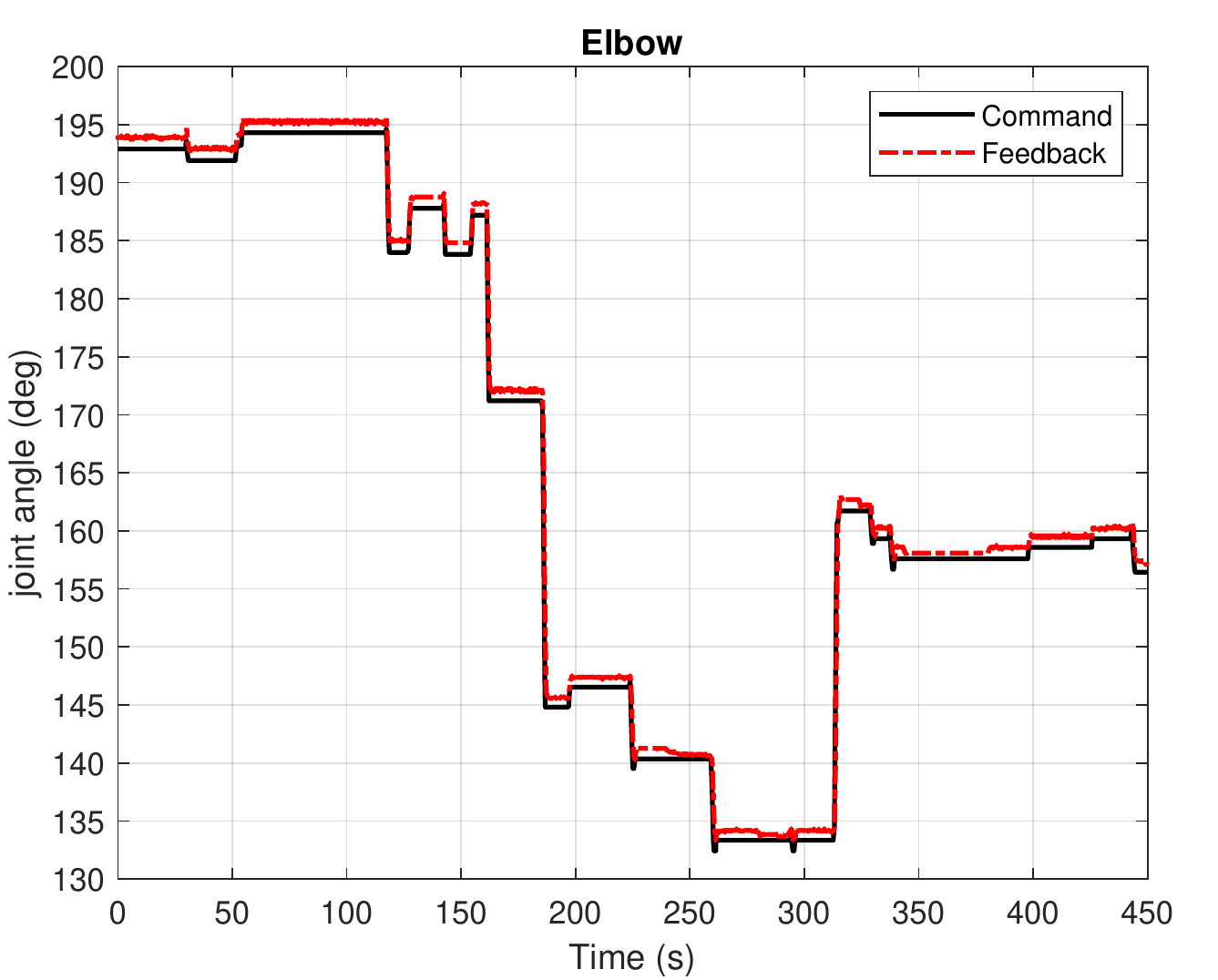}}\hfil
    \subfigure[Elbow Error Histogram]{\includegraphics[width=0.5\linewidth,trim={4cm 8.5cm 4cm 8.5cm},clip]{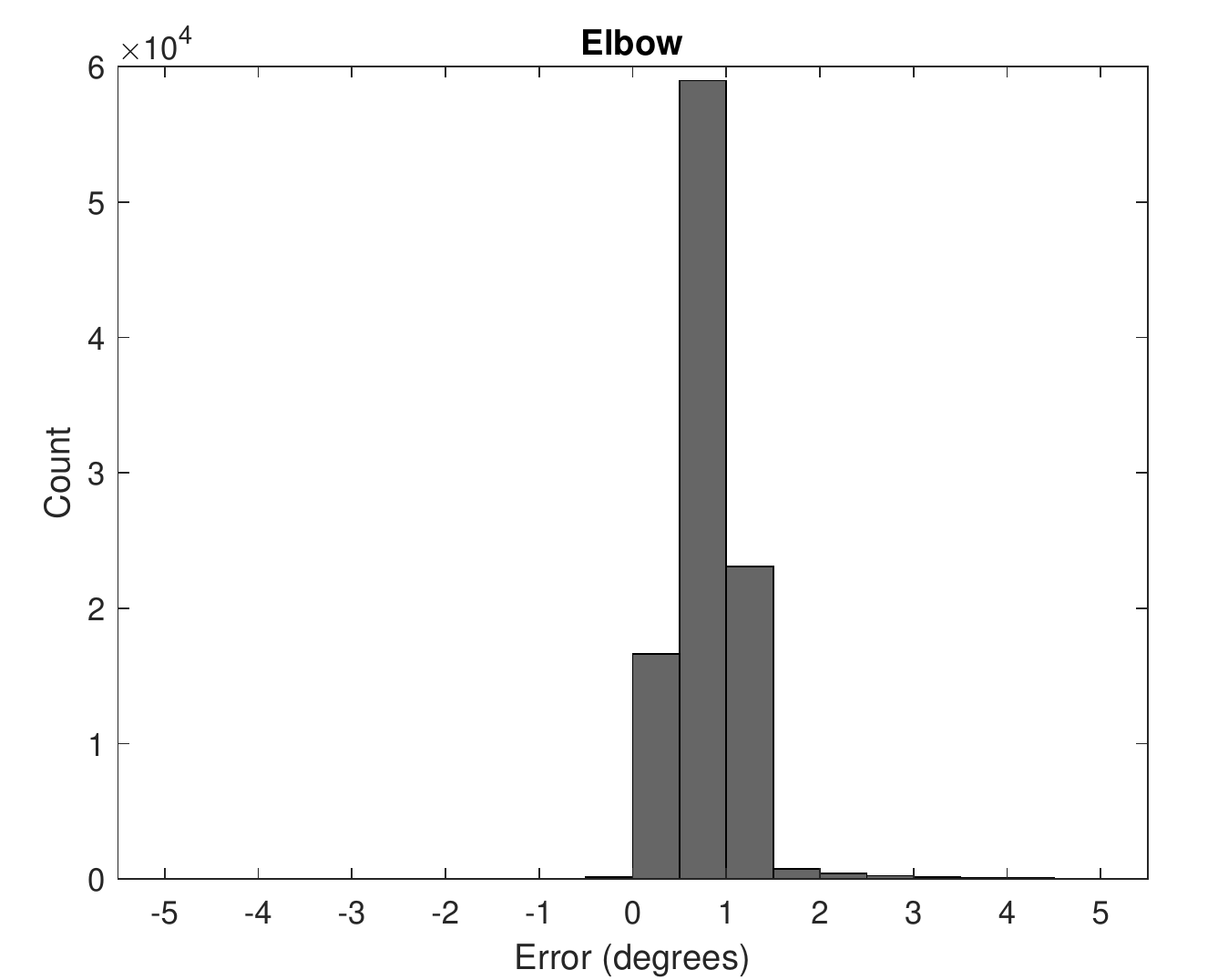}}
    \phantomcaption{}
\end{figure}
\begin{figure}[!ht]\ContinuedFloat
    \centering
    \subfigure[Wrist Pitch Trajectory]{\includegraphics[width=0.5\linewidth,trim={4cm 8.5cm 4cm 8.5cm},clip]{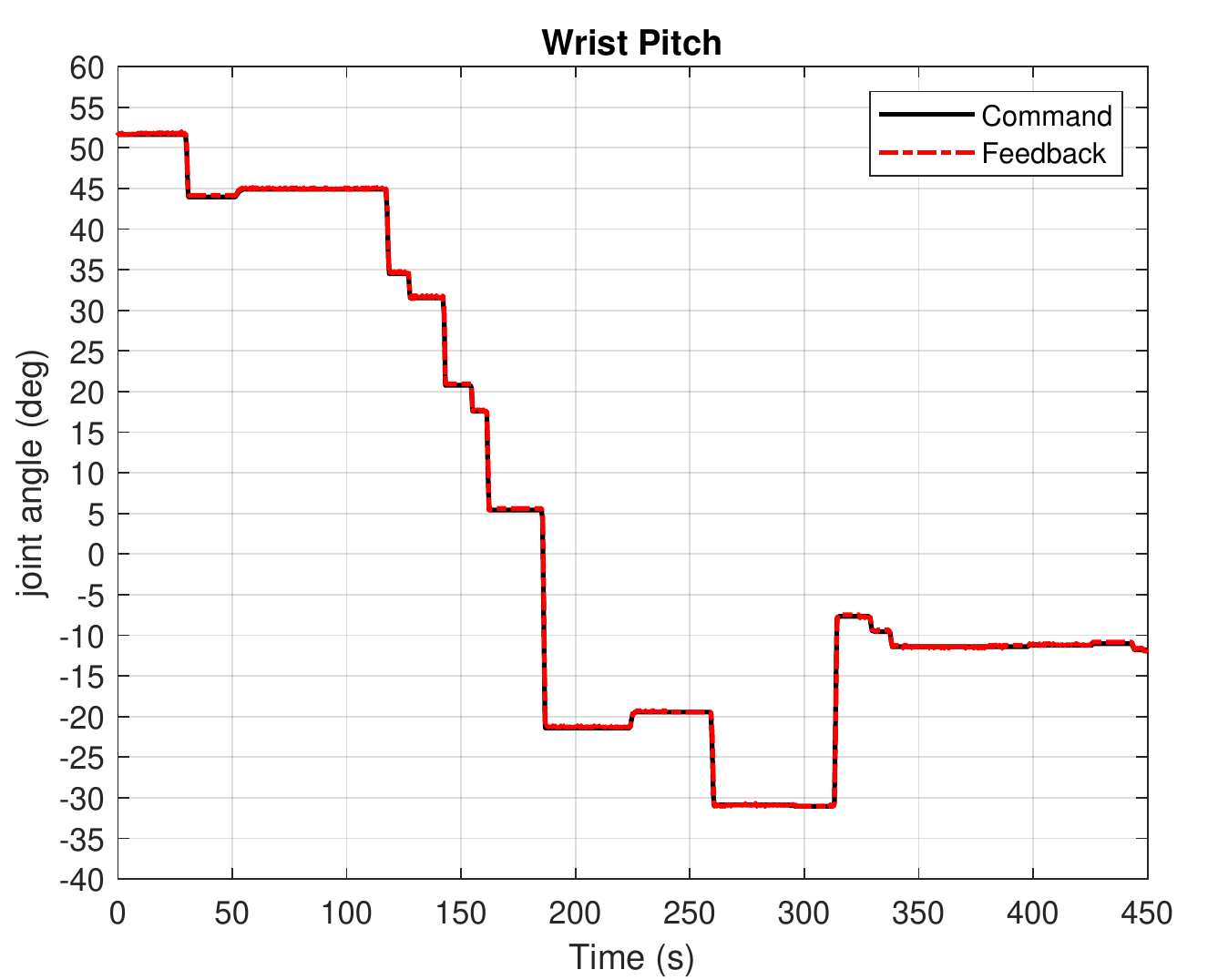}}\hfil
    \subfigure[Wrist Pitch Error Histogram]{\includegraphics[width=0.5\linewidth,trim={4cm 8.5cm 4cm 8.5cm},clip]{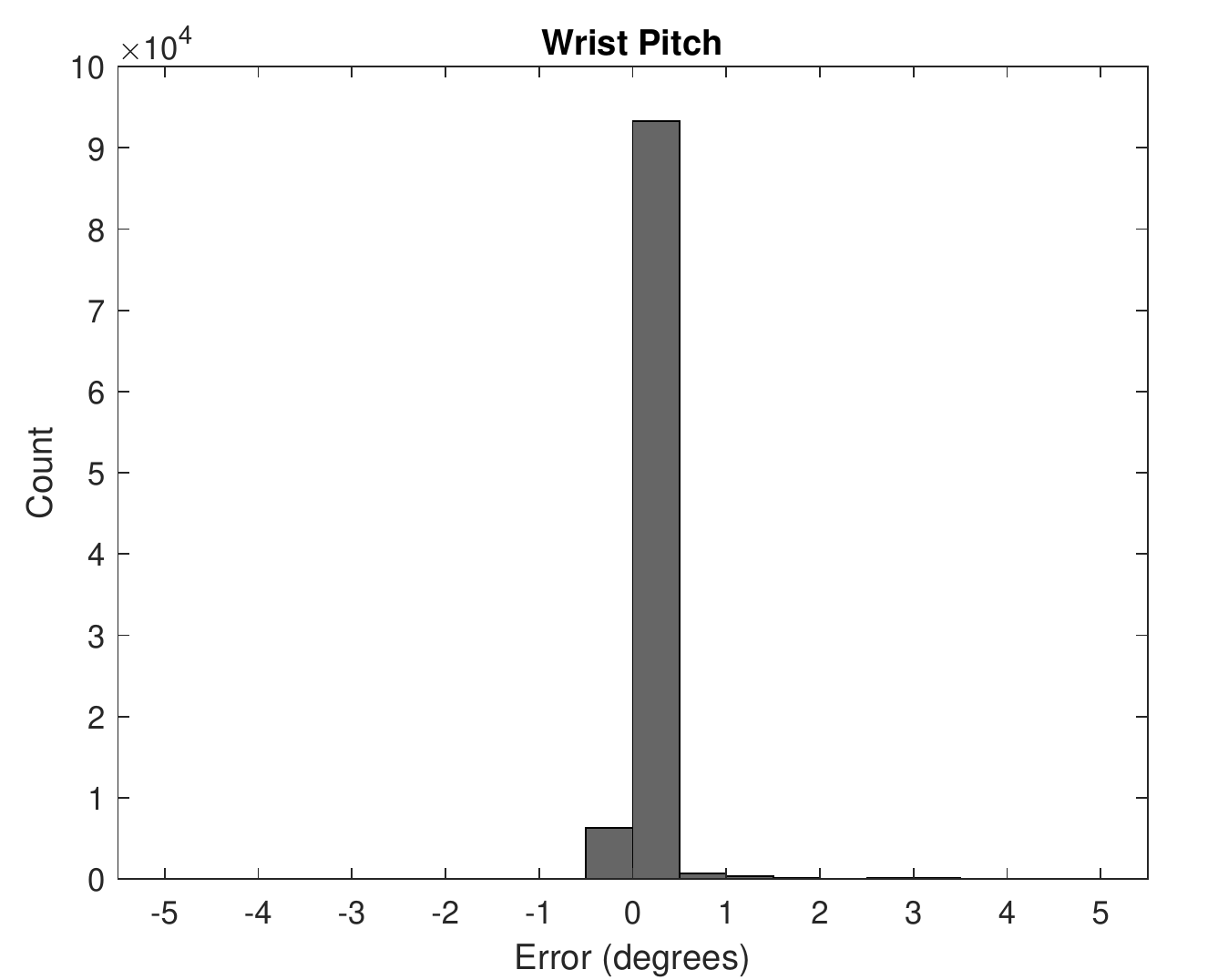}}\\
    \subfigure[Wrist Yaw Trajectory]{\includegraphics[width=0.5\linewidth,trim={4cm 8.5cm 4cm 8.5cm},clip]{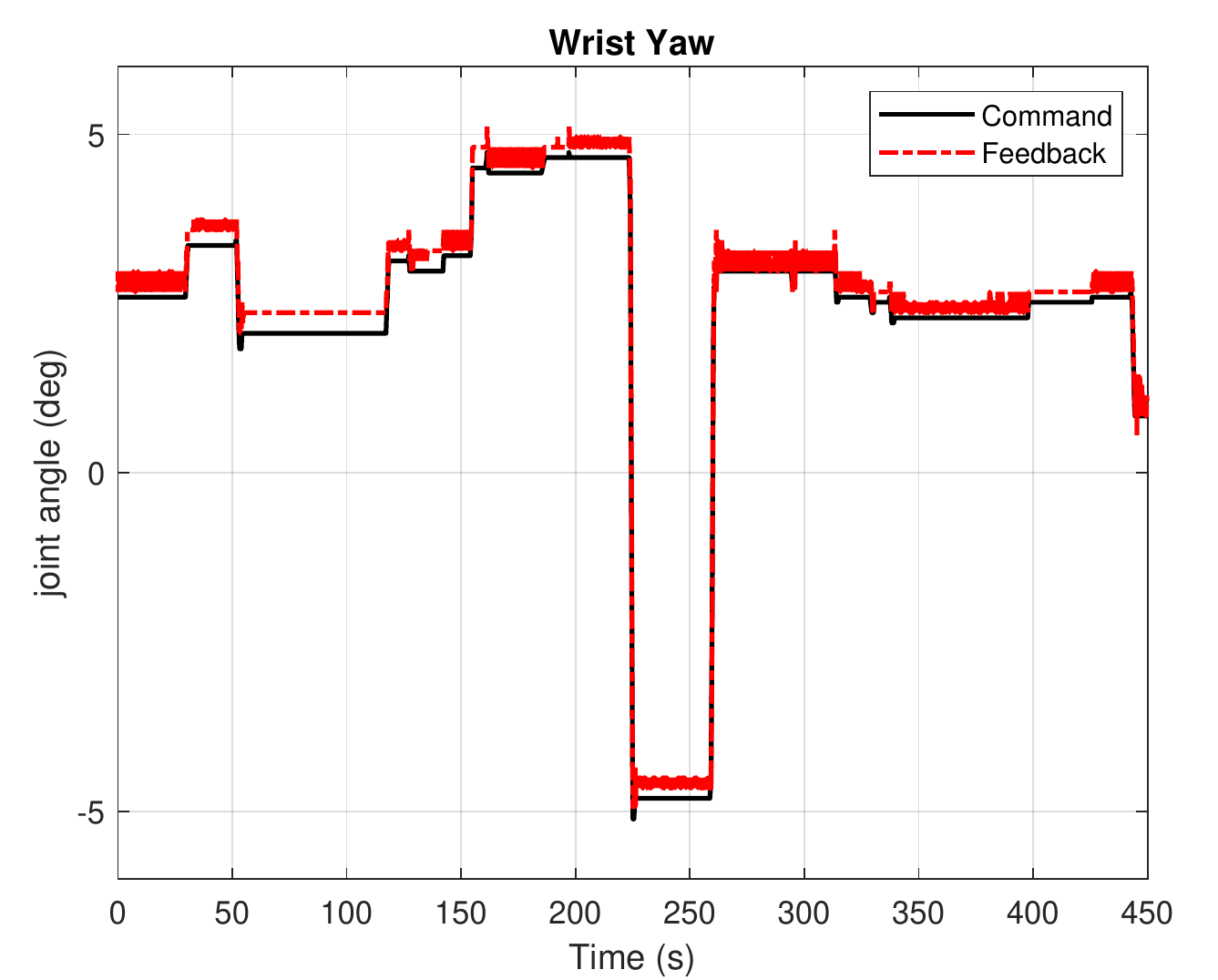}}\hfil
    \subfigure[Wrist Yaw Error Histogram]{\includegraphics[width=0.5\linewidth,trim={4cm 8.5cm 4cm 8.5cm},clip]{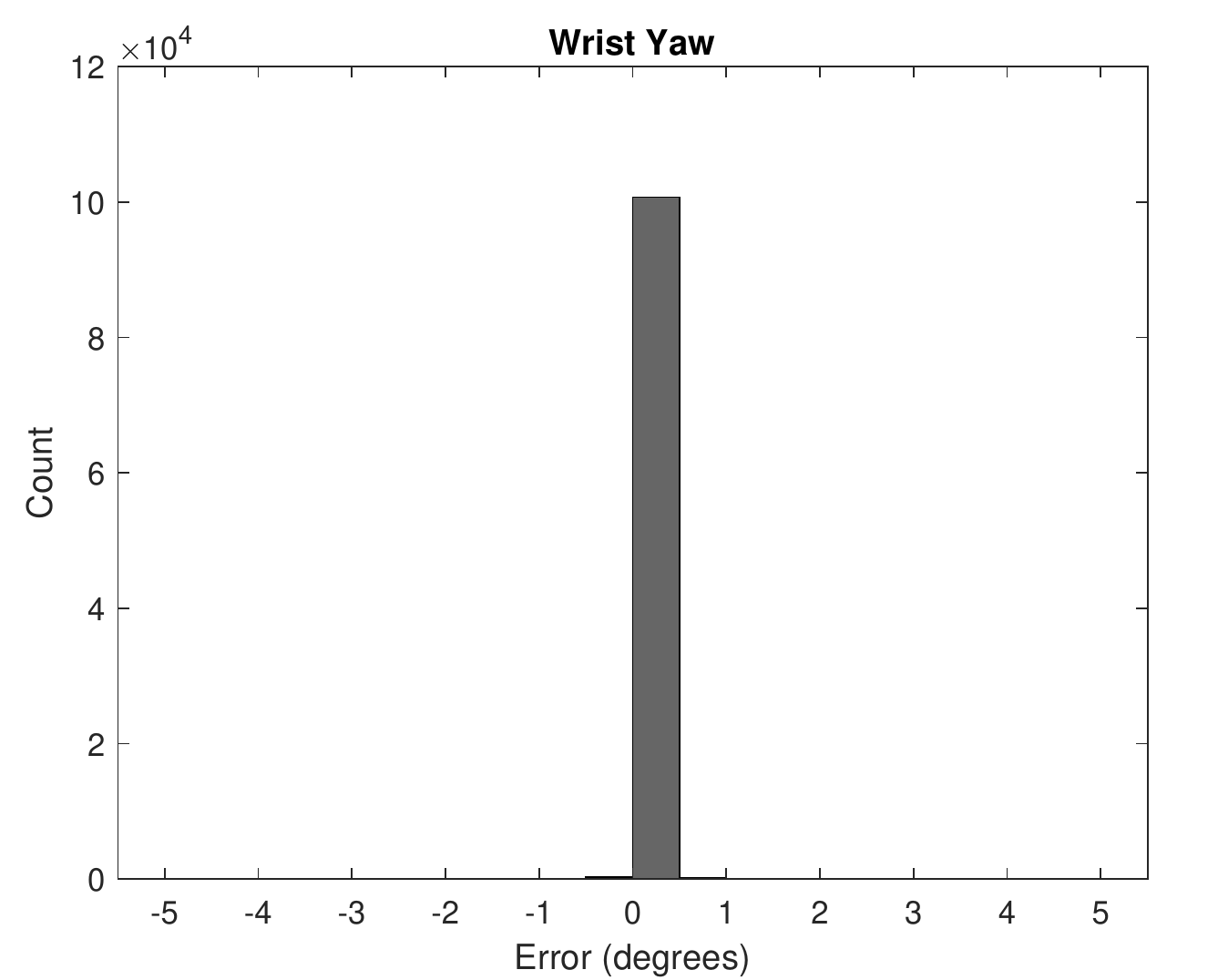}}\\
    \caption{Plot of commanded versus followed joint trajectories for the \textit{NUI} HROV manipulator during the Greece field trials.}
    \label{fig:joint_path_nui}
\end{figure}

\section{Discussion and Future Work}
\label{section:discussion}

During the course of our field trials, we identified operational challenges and failure modes for both the manipulator and vision systems. Addressing these issues is necessary to improve the robustness of the system and is the objective of ongoing research.

\revision{Underwater hydraulic manipulators have inherent characteristics which make them especially challenging to automate and can lead to mission failure if they are not accommodated by the planning and control systems. We identify three particular challenges. The first challenge is the senors that provide joint position feedback (e.g. potentiometers or encoders) can be noisy and prone to drift, resulting in an inaccurate estimate of the manipulator configuration, which can lead to self-collisions or collision with the vehicle or obstacles in the environment. This issue could be mitigated by continuously calibrating the arm using the vision system to detect and compensate for proprioceptive sensor drift. Such a fully automated kinematic calibration procedure is also a practical necessity for a system to be deployed on a space flight mission and would improve calibration accuracy over the manual procedure used in this report. Our ongoing work seeks to apply a feature-based mapping/structure-from-motion framework that jointly performs scene reconstruction and kinematic calibration of the manipulator using features from the fisheye camera. The second challenge is that hydraulic actuators can be imprecise. Typical hydraulic actuator characteristics include a bias between the commanded and reached joint positions, which we observed in the KRAFT manipulator, and hysteresis, where the offset between commanded and reached positions is variable with the direction of joint actuation and the position of the joint. These actuator effects could be mitigated through an adaptive control strategy that adjusts the joint commands to account for detected anomalies or offsets between the commanded and reached configurations. \citet{SIVCEV2018153} reported hysteresis as high as 1.5\degree ~in a Schilling Titan 2 manipulator and subsequently learned joint command offsets in a calibration procedure to compensate for it. The third challenge is that complete joint failure is common for underwater manipulators, reducing the degrees-of-freedom by at least one. Mitigation of this failure would require planning level adaptation to determine what manipulation tasks are still feasible. In this under-actuated operational state, the vehicle mobility might be considered within the kinematic planning to compensate for the loss of manipulator dexterity, drawing from the prior work on free-floating intervention.}
\begin{figure}[!t]
    \centering
    \subfigure[Good image quality and dense point cloud]{\includegraphics[width=1.0\linewidth]{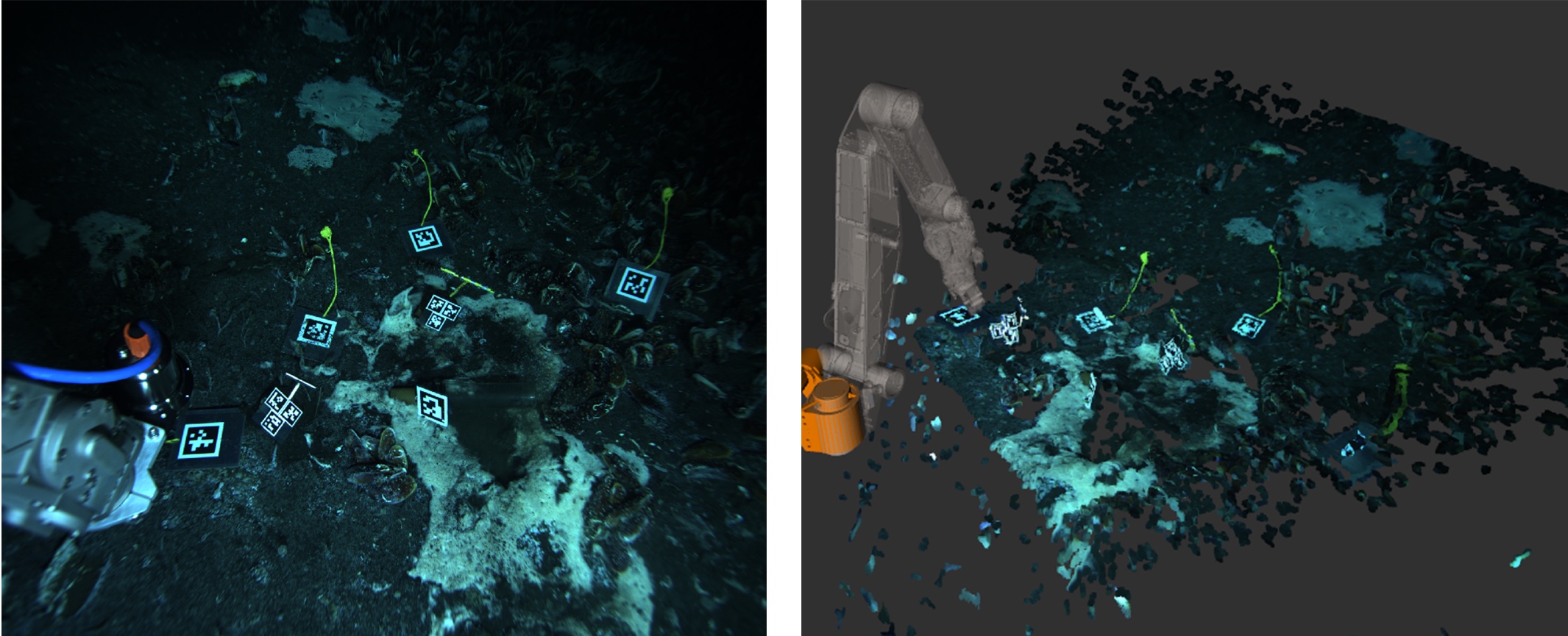}\label{fig:pointclouds_a}}\\
    \subfigure[Poor image quality and sparse point cloud]{\includegraphics[width=1.0\linewidth]{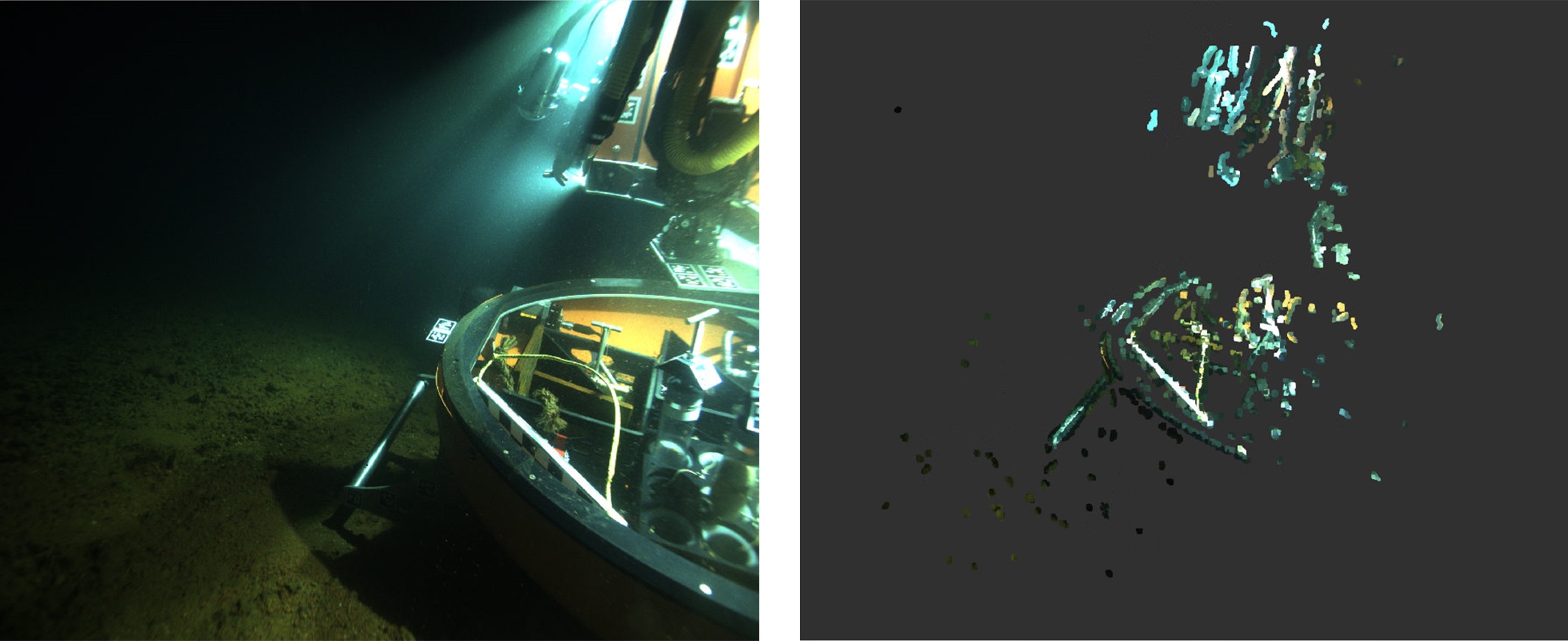}\label{fig:pointclouds_b}}
    \caption{The quality of stereo reconstruction is highly dependent on underwater conditions. Here, we compare stereo point clouds generated using the same camera system and stereo matching method, but with images captured within very different seafloor environments. The left images show the view from the left stereo camera, and the right images show the generated point clouds using a SGM-based stereo method. \subref{fig:pointclouds_a}~The top row was captured in the clear waters off Costa Rica, with even scene lighting and highly textured seafloor. \subref{fig:pointclouds_b}~The bottom row was captured in the Kolumbo caldera, with high backscatter and low texture microbial mats on the seafloor.}
    \label{fig:pointclouds}
\end{figure}

Existing visual reconstruction methods are typically sensitive to lighting, image contrast, and the presence of texture, all of which are highly variable in underwater environments. Figure~\ref{fig:pointclouds} compares point clouds generated using a standard SGM method from the same stereo camera under two different visual conditions. Under near-ideal conditions that include clear water, uniform illumination, and a richly textured seafloor as was the case during our Costa Rica expedition, the point cloud is highly detailed and exhibits a low amount of noise, resulting in a reconstruction that captures fine details of the scene. During the Kolumbo caldera operations, however, fine-grain unconsolidated sediments and amorphous microbial mats blanketed the seafloor, providing little texture for stereo matching. The illumination was uneven and particulates suspended in the water column caused turbidity and light scattering effects that degraded the quality of the images. Under these conditions, stereo matching is only able to recover the well defined edges of the vehicle, while very little of the seafloor is reconstructed. While these examples represent different extremes in underwater visual conditions, it is critical to develop scene reconstruction algorithms that can operate reliably across this range of conditions to achieve robust autonomy. We are currently investigating information-theoretic ways to exploit our ability to control the pose of the wrist-mounted fisheye camera and an adjacent light source to acquire targeted views and actively illuminate the scene in order to improve and extend reconstruction under both good and degraded visual conditions. \revision{Our system implementation currently assumes that the scene is semi-static, i.e., that the ROV position and scene state remain constant during the execution of a manipulator command. For example, if a command is given to grasp a detected tool, the pose of the tool in the scene is assumed to remain fixed during the execution of the grasp. If the tool were to move due to some disturbance before the grasp was completed, the grasp action would likely fail. Future work may integrate an obstacle-aware visual servoing controller to complete grasps or perform precise tool placement, which would reject disturbances to either the scene or the manipulator during task execution. Because our system relies on visual sensing, any disturbance to the scene that results in degraded water clarity, such as stirring up bottom sediment, can necessitate waiting for the water column to clear before the manipulation task can continue. While the KRAFT manipulator used in our field trials is particularly low power when idle, minimizing the energy cost of waiting for visual conditions to improve, future research may improve robustness of the system to degraded visual conditions by fusing acoustic imaging sonar data into the scene mapping framework. Compared to visual sensors, acoustic signals are not dependent on lighting conditions and are not degraded by haze in the water column or a sparsely textured seafloor.}

The technology presented in this report can be directly integrated onto terrestrial-based underwater manipulation platforms in order to decrease operational risk, reduce system complexity, and increase overall efficiency. The current standard for ROV manipulation requires one or more pilots to operate the UVMS based on image feeds from an array of cameras on the vehicle that are displayed on a set of monitors in a ship-side control van. Existing systems do not provide pilots with an estimate of the 3D scene structure, putting the system at risk of collision between the arm and the vehicle or workspace objects. This, together with the cognitive load imposed by having to interpret multiple sensor streams makes it extremely challenging for pilots to establish and maintain situational awareness. The technology presented in this report can be integrated at three different levels with existing ROV systems. At the first and most basic level, the system can act as a decision support tool that provides a detailed real-time 3D visualization of the scene, including the vehicle and manipulator configuration and a reconstruction of the workspace, enabling a pilot to position the manipulator with greater accuracy, speed, and safety. At the second level, the system can be integrated into the manipulator control system for execution monitoring to limit the motion of the manipulator based on scene structure, preventing the pilot from moving the manipulator into collision or a risky configuration~\citep{sivvcev2018collision}. At the third and highest level, manipulation tasks may be fully automated so that a pilot simply selects a desired function or indicates an intent through some mode of communication such as natural language, whereupon the system plans and executes the task while providing visual feedback to the pilot. In this case, it is critical that the pilot be able to override the automated process and take over control of the arm at will.
\begin{table}[h]
\caption{Comparison of the bandwidth requirements for direct teleoperation (top two rows) of an ROV manipulator system compared to operating our high-level autonomy system (bottom two rows), running onboard the vehicle with communication through natural language commands and only the necessary scene state feedback to inform the high-level commands.}
\begin{tabular}{lll}
\toprule
Mode             & Data Type                                & Bandwidth\\
\midrule
Teleoperation Cameras    & Compressed SD or HD @ 10--30\,Hz           &             100\,KB/s--3\,MB/s \\
Teleoperation Manipulator Coms      & 2 way $\times$ 15-200\,Hz $\times$ 18\,B                                  & 540\,B/s--7.2\,KB/s      \\
\hdashline
Natural Language & 1\,B/letter $\times$ $\sim$7 letters/word $\times$  $\sim$2.5\,words/s    &       17.5\,B/s  \\
Scene State Feedback & State and Compressed Images @ 0.1--1\,Hz      &     3--30\,KB/s \\
\bottomrule
\end{tabular}
\label{tab:rates}
\end{table}

\revision{For teleoperation of ROV manipulators, it is standard practice to stream multiple high-definition (HD) camera feeds at 30\,Hz to the operating pilots. In the most bandwidth constrained circumstances, Compressed standard-definition (SD) cameras can be streamed at 10\,Hz to the pilots. At lower image resolutions or framerates, it becomes difficult for pilots to teleoperate the manipulator safely. Our system enables high-level command of the manipulator and mitigates the need for continuous image streams back to the controlling pilot. Single image frames need only be sent when a scene change is detected or on request. Future work on the vision system will develop methods for semantic-level scene understanding, which will further reduce the need for direct image streams back to the pilot. For a semantic aware system, natural language is well suited for human-machine interaction and can drastically reduce the data communication load between the vehicle platform and a remote operator by on-boarding data heavy computation (e.g., image processing) onto the vehicle's local compute system and interfacing with the remote operator through small bandwidth language packets. For our system to operate with pilot oversight, high level commands and sensory feedback need only be streamed at rates which match the dynamics of the scene. In the scenario where the vehicle is set down on the seafloor to collect samples, the relevant scene dynamics can be on the order of seconds, minutes or longer, enabling significant reduction of the communication bandwidth which is vital for remote operations over bandwidth limited connections, such as satellite links. Table~\ref{tab:rates} shows estimated bandwidth range requirements for the manipulator coms and image streams necessary to support direct teleoperation of an ROV manipulator system compared to the bandwidth requirements for natural language communication with the vehicle and only the necessary scene state feedback to inform the high level commands. In the case of direct teleoperation, the manipulator coms can range from 15\,Hz to 200\,Hz two-way communication with a typical packet size of 18\,B. We estimate the image bandwidth for a single SD or HD camera with compressed data streamed at 10\,Hz--30\,Hz, though generally multiple camera views are streamed simultaneously back to the pilot for safe manipulator control. In the case of our high-level automation system, the natural language data rates are based on approximate estimates for the average letter count per word and the speech rate. This data rate represents the expected maximum bandwidth load when transmitted in real-time, as language based communication is intermittent and can be compressed. The scene state feedback includes the vehicle state such as the manipulator joint states and semantic information, such as the type and pose of detected tools. However, the visual scene state feedback takes up the bulk of the bandwidth and is assumed to be encoded as a compressed camera frame or view of the 3D scene reconstruction. As demonstrated in the table, communication requirements to support our high-level system reduce the necessary bandwidth load by at least an order of magnitude compared to the requirements of the most limited direct teleoperation modality.}

Despite the technological challenges in reaching extraterrestrial worlds, the NASA Science Mission Directorate (SMD) sets its first priority ``to discover the secrets of the universe, to search for life, and to protect and improve life on Earth''~\citep{nasa_science_mission_directorate_2020} and ``is undertaking a flagship mission to Jupiter's moon Europa, as its subsurface ocean has great potential to harbor extraterrestrial life.'' A Europa mission concept for a surface lander has reached relative maturity, having passed its delta Mission Concept Review~\citep{hand2021europa}. The sampling system is recognized as being critical to the success of the mission and relies on a robotic arm for ``excavation, collection, and presentation (or transfer) of samples to scientific instruments for observation and analysis''~\citep{europa_lander_2016_report}. Due to the anticipated communication limitations, it is likely that the lander will be required to self-select sampling sites, in which ``the sampling system would be capable of conducting a sampling cycle in a fully autonomous fashion with no input from ground operators, from target selection to sample delivery. This autonomous capability is to guard against a prolonged telecommunications fault during the short mission lifetime, and will be in place to provide added assurance that the mission threshold science would be met''~\citep{europa_lander_2016_report}. Challenges to the sampling system will be exacerbated by ``poorly-characterized terrain at small scales", and ``the terrain immediately in front of the landing spot must suffice for sampling locations; there is no mobility system that can be used to search for a better site''~\citep{europa_lander_2016_report}. The methods we demonstrated in this report for automated manipulator control and sample collection are directly applicable to operations focused on a Ladder of Life detection mission scenario~\citep{neveu2018ladder}. With the exception of the wrist mounted camera, the manipulator and imaging system used in our demonstrations are very similar to the hardware for the Europa lander concept, consisting of a multi-DoF manipulator and vehicle-mounted stereo pair. A primary limiting factor on the integration of our autonomy methods with the lander would be the available computational power. However, for a stationary lander, the visual processing, which is the primary computational bottleneck, could operate at low-frame rates suitable for extraterrestrial exploration, assuming the environment dynamics are sufficiently slow. The methods we describe are also suitable to run on embedded systems and may be optimized accordingly.

\section{Conclusions}
\label{section:conclusion}

An exobiology search mission to distant ocean worlds will require a highly automated exploratory vehicle, capable of operating in extreme conditions for an extended period of time. Such a platform will likely be outfitted with a manipulator to maximize the types of samples that could be collected. In this report we describe a vision system and control framework for automating an ROV manipulator. This architecture is readily integrated onto a wide array of vehicle platforms, and we have demonstrated the viability of the system in the field on two ROVs with different manipulators, including the \textit{NUI} HROV which is dynamically reconfigurable. In November of 2019, we demonstrated planner-controlled sample collection and return within active submarine volcanoes that host diverse assemblages of extremophile organisms. These operation locations served as analogs to environments that may exist within other ocean worlds in our solar system and beyond.

A current limitation of our approach is a semi-static vehicle and scene assumption, where the ROV is held stationary and the scene does not change during execution of a manipulator motion, though the vehicle and scene state may change between motions. The vehicle is typically kept stationary by setting it down on the seafloor before manipulation is initiated. This assumption limits the type of sampling tasks that may be performed with the described system. For example collecting samples from a vertical wall, the underside of an ice shelf, or other moving objects would require free-floating control. Free-floating manipulation is an open problem in robotics, and a promising research direction that directly builds on our demonstrated system is obstacle aware disturbance rejection control of the manipulator. This method is similar to obstacle aware visual servoing, using feature based SLAM with the vision system to compensate for vehicle motions and stabilize the end-effector. A disturbance rejection approach would enhance the flexibility of the system to be easily integrated on different vehicles and manipulators without requiring the generation of complex vehicle and manipulator dynamic models.

While the demonstrated system represents a significant step towards autonomous sample collection and return from seafloor environments, more advancements are required before the system can be deployed reliably in a fully automated fashion. In particular, visual methods must be developed that are robust to the optical challenges of the underwater environment in order to enable safe and targeted sample collection and precision tool handling. These methods must be robust to dynamic scenes, insensitive to the intensity inconsistency of underwater lighting and perform well in sparsely featured and low-textured environments. Fusion of sparse feature based methods for SLAM with learning-based methods for dense scene reconstruction and high-level semantic scene understanding, such as segmentation, object detection and tool pose estimation may provide an appropriate path forward to overcome this challenge.

In summary, automated exploration of unstructured seafloor environments is within reach of current underwater robotic technology. More development is needed, particularly in methods for scene reconstruction and understanding, to make this technology sufficiently reliable for fully automated deployment, but results from our oceanographic expeditions described in this report demonstrate that a wide range of existing ROVs and manipulator systems can be adapted, with moderate effort, for high level automation capabilities.

\subsubsection*{Acknowledgments}
This work was funded under a NASA PSTAR grant, number NNX16AL08G, and by the National Science Foundation under grants IIS-1830660 and IIS-1830500. The authors would like to thank the Costa Rican Ministry of Environment and Energy and National System of Conservation Areas for permitting research operations at the Costa Rican shelf margin and the Schmidt Ocean Institute (including the captain and  crew of the \textit{R/V Falkor}, and  ROV \textit{SuBastian}) for their generous support and making the FK181210 expedition safe and highly successful. Additionally, the authors would like to thank the Greek Ministry of Foreign Affairs for permitting the 2019 Kolumbo Expedition to the Kolumbo and Santorini calderas, as well as Prof.\ Evi Nomikou and Dr.\ Aggelos Mallios for their expert guidance and tireless contributions to the expedition. We would also like to thank Maritech and crew of the \textit{CLV OceanLink}, and the HROV \textit{NUI} crew for their skillful and friendly assistance with integration, testing, and field operations. Finally we would like to thank Prof.\ Blair Thornton for graciously sharing his underwater camera housing design, which was used in our work described here.

\clearpage
\bibliographystyle{apalike}
\bibliography{references}

\end{document}